# An Algorithmic Perspective on Imitation Learning

Takayuki Osa
University of Tokyo
osa@edu.k.u-tokyo.ac.jp

Joni Pajarinen
Technische Universität Darmstadt
pajarinen@ias.tu-darmstadt.de

Gerhard Neumann
University of Lincoln
gneumann@lincoln.ac.uk

J. Andrew Bagnell
Carnegie Mellon University
dbagnell2@andrew.cmu.edu

Pieter Abbeel
University of California, Berkeley
pabbeel@cs.berkeley.edu

Jan Peters
Technische Universität Darmstadt
mail@jan-peters.net

# Contents

## Abstract

As robots and other intelligent agents move from simple environments and problems to more complex, unstructured settings, manually programming their behavior has become increasingly challenging and expensive. Often, it is easier for a teacher to *demonstrate* a desired behavior rather than attempt to manually engineer it. This process of learning from demonstrations, and the study of algorithms to do so, is called *imitation learning*. This work provides an introduction to imitation learning. It covers the underlying assumptions, approaches, and how they relate; the rich set of algorithms developed to tackle the problem; and advice on effective tools and implementation.

We intend this paper to serve two audiences. First, we want to familiarize machine learning experts with the challenges of imitation learning, particularly those arising in robotics, and the interesting theoretical and practical distinctions between it and more familiar frameworks like statistical supervised learning theory and reinforcement learning. Second, we want to give roboticists and experts in applied artificial intelligence a broader appreciation for the frameworks and tools available for imitation learning.

We organize our work by dividing imitation learning into directly replicating desired behavior (sometimes called *behavioral cloning* [Bain and Sammut, 1996]) and learning the hidden objectives of the desired behavior from demonstrations (called *inverse optimal control* [Kalman, 1964] or *inverse reinforcement learning* [Russell, 1998]). In addition to method analysis, we discuss the design decisions a practitioner must make when selecting an imitation learning approach. Moreover, application examples—such as robots that play table tennis [Kober and Peters, 2009] and programs that play the game of Go [Silver et al., 2016]— illustrate the properties and motivations behind different forms of imitation learning. We conclude by presenting a set of open questions and point towards possible future research directions.

# 1

## Introduction

Programming autonomous behavior in machines and robots traditionally requires a specific set of skills and knowledge. However, human experts know how to demonstrate the desired task even if they do not know how to program the necessary behavior in a machine or robot. The purpose of imitation learning is to efficiently learn a desired behavior by imitating an expert's behavior. The application of imitation learning is not limited to physical systems. It can be a powerful tool to design autonomous behavior in systems such as web sites, computer games, and mobile applications. Any system that requires autonomous behavior similar to human experts can benefit from imitation learning.

However, imitation learning may be essential for robotics. It is now considered to be a key technology for applications such as manufacturing, elder care, and the service industry. These robots will be expected to work closely with humans in a dramatic shift from prior uses of robots. Powerful robotic manipulators are dangerous and have therefore been used mainly in constrained, predefined industrial applications; employees must undergo special training before working with them. This is changing due to recent advances in robotics from compute to the use of light, compliant, and safe robotic manipulators. They are ideal for applications where robots work alongside people, such as collaborating with human operators and reducing the physical workload of care givers. These applications require efficient, intuitive ways to teach robots the motions they need to perform from domain experts who may not possess special skills or knowledge about robotics.

In recent years, imitation learning has been investigated as a way to efficiently and intuitively program autonomous behavior[Schaal, 1999, Argall et al., 2009, Billard et al., 2008, Billard and Grollman, 2013, Bagnell, 2015, Billard et al., 2016]. In imitation learning, a human demonstrates how to perform a task. A robotic system learns a policy to execute the given task by imitating the demonstrated motions. Numerous imitation learning methods have been developed and imitation learning has become a gigantic field of research. As a consequence, capturing the entire field of imitation learning is not a trivial task.

The purpose of this survey is to provide a structural understanding of existing imitation learning methods and its relationship with other fields from supervised learning to control theory. We will describe what has been developed in the field of imitation learning and what should be developed in the future.

## 1.1 Key successes in Imitation Learning

One of the earliest and most well-known imitation learning success stories was the autonomous driving project Autonomous Land Vehicle In a Neural Network (ALVINN) at Carnegie Mellon University [Pomerleau, 1988]. In ALVINN, a neural network learned how to map input images to discrete actions in order to drive a vehicle. ALVINN's neural network had one hidden layer with five units. Its input layer had 30 by 32 units; its output layer had 30 units. Although the structure of this network was simple compared to modern neural networks with millions of parameters, the system succeeded at driving autonomously across the North American continent.

The Kendama robot developed by Miyamoto et al. [1996] is another successful application of imitation learning. In the early days of imitation learning, roboticists were mainly interested in teaching higher-level tasks from human demonstrations, such as "pick," "move," and "place" Kang and Ikeuchi [1993], Kuniyoshi et al. [1994]. In those settings, lower-level tasks were often considered to be simple, point-to-point motions. In the late 1990s, this focus shifted from task-level planning to trajectory-level planning. The term "learning from demonstration" has become very popular since its use by S. Schaal and G. Atkeson [Schaal, 1997, Atkeson and Schaal, 1997]. Since then, learning robot motions has been a key domain of imitation learning.

Recently, learning from human demonstrations has benefited from developments in deep neural networks. Recurrent neural networks such as long short-term memory (LSTM) networks Hochreiter and Schmidhuber [1997] have played a significant role in demonstrating how to succeed in many previously difficult sequential tasks by learning from demonstrated data. This includes tasks for generating handwriting Chung et al. [2015], natural language Wen et al. [2015], or image captions Karpathy and Fei-Fei [2015]. Furthermore, AlphaGo, the algorithm which was able to beat a human Go master and which we discuss in more detail in §3.4.2, initializes a deep neural network policy from human demonstrations Silver et al. [2016]. Often these recent approaches require a large amount of data. In §3, we will discuss how to learn from data to reproduce observed behavior in specific problem settings.

## 1.2 Imitation Learning from the Point of View of Robotics

Imitation learning is a class of methods that reproduces desired behavior based on expert demonstrations. In many cases, the experts are human operators and the learners are robotic systems, Thus, imitation learning is a technique that enables skills to be transferred from humans to robotic systems. To perform imitation learning, we need to develop a system that records demonstrations by experts and learns a policy to reproduce the demonstrated behavior from the recorded data. For this purpose, we need to answer the following questions.

**General Aspects:**

1. **Why and when should imitation learning be used?** This question clarifies the motivation for using imitation learning and what we should do with it.

2. **Who should demonstrate?** In many cases, the experts are human operators. Many imitation learning methods implicitly assume that demonstrations are provided by a single expert. When multiple experts are available, we need to decide which one should be imitated or how we can incorporate demonstrations from multiple experts.

3. **How should we record data of the expert demonstrations?** There are multiple ways of recording the behavior of experts. For example, motion capture systems and teleoperated robotic systems record data from expert behavior. This choice is closely related to the embodiment problem between experts and learners, which will be discussed in §3.7.1.

4. **What should we imitate?** The recorded data often includes redundant information about expert behavior. In such cases, features appropriate for the desired behavior should be selected. Meanwhile, the recorded data also includes unnecessary motions, which should not be imitated. The data must be segmented to extract the motions to be imitated.

**Algorithmic Aspects:**

5. **How should we represent the policy?** Expert behavior can be represented using methods such as symbolic representation, trajectory-based representation, and state-action space representation. The choice depends largely on the design of the entire system.

6. **How should we learn the policy?** Many algorithms for learning the policy have been developed over the past several decades. The choice of the algorithm for learning the policy is closely related to the choice of policy representation.

With regard to the first four questions, several survey papers on imitation learning [Argall et al., 2009, Billard et al., 2008, Billard and Grollman, 2013, Billard et al., 2016], provide a taxonomy of imitation learning from the perspective of robotics. Argall et al. [2009] indicate that it is essential to design an imitation learning system by considering the correspondence between the expert and the learner, data acquisition methods, and limitations of the demonstration dataset. Billard et al. [2008, 2016] provide an overview of imitation learning methods and highlight techniques for trajectory learning. However, none of the previous review articles focused on the *design decisions needed to develop new imitation learning algorithms* to enable answering questions five and six related to the algorithmic aspects discussed above. In addition, these articles did not discuss the algorithmic details of existing methods because the enormous amount of prior work on imitation learning makes it challenging to cover the entire range of previous studies.

In this survey, we provide an overview of existing methods from the algorithmic point of view, which will be useful for both readers beginning the practice of imitation learning and readers who want to achieve a deeper understanding of the theoretical aspects of imitation learning. We discuss the design choices which one should consider in order to develop novel imitation learning algorithms. Although our survey cannot be exhaustive, we discuss the algorithmic details of existing algorithms as much as possible, which will be useful to readers who want to implement imitation learning techniques. Additionally, we develop an information theoretic understanding of existing methods, which will help readers to understand how existing methods relate to each other and figure out how they could be extended.

Let us illustrate how different design choices of imitation learning algorithms can be made in different applications. Figure 1.1 shows three applications of imitation learning: 1) an RC helicopter, 2) robotic surgery, and 3) quadruped robot locomotion. In these applications, design of the policies for motion planning and control vary. Abbeel et al. [2010] demonstrates acrobatic RC helicopter flight by learning from trajectories demonstrated by a human expert. In this system, the desired

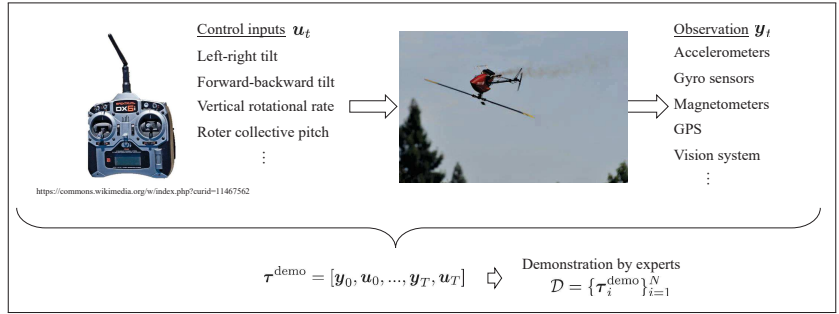

(a) Learning of acrobatic RC helicopter maneuvers [Abbeel et al., 2010]. The trajectories for acrobatic flights are learned from a human expert's demonstrations. To control the system with highly nonlinear dynamics, iterative learning control was used.

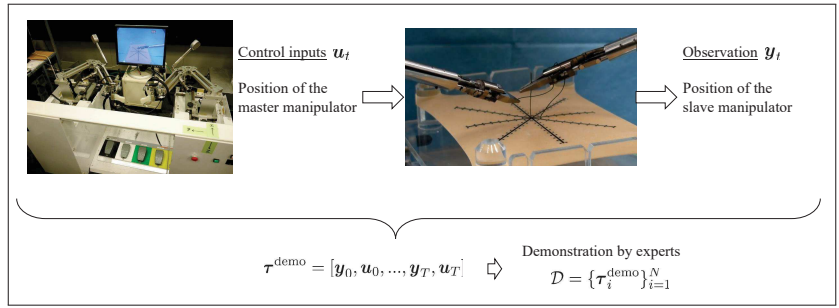

(b) Learning with a teleoperated system [Osa et al., 2014] where a position/velocity controller is available. To generalize the trajectory to different situations, a mapping from task situations to trajectories is learned from demonstrations under various situations.

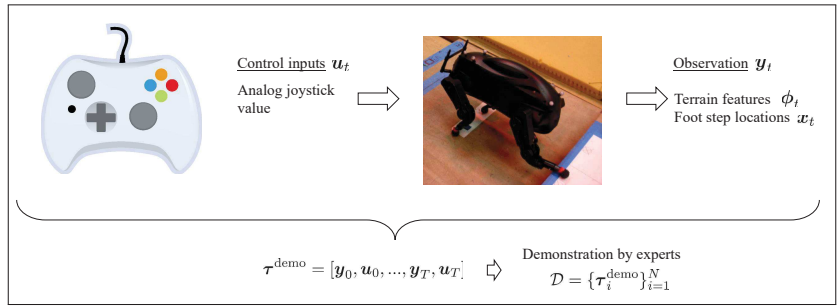

(c) Learning quadruped robot locomotion [Zucker et al., 2011]. The footstep planning was addressed as an optimization of the reward/cost function, which was recovered from the expert demonstrations. Learning the reward/cost function allows the footstep planning strategy to be generalized to different terrains.

**Figure 1.1:** Observations $y$ and control inputs $u$ for imitation learning in (a) helicopter flight, (b) surgery, and (c) locomotion. Motion planning is formulated in different ways in these examples.

trajectories of acrobatic flights were learned from demonstrations with a supervised learning method. Osa et al. [2017b] also learned trajectories for autonomous knot tying from demonstrations by a human expert. To generalize a trajectory, Osa et al. [2017b] learned a direct mapping from task situations (contexts) to trajectories using demonstrations recorded under various situations. Contrary to [Abbeel et al., 2010, Osa et al., 2017b], Zucker et al. [2011] formulated footstep planning for quadruped robot locomotion as an optimization of the reward/cost function. The reward/cost function was recovered from demonstrations. In [Zucker et al., 2011], learning the reward/cost function as a function of terrain features enables the footstep planning strategy to be generalized to different terrains. Learning such reward/cost functions for manipulation tasks like as knot-tying [Osa et al., 2017b] is not trivial, since complex manipulation tasks often require nonlinear reward/cost functions.

Methods for learning policies also differ between applications. The observation and control inputs of the RC helicopter system are much noisier than those of the other two systems, and its dynamics are highly nonlinear [Abbeel et al., 2010]. Therefore, it is essential to estimate the true state using various sensory information and learn an adaptive controller through iterations of trials to achieve acrobatic RC helicopter flight. On the other hand, we can assume that the system state is precisely known and a position/velocity controller is available in the case of the tele-operation system in [Osa et al., 2014], which simplifies imitation learning significantly. In [Osa et al., 2014], the conditional trajectory distribution given a context can be learned with a simple regression method, and the planned trajectory can be executed by a standard velocity controller. In locomotion planning for a quadruped robot in [Zucker et al., 2011], estimating the reward/cost function requires an iterative learning process with virtual simulation of the learned policy. As one can see from these examples, learning methods can be very different between applications.

To apply imitation learning, it is essential to identify the structure of the system, formulate a given problem, and design an algorithm to solve the problem efficiently. In this survey, we focus on the algorithmic aspects of imitation and discuss necessary design choices, exploring various solutions proposed by previous studies.

In the rest of this chapter, we introduce several concepts in machine learning that are essential to understand imitation learning algorithms. We discuss the design choices of imitation learning algorithms in Chapter 2. We describe the details of behavioral cloning methods and inverse reinforcement learning methods in Chapters 3 and 4, respectively. To conclude, we list open questions of imitation learning in Chapter 5.

## 1.3 Key Differences between Imitation Learning and Supervised Learning

The imitation learning problem has special properties that distinguish it from the better known supervised learning setting [Shalev-Shwartz and Ben-David, 2014] : 1) the solution may have important structural properties including constraints (for example, robot joint limits), dynamic smoothness and stability, or leading to a coherent, multi-step plan [Bagnell, 2015]; 2) the interaction between the learner's decisions and its own input distribution (an *on-policy* versus *off-policy* distinction) , and 3) the increased necessity of minimizing the typically high cost of gathering examples.

As we learn a policy $\pi$ from a dataset $\mathcal{D}$, imitation learning is closely related to supervised learning, and is particularly related to the field of *structured prediction* [Daumé III et al., 2009, Ratliff et al., 2006a, Taskar, 2005] , where the task is to learn a mapping from inputs $\boldsymbol{x}$ to a complex, structured output $\boldsymbol{y}$ (plans, parse trees, complex motions). Reductions of structured prediction to sequential decision [Daumé III et al., 2009], and reductions of imitation learning to structured prediction [Ratliff et al., 2006b] show the close connection, and cross-fertilization between these research areas has been important for both. In practice, distinctions arise because of the structural properties of policies we attempt to imitate, and the difficulty of "resetting" state and restarting predictions is too costly or even infeasible in most imitation learning settings because a physical system is often involved.

In addition, it is often the case that the embodiments of the expert and the learner are different. For example, when transferring human skills to a humanoid robot, the motion captured from a human expert may be infeasible for the humanoid. In such a case, the demonstrated motion needs to be adapted to be feasible for the humanoid. This kind of adaptation is less common in the standard supervised learning.

In machine learning, the prediction problem where the source domain distribution and the target domain distribution are different is often referred to as "*covariate shift*" or "*domain adaptation*" [Sugiyama, 2015]. In imitation learning, the source domain corresponds to expert demonstrations and the target domain to learner reproductions. In imitation learning, the demonstration dataset does not cover all possible situations since collecting expert demonstrations to cover all situations is usually too expensive and time-consuming. As a result, the learner often encounters states which were not encountered by the expert during demonstrations, which means that the target domain distribution is different from the source distribution. Therefore, covariate shift or domain adaptation is closely related to imitation learning [Bagnell, 2015].

Imitation learning is also closely related to reinforcement learning (RL), which tries to obtain a policy that maximizes an expected reward [Sutton and Barto, 1998] signal. In RL, we employ a reward function that encourages a desired behavior. However, in imitation learning we often assume optimal (or at least "good") expert demonstrations which are not available in basic reinforcement learning, and which provide prior knowledge that allows for dramatically more efficient methods. Recent work by Sun et al. [2017] demonstrates a potentially *exponential* decrease in sample complexity in learning a task by imitation rather than by trial-and-error reinforcement learning, and empirical results have long shown such benefits [Silver et al., 2016, Kober and Peters, 2009, Abbeel et al., 2010]. Moreover, in the imitation learning setting, as we detail below, we may or may not have access to a true reward function.

## 1.4 Insights for Machine Learning and Robotics Research

As imitation learning offers intuitive ways to program robotic motions by demonstrating the desired motion, imitation learning attracted interests from robotic researchers. The robotics community has devel-oped many imitation learning methods for motion planning and robot control. When planning a trajectory for a robotic system, it is often necessary to make sure that a planned trajectory satisfies some constraints such as smooth convergence to a new goal state. For this reason, robotics researchers have developed "custom" trajectory representations that explicitly satisfy constraints necessary for robotic applications. Machine learning techniques are often used as a part of such frameworks. However, robotics researchers need to be aware that rich set of algorithms have been developed by the machine learning community and some of new algorithms might eliminate the need for customizing policy or trajectory representation.

For machine learning researchers, imitation learning offers interesting practical and theoretical problems, which differ from standard supervised and reinforcement learning settings. Although imitation learning is closely related to structured prediction, it is often challenging to apply existing machine learning methods to imitation learning, especially robotic applications. In imitation learning, collecting demonstrations and performing rollouts are often expensive and time-consuming. Therefore, it is necessary to consider how to minimize these costs and perform learning efficiently. In addition, embodiments and observability of the learner and the expert are different in many applications. In such cases, the demonstrated motion needs to be adapted based on the learner's embodiment and observability. These difficulties in imitation learning present new challenges to machine learning researchers.

## 1.5   Statistical Machine Learning Background

To understand imitation learning algorithms, familiarity with several concepts in statistical machine learning is essential. In this section, we briefly introduce the notation we use and these concepts.

### 1.5.1   Notation and Mathematical Formalization

Before introducing important concepts in machine learning, we introduce the notation in this article. Table 1.1 summarizes our notation. Throughout this survey, we use the bold style for vector values, and the non-bold style for scalar values. Demonstrations by an expert are often given as a set of trajectories. In this case, the dataset of demonstrations is given by $\mathcal{D} = \{\boldsymbol{\tau}^0, \ldots, \boldsymbol{\tau}^m\}$. We use the lower script to denote the time index; $\boldsymbol{x}_t$ represents the state of the system at time step $t$. We review many methods that manipulate probability distributions in various ways. To make equations concise, the probability distribution induced by the experts' policy is denoted by $q$, and the distribution induced by the learner's policy is denoted by $p$. For example, $p(\boldsymbol{\tau})$ represents the probability distribution over trajectories induced by the learner's policy. The term "action" is mainly used in machine learning community, and "control input" is mainly used in robotic community and control theory community. Since imitation learning methods have been developed in all of these communities, we use the word "action" and "control input" interchangeably. We use the term "context" to refer to the condition relevant to the task. The context $s$ can be the initial state of the system $x_0$ or the state of relevant objects. For instance, the position of the ball can be part of the context in a hitting-a-ball task. We use $T$ to denote the finite horizon of the trajectory. Therefore, the total number of the time steps of a single trajectory is $T+1$ in our notation.

**Table 1.1:** Table of Notation. We use a notation common in the control literature for states and controls.

| | |
|---|---|
| $\boldsymbol{x}$ | system state |
| $\boldsymbol{s}$ | context |
| $\boldsymbol{\phi}$ | feature vector |
| $\boldsymbol{u}$ | control input/action |
| $\boldsymbol{\tau}$ | trajectory |
| $\pi$ | policy |
| $\mathcal{D}$ | dataset of demonstrations |
| $q$ | probability distribution induced by an expert's policy |
| $p$ | probability distribution induced by a learner's policy |
| $t$ | time |
| $T$ | finite horizon |
| $N$ | number of demonstrations |
| E | superscript representing an expert <br> e.g. $\pi^{\text{E}}$ denotes an expert's policy |
| L | superscript representing a learner <br> e.g. $\pi^{\text{L}}$ denotes a learner's policy |
| demo | superscript representing a demonstration by an expert <br> e.g. $\boldsymbol{\tau}^{\text{demo}}$ denotes a trajectory demonstrated by an expert |

### 1.5.2 Markov Property

A sequence of states $x_0, ..., x_t$ is a Markov chain if at any time $t$, the future states $x_{t+1}, x_{t+2}, ...$ depend on the history $x_0, ..., x_t$ only through the present state $x_t$ [Serfozo, 2009]. In other words, the next state $x_{t+1}$ only depends on the current state $x_t$ in a Markov chain. This property is called the *Markov property*.

### 1.5.3 Markov Decision Process

A Markov decision process (MDP) is a process that satisfies the Markov property. If the state and action spaces are finite, then it is called a finite Markov decision process (finite MDP) [Sutton and Barto, 1998]. An MDP is defined as a tuple $(\mathcal{X}, \mathcal{U}, \mathcal{P}, \gamma, D, R)$. $\mathcal{X}$ is a finite set of states; $\mathcal{U}$ is a set of control inputs; $\mathcal{P}$ is a set of state transitions probabilities; $\gamma \in [1, 0)$ is a discount factor; $D$ is the initial-state distribution from which the initial state $x_0$ is drawn; and $R : \mathcal{X} \mapsto \mathbb{R}$ is the reward function.

### 1.5.4 Entropy

Given the random variable $x$ and its probability distribution $p(x)$, the entropy

$$H(p) = -\int p(x) \ln p(x) dx \qquad (1.1)$$

is defined as the amount of information conveyed by transmitting $x$ [Bishop, 2006]. Note that the entropy $H(x)$ is a convex function.

### 1.5.5 Kullback-Leibler (KL) Divergence

In the field of information geometry, the KL divergence is used to quantify a difference between two probability distributions[Kullback and Leibler, 1951], i.e.,

$$D_{\text{KL}}\left(p(\boldsymbol{x})||q(\boldsymbol{x})\right) = \int p(\boldsymbol{x}) \ln \frac{p(\boldsymbol{x})}{q(\boldsymbol{x})} d\boldsymbol{x}. \tag{1.2}$$

Since the KL divergence identifies a difference between two probability distributions, it is useful for cases in which stochastic policies are going to be learned, or stochastic trajectories result from a deterministic policy. Please note that the KL divergence is not symmetric, therefore $D_{\text{KL}}(p||q) \neq D_{\text{KL}}(q||p)$. The KL divergence can be obtained as a Bregman divergence derived from the negative entropy [Amari, 2016] and is widely used as a measure in multiple imitation learning approaches.

### 1.5.6 Information and Moment Projections

One common approach to learning a policy from a dataset is to consider "projecting" that dataset onto the space of the policy model. Information theory emphasizes two kinds of projections: the Information(I)-projection and the Moment(M)-projection [Bishop, 2006]. Using the Kullback-Leibler (KL) divergence [Kullback and Leibler, 1951], the I-projection is

$$p^* = \arg\min_p D_{\text{KL}}(p \parallel q), \tag{1.3}$$

and, the M-projection

$$p^* = \arg\min_p D_{\text{KL}}(q \parallel p). \tag{1.4}$$

As the KL divergence is not symmetric, these two projections result in different solutions when a given distribution is multi-modal as shown in Figure 1.2. While the M-projection averages over the several modes, the I-projection concentrates on a single mode. Performing the I-projection is often not straight-forward, although the M-projection can often be performed relatively easily by maximizing the likelihood with respect to a given training dataset [Bishop, 2006].

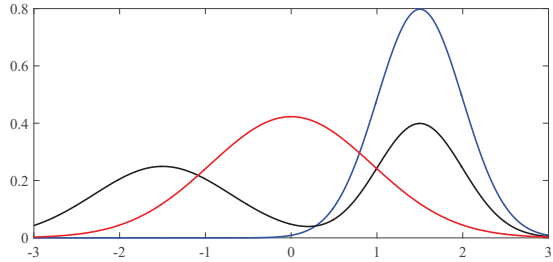

**Figure 1.2:** Illustration of I- and M- projections. Given a distribution with two modes as shown in black, M-projection will give a solution that averages over two modes as shown in red. On the contrary, I-projection will give a solution that concentrates on one of the modes.

### 1.5.7 The Maximum Entropy Principle

Let us consider a probability distribution $p(\boldsymbol{x})$ that matches the features of an unknown distribution $q$, i.e. it satisfies

$$\mathbb{E}_p[\boldsymbol{\phi}(\boldsymbol{x})] = \mathbb{E}_q[\boldsymbol{\phi}(\boldsymbol{x})],$$

where $q(\boldsymbol{x})$ is an unknown probability distribution and $\mathbb{E}_q[\boldsymbol{\phi}(\boldsymbol{x})]$, which is the expectation of a feature function $\boldsymbol{\phi}(\boldsymbol{x})$, is available. As there are typically an infinite amount of such distributions, we need an additional constraint to obtain a unique solution [Amari, 2016].

The maximum entropy principle [Jaynes, 1957] suggests to choose a distribution that maximizes the entropy

$$H(p) = -\int p(\boldsymbol{x}) \ln p(\boldsymbol{x}) d\boldsymbol{x}$$

among the distributions that satisfy $\mathbb{E}_p[\boldsymbol{\phi}(\boldsymbol{x})] = \mathbb{E}_q[\boldsymbol{\phi}(\boldsymbol{x})]$. From this constrained optimization program, the maximum entropy distribution can be computed as

$$p(\boldsymbol{x}) \propto \exp\left(\boldsymbol{w}^\top \boldsymbol{\phi}(\boldsymbol{x})\right), \quad (1.5)$$

where $\boldsymbol{w}$ is a vector-valued Lagrangian multiplier for the feature matching constraint. While the maximum entropy principle does not directly translate into a practical algorithm, it uncovers an interesting observation. Every distribution that is in a log-linear representation given by Equation 1.5, is the maximum entropy distribution that can match specific feature expectations given by the feature vector $\boldsymbol{\phi}(\boldsymbol{x})$. This is true for typical distributions from the exponential family such as the Gaussian distribution, which is the maximum entropy distribution that matches first and second order moments. The notion of Maximum Entropy generalizes to Maximum Causal Entropy, which turns out to be a natural notion of uncertainty for dynamical systems [Ziebart et al., 2013].

### 1.5.8 Background: Reinforcement Learning

Reinforcement learning is a class of methods that autonomously learns policies through iterations of trials and evaluations. The goal of reinforcement learning is to learn a policy $\pi$ that maps the state of the system to the control input so as to maximize the expected reward $J(\pi)$. The reward $r_t$ represents the quality of the given state, action or trajectory at time $t$. For example, $r_t$ could be large when a robot is close to the desired trajectory and small when the robot is far from the trajectory, or, $r_t$ could be large for stable robot grasps and small for unstable ones. With a finite horizon $T$, the expected return is given by the accumulation of the reward at each time step,

$$J(\pi) = \mathbb{E}\left[\sum_{t=0}^{T} r_t \,\middle|\, \pi\right]. \tag{1.6}$$

Alternatively, the discounted accumulated reward is used for the infinite horizon scenario, i.e.,

$$J(\pi) = \mathbb{E}\left[\sum_{t=0}^{\infty} \gamma^t r_t \,\middle|\, \pi\right], \tag{1.7}$$

where the discounted factor $\gamma$ controls the trade-off between shorter term rewards and longer term rewards. The desired policy $\pi^*$ is given by

$$\pi^* = \arg\max_{\pi} J(\pi). \tag{1.8}$$

The value of a state $\boldsymbol{x}$ under a policy $\pi$ can be computed as the expected reward when starting from $\boldsymbol{x}$ and following $\pi$

$$V^\pi(\boldsymbol{x}) = \mathbb{E}\left[\sum_{t=0}^{\infty} \gamma^t r_t \,\middle|\, \boldsymbol{x}_0 = \boldsymbol{x}, \pi\right]. \tag{1.9}$$

$V^\pi(\boldsymbol{x}_t)$ is often called the *value function* [Sutton and Barto, 1998]. Likewise, the value of taking action $\boldsymbol{u}$ in state $\boldsymbol{x}$ under a policy $\pi$ can be computed as the expected reward when starting from the action $\boldsymbol{u}$ in a state $\boldsymbol{x}$ and thereafter following policy $\pi$

$$Q^\pi(\boldsymbol{x}, \boldsymbol{u}) = \mathbb{E}\left[\sum_{t=0}^{\infty} \gamma^t r_t \bigg| \boldsymbol{x}_0 = \boldsymbol{x}, \boldsymbol{u}_0 = \boldsymbol{u}, \pi\right]. \qquad (1.10)$$

$Q^\pi(\boldsymbol{x}_t, \boldsymbol{u}_t)$ is often called the *action-value function* [Sutton and Barto, 1998].

For an overview of reinforcement learning methods, please refer to [Sutton and Barto, 1998, Szepesvari, 2010, Wiering and van Otterlo, 2012, Sugiyama et al., 2013] and for an overview in reinforcement learning in robotics, please refer to Kober et al. [2013], Deisenroth et al. [2013b].

## 1.6  Formulation of the Imitation Learning Problem

The goal of imitation learning is to learn a policy that reproduces the behavior of experts who demonstrate how to perform the desired task. Suppose that the behavior of the expert demonstrator (or the learner itself) can be observed as a trajectory $\boldsymbol{\tau} = [\boldsymbol{\phi}_0, ..., \boldsymbol{\phi}_T]$, which is a sequence of features $\boldsymbol{\phi}$. The features $\boldsymbol{\phi}$, which can be the state of the robotic system or any other measurements, can be chosen according to the given problem. Please note that the features $\phi$ do not have to be manually specified, and $\boldsymbol{\phi}$ could be as general as simply pixels in raw images.

Often, the demonstrations are recorded under different conditions, for example, grasping an object at different locations. We will refer to these task conditions as context vector $\boldsymbol{s}$ of the task which is stored together with the feature trajectories. The context $\boldsymbol{s}$ can contain any information relevant to the task, e.g., the initial state of the robotic system or positions of target objects. Note that, as the context describes the current task, it is typically fixed during task execution and the only dynamic aspects of the problem are the state features $\boldsymbol{\phi}_t$. Optionally, a reward signal $r$ that the expert is trying to optimize is also available in some problem settings [Ross and Bagnell, 2014].

In imitation learning, we collect a dataset of demonstrations $\mathcal{D} = \{(\boldsymbol{\tau}_i, \boldsymbol{s}_i, r_i)\}_{i=1}^{N}$ that consists of pairs of trajectories $\boldsymbol{\tau}$, contexts $\boldsymbol{s}$, and optionally reward signals $r$. The data collection process can be both offline and online. Using the collected dataset $\mathcal{D}$, a common *optimization-based strategy* learns a policy $\pi^*$ that satisfies

$$\pi^* = \arg\min D\left(q(\boldsymbol{\phi}), p(\boldsymbol{\phi})\right), \qquad (1.11)$$

where $q(\boldsymbol{\phi})$ is the distribution of the features induced by the experts' policy, $p(\boldsymbol{\phi})$ is the distribution of the features induced by the learner, and $D(q, p)$ is a similarity measure between $q$ and $p$. Both offline and online learning scenarios of this problem have been considered [Ross et al., 2011]. Please note that, when the dataset contains demonstrations of multiple tasks and the contexts include information of each task, this problem can be considered multitask learning as in recent work by Duan et al. [2017], Finn et al. [2017a,b].

In addition, we often have access to an environment such as a simulator or a physical robotic system where we can perform and evaluate a policy through interaction. This simulator can be used to gather new data and iteratively improve the policy to better match the demonstrations.

# 2

# Design of Imitation Learning Algorithms

In this chapter, we discuss the design choices of imitation learning methods. First, we describe what design choices need to be considered, and we then discuss what options we can consider for each design decision. Thereafter, we discuss imitation learning methods from an information theoretic point of view.

## 2.1 Design Choices for Imitation Learning Algorithms

When developing an imitation learning method, it is necessary to make several design choices to formalize the problem. In this section, we present a list of some of these design choices.

- **Access to the reward function: imitation learning or reinforcement learning.** A central distinction in imitation learning is whether or not the learner has access to both an expert demonstrator *and* a reward signal that the expert is attempting to optimize. For instance, in learning to play Atari games [Mnih et al., 2015] or play Go [Silver et al., 2016] there is an unambiguous score metric. On the other hand, there exists tasks where it is feasible for the expert to demonstrate the optimal behavior and it is hard to define the reward manually including, learning to drive a car by demonstration [Pomerleau, 1988] and complex manipulation such as knot-tying [Osa et al., 2017b].

One might naturally ask what benefit is conferred by an expert if a reward signal is available– surely we can simply solve the problem by reinforcement learning? The expert's role is to reign in the need for tremendous and expensive *global* exploration. This has been consistently demonstrated empirically to speed learning even on problems with a clear metric (e.g., the ball-in-a-cup task in [Kober and Peters, 2009]) and recently shown theoretically to provide a potentially exponential improvement in the number of samples required to learn [Sun et al., 2017]. The most common approach to leverage such information is initialize a policy by imitation learning with coarse demonstration and refined by reinforcement learning through trial and error [Silver et al., 2016, Tesauro, 1995]. Algorithms like SEARN [Daumé III et al., 2009] and AggreVaTe [Ross and Bagnell, 2014, Sun et al., 2017], intermix the process of imitation and reinforcement– the learner attempts multiple actions and the expert provides the best strategy or an estimate of *cost-to-go* given the learner's decision. This intermixing ensures that the learner is able (with enough samples and representational power) to recover a policy that is guaranteed to be nearly as good as the expert (and can be much better), and prevents small mistakes from cascading into poor overall behavior.

The emergence of the "V-style jump" [Maryniak et al., 2009] shown in Figure 2.1 in ski jumping is a textbook example of such imitation learning by humans. Although it took decades to be recognized, soon after some athletes achieved successful results with the V-style jump in 1990s, it has become prominent in the sport and has been mastered by all the athletes performing ski jumps. This example illustrates that local optimization around the initial demonstration can only find local optima while imitation learning leads to fast skill acquisition.

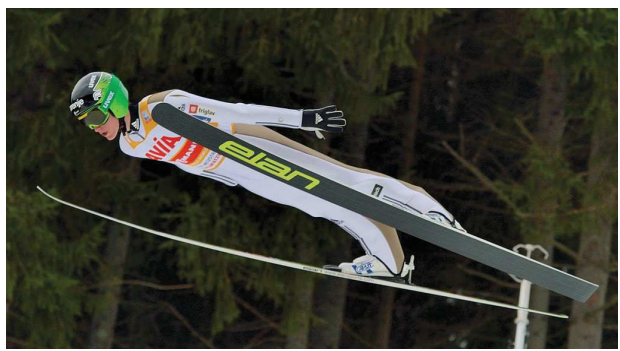

**Figure 2.1:** A ski jumper flies through the air using the highly aerodynamic "V-style". "V-style" was adopted by most ski jumpers in the 1990s after some jumpers demonstrated impressive results with the style (public domain picture from Wikimedia Commons).

- **Parsimonious description of the desired behavior: behavioral cloning or inverse reinforcement learning.** Data efficient learning demands we identify the most compact representation of a behavior. Often a direct mapping from features to trajectories/actions is the most parsimonious description of the policy and the approach known as behavioral cloning approach is used. However, particularly for problems where the behavior is, crudely speaking, *deliberative* and focused on long-horizon planning, the most parsimonious description of the policy may be to encode the policy as the solution of an optimization or planning problem [Ratliff et al., 2009, Bagnell, 2015] Inverse Optimal Control approaches learn a (surrogate) cost function so that the behavior that results from solving that optimization is in some sense similar to that demonstrated by the expert.

- **Access to system dynamics: model-based or model-free.** Access to system dynamics is required for making some problems tractable. For instance, estimation of the system dynamics is often required for motion planning in under-actuated robots, in which accurate controllers are not available. Meanwhile, access to the system is not necessary when a controller of sufficient quality is available. It is desirable to avoid learning of the system dynamics because it is not a trivial problem. Thus, it is essential to identify whether access to system dynamics is necessary for controlling the given system or not.

- **Similarity measure between policies.** In the event that there is not a clear notion of reward function being optimized, a surrogate notion of similarity between the experts' policy and the learner's policy needs to be established to reproduce the behavior of the expert. This similarity can be defined at the level of individual decisions, although it is usual preferred that the notion of similarity be defined over *trajectories* the learner and system take together [Ziebart et al., 2013].

- **Features.** It is essential to select appropriate features that enable the desired behavior to be expressed. Features should contain enough information to solve the problem while limiting the complexity of learning. The features can be various measurements related to the desired task, such as kinematic/dynamic state of the robotic system and/or the surrounding objects. Learning techniques, based on deeper representations have enabled features representations to be at least partially extracted automatically, e.g., using deep learning [Ratliff et al., 2006a, Bradley, 2010, Grubb and Bagnell, 2010, Levine et al., 2016, Ho and Ermon, 2016, Finn et al., 2016b].

- **Policy representation.** Policy representation needs to be chosen such that the desired behavior can be properly captured. For example, a policy can be represented by a neural network or a linear function. With respect to the task abstraction level, we need to decide at which level of the task we learn, such as task level, trajectory level, and action-state level. While it is necessary to select a sufficiently informative representation to model the desired behavior, increasing the complexity of policy representation usually leads to the increase of the required training data and learning time.

As one can see above, these design choices are not independent and the order of these design choices are flexible. For example, the choice of similarity measures between policies is related to the choice of policy representations. In the following sections, we present possible options for some of these design choices.

## 2.2 Behavioral Cloning and Inverse Reinforcement Learning

One way to obtain a policy that reproduces the demonstrated behavior is to learn a policy that directly maps from the input to the action/trajectory. In problems, where a dataset of demonstrated trajectories with state-action pairs and contexts $\mathcal{D} = \{(\boldsymbol{x}_t, \boldsymbol{s}_t, \boldsymbol{u}_t)\}$ is given, we can directly compute a mapping from states or/and contexts to control inputs as

$$\boldsymbol{u} = \pi(\boldsymbol{x}_t, \boldsymbol{s}_t). \tag{2.1}$$

This kind of policy can be usually obtained through a standard supervised learning method. Learning a policy that directly maps from the state or/and the context to the control input is often referred to as *Behavioral Cloning* (BC) [Bain and Sammut, 1996].

Alternatively, given a reward signal, a policy can be obtained so as to maximize the expected return. Such a policy can be expressed as

$$\pi = \arg\max_{\hat{\pi}} J(\hat{\pi}), \tag{2.2}$$

where $J(\hat{\pi})$ is the expectation of the accumulated reward given the policy $\pi$ as in (1.7). However, the reward function is considered unknown and needs to be recovered from expert demonstrations under the assumption that the demonstrations are (approximately) optimal w.r.t. this reward function. Recovering the reward function from demonstrations is often referred to as *Inverse Reinforcement Learning* (IRL) [Russell, 1998] or *Inverse Optimal Control* (IOC) [Moylan and Anderson, 1973].

BC and IRL form two major classes of imitation learning methods. In order to select one of BC and IRL, it is essential to consider *what is the most parsimonious description of the desired behavior*? The policy learned by an IRL method is valid as long as the estimated reward function represents the desired behavior appropriately, while a policy learned by a BC method is valid as long as the learned mapping from states to actions is valid. A choice between BC and IRL is to select the best way to describe the desired behavior, which is totally dependent on a given problem setting. It is essential to analyze how the desired behavior should be performed when applying imitation learning methods.

## 2.3 Model-Free and Model-Based Imitation Learning Methods

Whether we access the system dynamics for imitation learning or not is one of the crucial design decisions. Although learning and leveraging the system dynamics often enables data-efficient learning with a system that has nonlinear and unknown dynamics, learning the system dynamics can be often challenging. In the reinforcement learning literature, methods that learn a forward model of the system and leverage it for learning a policy are often referred to as *model-based*, while methods that do not explicitly learn a forward model of the system are referred to as *model-free* [Kober et al., 2013, Deisenroth et al., 2013b]. In this survey, we apply the same categorization to imitation learning methods. Table 2.1 shows a summary of the advantages and disadvantages of model-free and model-based methods in imitation learning.

*Model-free* imitation learning methods attempt to learn a policy that reproduce the behavior demonstrated by experts without learning/using a forward model of the system. Therefore, there is no need to estimate the system dynamics in model-free imitation learning method. Yet, the system dynamics is encoded only implicitly in policies learned by model-free methods. In many robotic systems, especially in industrial applications, position/velocity controllers are often available for controlling joints. In such cases, we can assume that the robot is fully actuated, and the dynamics of the system is almost negligible in motion planning if a reasonably smooth trajectory is used. Model-free imitation learning methods can be easily applied to motion planning for such (nearly) fully-actuated robotic systems when the demonstrations by ex-perts are available. For this reason, behavioral cloning methods which learn a direct mapping from states/contexts to actions have focused on model-free methods until recent years.

For motion planning of underactuated systems, it is often necessary to plan a feasible trajectory by considering the system dynamics. It can be challenging to use model-free BC methods to learn trajectories in such underactuated systems where the reachable states are limited. However, recent IRL work by Boularias et al. [2011], Finn et al. [2016b], Ho and Ermon [2016] shows how one can learn skills in underactuated systems through iterative rollouts without explicitly learning a dynamics model.

*Model-based* imitation learning methods attempt to learn a policy that reproduces the demonstrated behavior by learning/using the system dynamics, e.g. a forward model of the system. This property can be critical especially for underactuated robots. Since underactuation limits the number of reachable states, it is essential to take into account the dynamics of the system when planning feasible trajectories. Moreover, the prior knowledge of the system dynamics makes inverse reinforcement learning easier since the learner's performance can be easily predicted when the system dynamics is known. However, in a real robotic system, it is often challenging to learn the system dynamics. For example, it is hard to model the contact between deformable objects, and it will be difficult to apply model-based methods to tasks that involve such contacts.

**Table 2.1:** Advantages and disadvantages of model-based and model-free methods in imitation learning. Model-free methods learn a policy without knowledge on the system dynamics, and the system dynamics is encoded only implicitly in policies. Model-based methods learn a policy that explicitly satisfies the system dynamics by leveraging the system dynamics. However, learning/estimating the system dynamics can be challenging.

|                   | **Model-free**                                                                                                  | **Model-based**                                                                           |
| ----------------- | --------------------------------------------------------------------------------------------------------------- | ----------------------------------------------------------------------------------------- |
| **Advantages**    | A policy can be learned without learning/estimating the system dynamics.                                        | The learning process can be data-efficient. A learned policy satisfies the system dynamics. |
| **Disadvantages** | The prediction of future states is difficult. The system dynamics is only implicitly considered in the resulting policy. | Model learning can be difficult. Computationally expensive.                               |

Existing imitation learning methods can be categorized into behavioral cloning and inverse reinforcement learning with a distinction between model-free and model-based methods as shown in Table 2.2. At a glance, one can see that studies on behavioral cloning have focused on model-free methods and studies on inverse reinforcement learning have focused on model-based methods, although recent studies on IRL have proposed model-free methods. BC methods have been mainly focused on trajectory planning for robotic systems in which a lower-level controller is available. A model-free approach is a reasonable choice in such applications because the dynamics of the system is not crucial. On the other hand, IRL has focused on learning a policy in action-state space which needs to be iteratively evaluated in a given system. A model-based approach is suitable for such applications, and this is why many model-based methods have been developed for IRL.

**Table 2.2:** Categorization of existing imitation learning methods with distinction between model-free and model-based methods. Model-free methods are dominant in behavioral cloning, and model-based methods are dominant in inverse reinforcement learning. Recent studies on IRL have proposed model-free methods.

|  | **Model-free** | **Model-based** |
|---|---|---|
| **Behavioral Cloning** | Widrow and Smith [1964], Chambers and Michie [1969], Pomerleau [1988], Schaal et al. [2004], Schaal [1999], Ijspeert et al. [2013], Calinon et al. [2007], Khansari-Zadeh and Billard [2011], Paraschos et al. [2013], Osa et al. [2014], Ross and Bagnell [2010], Ross et al. [2011], Takano and Nakamura [2015], Maeda et al. [2016], Deniša et al. [2016], Ho and Ermon [2016] | Ude et al. [2004], Englert et al. [2013], van den Berg et al. [2010] |
| **Inverse Reinforcement Learning** | Boularias et al. [2011] Kalakrishnan et al. [2013] | Abbeel and Ng [2004], Ratliff et al. [2006b], Silver et al. [2010], Ziebart et al. [2008], Ziebart [2010], Levine et al. [2011], Levine and Koltun [2012], Hadfield-Menell et al. [2016], Finn et al. [2016b] |

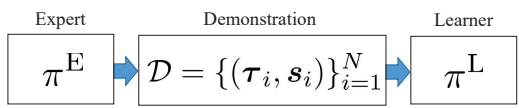

**Figure 2.2:** Diagram of general imitation learning. The learner cannot directly observe the expert's policy in many problems. Instead, a set of trajectories induced by the expert's policy is available in imitation learning. The learner estimates the policy that reproduces the expert's behavior using the given demonstrations. Please note that the process of querying the demonstration and updating the learner's policy can be interactive.

## 2.4 Observability

The main goal of many imitation learning methods is to learn a policy that reproduces the expert's behavior. Since the expert's policy cannot be directly observed, the learner recovers the policy from the expert's demonstrations. The diagram in Figure 2.2 illustrates the imitation learning process. To formulate a imitation learning problem, it is necessary to consider the observability in practice.

For a formal definition, it is necessary to figure out observability of the state. Observability can vary significantly between different applications leading to different kinds of learning methods.

### 2.4.1 Trajectories in Fully Observable Settings

When the state of the system is fully observable, we can obtain a trajectory as a sequence of the state and the control input as

$$\boldsymbol{\tau} = [\boldsymbol{x}_0, \boldsymbol{u}_0, \boldsymbol{x}_1, \boldsymbol{u}_1, \ldots, \boldsymbol{x}_T, \boldsymbol{u}_T]. \tag{2.3}$$

For instance, both the state and the control inputs are observable in a teleoperated system in [Abbeel et al., 2010, van den Berg et al., 2010, Osa et al., 2014, Ross et al., 2011], although observation can be noisy.

### 2.4.2 Trajectories in Partially Observable Settings

In some settings of imitation learning, the control input by the experts is not observable in demonstrations, and only the states of the system during the demonstrations are given. In such cases, the trajectory is given as a sequence of the state of the system,

$$\boldsymbol{\tau} = [\boldsymbol{x}_0, \boldsymbol{x}_1, \ldots, \boldsymbol{x}_T]. \tag{2.4}$$

For example, the control inputs to achieve the demonstrated trajectory are often unobservable in kinesthetic teaching [Kober and Peters, 2009, Englert et al., 2013, Maeda et al., 2016]. Also, when transferring motions captured from a human expert to a humanoid robot the control inputs to achieve the desired motion in the learner's embodiment cannot be observed [Ijspeert et al., 2002b, Grimes et al., 2006b, Grimes and Rao, 2009]. In addition, the state of the system is often partially observable. In this case, the trajectory is given as a sequence of the partial observation of the system,

$$\boldsymbol{\tau} = [\boldsymbol{y}_0, \boldsymbol{y}_1, \ldots, \boldsymbol{y}_T]. \tag{2.5}$$

where $\boldsymbol{y}_t$ is the partial observation of the system, which is often given by $\boldsymbol{y}_t = \boldsymbol{f}_o(\boldsymbol{x}_t)$ where $\boldsymbol{f}_o$ is the observation function. As a special case, the observation $\boldsymbol{y}$ can be linear w.r.t. the state $\boldsymbol{x}$ as $\boldsymbol{y}_t = \boldsymbol{H}_t \boldsymbol{x}_t$ where $\boldsymbol{H}_t$ is the observation matrix.

### 2.4.3 Differences in observability between the expert and the learner

In imitation learning, the expert and the learner often observe the environment differently. For example, in robotic manipulation tasks a human expert often obtains much richer sensory information compared to a robot learner due to the differences in their sensory embodiments. As another example, a robotic learner may be able to record sensory information more accurately and at a higher rate than a human expert. In such cases, the information of the learner about the environment/system differs from the information of the expert and should be taken into account when formalizing the imitation learning problem. In general, the observability of the expert and learner can manifest in different ways:

- The expert observes partially
    - the system state
    - the control inputs by the expert
    - learner's observations
- The learner observes partially
    - the system state
    - expert's observations
    - the control inputs by the expert
    - the control inputs by the learner

These cases need to be taken into account when deciding on the imitation learning approach for a specific application. When the expert observes the system state partially, the expert demonstrations can become sub-optimal requiring careful consideration. Moreover, when the expert observes the learner, the learner may have more information about its own embodiment. For example, if a human expert uses kinesthetic teaching to show how to grasp an object, the demonstration may be sub-optimal for a robot learner if the expert does not see what the robot observes.

In imitation learning, the expert is often assumed to behave optimally. However, this optimality is often based on partial observations which may differ significantly from the observations of the learner. For example, if the human expert performs a motion which goes around an obstacle which the robot learner does not observe, a robot learner learns to perform a similar circumnavigation motion even when there are no obstacles. Moreover, when the learner observes only partially expert observations the learner can make wrong predictions about the policy behind expert behavior.

## 2.5 Policy Representation in Imitation Learning

One of the important design choices in imitation learning is policy representation. In this section, we discuss the design choices related to policy representation.

### 2.5.1 Levels of Policy Abstraction

For imitation learning, several types of policy abstractions can be used. We can categorize the policy representations into three types: 1) symbolic-level abstraction, 2) trajectory-level abstraction, and 3) action-state space abstraction. In task level planning, the learner learns a policy that generates an option $o \in \mathcal{O}$ where $\mathcal{O}$ is the set of options. *Options* are often defined as policies of taking actions over a period of time [Sutton et al., 1999]. In this task-level planning, each option often consists of a set of actions or trajectories. A policy maps given states $x_t$ and contexts $s$ to sequences of options in the task-level abstraction.

$$\pi : x_t, s \mapsto [o_1, \ldots, o_T], \tag{2.6}$$

where $T$ is the horizon of the task. A complex task is often hard to model as a single movement. The task-level abstraction enables modeling such complex task as a sequence of simple movements. BC methods such as [Konidaris et al., 2011, Niekum et al., 2014, Kroemer et al., 2015] model complex task as a sequence of movement primitives.

In trajectory planning, a policy maps a context $s$ to a trajectory $\tau$ that is a sequence of the state of the system $x$ (and control inputs $u$)

as

$$\pi : s \mapsto \tau. \tag{2.7}$$

BC methods such as DMP [Schaal et al., 2004, Ijspeert et al., 2013] and ProMP [Paraschos et al., 2013, Maeda et al., 2016] learn such trajectory-based policies.

In the action-state space level, a policy maps states of the system $x_t$ and contexts $s$ to control inputs $u_t$ as

$$\pi : x_t, s \mapsto u_t. \tag{2.8}$$

BC methods such as [Chambers and Michie, 1969, Pomerleau, 1988, Khansari-Zadeh and Billard, 2011, Ross et al., 2011] and IRL methods such as [Abbeel and Ng, 2004, Ziebart et al., 2008, Boularias et al., 2011, Finn et al., 2016b] learn policies in action-state space. These abstractions are summarized in Table 2.3.

Existing imitation learning methods can be categorized based on task abstractions as shown in Table 2.4. The table displays an abundance of model-free methods for trajectory learning. On the contrary, many model-based IRL methods have been developed with action-space space abstractions. Since commercially available robotic manipulators often have a position/velocity controller, model-free methods are preferred for trajectory planning in such systems. This is especially pronounced in motion planning methods designed for robotic manipulators in the robotics research community. On the other hand, the machine learning community have developed many IRL methods for learning a policy in action-state space.

**Table 2.3:** Abstraction and the related policy in imitation learning. In a task-level abstraction, the policy maps from the initial state $x_0$ to a sequence of discrete options, where an option at time step $t$ is denoted with $o_t$. In a trajectory-level abstraction, the policy maps from an initial state $x_0$ to a trajectory $\tau$. In an action-state space abstraction, the policy maps from the current state $x_t$ to a control $u_t$.

| Abstraction Level | Policy |
| --- | --- |
| Task-level abstraction | $\pi : x, s \mapsto [o_1, \ldots, o_T]$ |
| Trajectory-based abstraction | $\pi : x_0, s \mapsto \tau$ |
| Action-state space abstraction | $\pi_t : x_t, s \mapsto u_t$ |

### 2.5.2 Hierarchical vs Monolithic Policies

When we consider a single abstraction level of policy, the policy will be non-hierarchical/monolithic. BC methods such as [Chambers and Michie, 1969, Pomerleau, 1988, Schaal et al., 2004, Khansari-Zadeh and Billard, 2011, Paraschos et al., 2013, Ross et al., 2011] and IRL methods such as [Abbeel and Ng, 2004, Ratliff et al., 2006b, Ziebart, 2010, Finn et al., 2016b] are monolithic. Thus far, numerous methods have been developed for learning a monolithic policy. However, we need to employ a complex policy representation such as a neural network

Table 2.4: Categorization of imitation learning methods based on different policy abstractions with distinction between model-free and model-based methods. Many model-free methods have been developed for imitation learning with trajectory-based abstractions. On the contrary, many model-based IRL methods have been developed with action-space space abstractions.

|  | **Model-free** | **Model-based** |
|---|---|---|
| **Task-level abstraction** | Takano and Nakamura [2015], Niekum et al. [2014], Konidaris et al. [2014], Inamura et al. [2004] | - |
| **Trajectory-based abstraction** | Schaal et al. [2004], Schaal [1999], Ijspeert et al. [2013], Calinon et al. [2007], Khansari-Zadeh and Billard [2011], Paraschos et al. [2013], Osa et al. [2014], Maeda et al. [2016], Deniša et al. [2016] | Ude et al. [2004], Englert et al. [2013], van den Berg et al. [2010], Abbeel et al. [2010] |
| **Action-state space abstraction** | Chambers and Michie [1969], Widrow and Smith [1964], Pomerleau [1988], Ross and Bagnell [2010], Ross et al. [2011], Boularias et al. [2011], Kalakrishnan et al. [2013], Ho and Ermon [2016] | Abbeel and Ng [2004], Ratliff et al. [2006b], Silver et al. [2010], Ziebart et al. [2008], Ziebart [2010], Levine et al. [2011], Levine and Koltun [2012], Hadfield-Menell et al. [2016], Finn et al. [2016b] |

policy in [Finn et al., 2016b] in order to learn a complex task with a monolithic policy.

On the contrary, by combining the different levels of abstraction, we can learn a hierarchical policy where the lower-level policies learn to perform the primitive behavior and the upper-level policy learns to plan a sequence of the lower-level policies. BC methods such as [Niekum et al., 2014, Konidaris et al., 2014, Kroemer et al., 2015] and IRL methods such as [Kolter et al., 2008, Choi and Kim, 2015, Krishnan et al., 2016] learn hierarchical policies. Since a hierarchical policy can be decomposed into a sequence of the lower-level policies, we do not have to use complex policy representation for the lower-level policies. On the other hand, it is not trivial to learn all of the lower-level and upper-level policies simultaneously.

### 2.5.3  Feedback vs Open-Loop/Feedback-Free Policies

With regard to feedback of the state, policies can be categorized into two types: feedback and open-loop/feedback-free policies. A *feedback* policy iteratively determines the control input/desired behavior based on the feedback from the environment. In other words, a feedback policy considers the changes of the environment caused by the previous control input in sequential decision making. A policy for determining the torque input to a robotic manipulator is often learned in robotic applications such as [Boularias et al., 2011, Englert et al., 2013]. Such a torque-based control is often learned as a feedback policy since it is essential to consider the state of the system in sequential decision making where a small mistake can cause a big error in the next state.

In contrast, an *open-loop/feedback-free* policy determines the control input/desired behavior just based on the initial input. Therefore, once a open-loop policy starts running, it does not use the feedback from the environment. A policy for planning a desired trajectory can be often learned as an open-loop policy since it can be addressed as a one shot decision making for a given situation such as in [Calinon et al., 2007, Takano and Nakamura, 2015]. However, there are methods for planning and updating the desired trajectory during the task execution such as [Ijspeert et al., 2013, Paraschos et al., 2013, Schulman et al., 2013, Osa et al., 2017b]. For example, in the framework of [Schulman et al., 2013], the trajectory is learned as a direct function of the pixel values observed, and the desired trajectory is updated online.

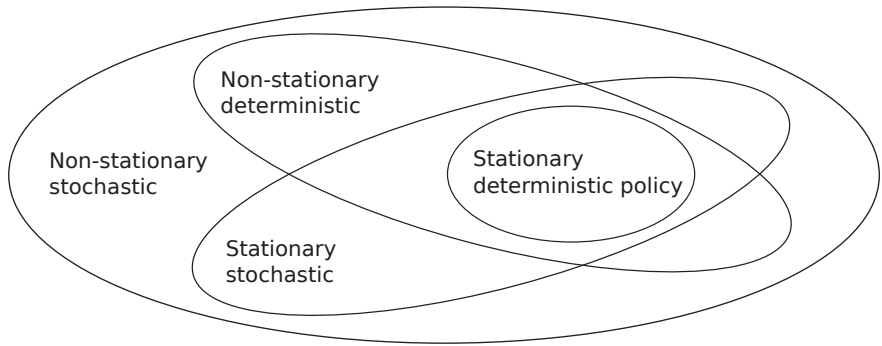

**Figure 2.3:** Illustration of the relationships between basic policy classes. Stationarity is a special case of non-stationarity and determinism is a special case of stochasticity. We use the terms "stationary" and "time-invariant" interchangeably. Likewise, "non-stationary" and "time-variant" are used interchangeably. Please see § 2.5.4 for more details.

Different policy types can be used in the same system at the different level. In the acrobatic helicopter flights by Abbeel et al. [2010], the scheme for planning the desired trajectory can be interpreted as an open-loop policy because the system does not update the desired trajectory during the flight. Meanwhile, an iterative LQR controller for the lower-level control in [Abbeel et al., 2010] can be considered as a feedback policy because it determines the control input based on the observation of the system.

### 2.5.4 Stationarity and Stochasticity of Policies

With respect to stationarity, we can categorize policies into stationary and non-stationary policies, depending on whether the policy depends on time. Moreover, we can categorize policies into deterministic and stochastic policies in terms of stochasticity. Note that stationarity is a special case of non-stationarity and determinism is a special case of stochasticity. Figure 2.3 illustrates relationships between these policy classes.

#### 2.5.4.1 Stationary vs. Non-Stationary Policies

A non-stationary (time-variant) policy depends on time. Typically trajectory based policies are non-stationary since the policy depends on the time step or phase of the trajectory. For example, a complex movement of a robot arm through space [van den Berg et al., 2010, Osa et al., 2014] needs to be performed such that the learned speed is similar to the demonstrated speed over the whole trajectory, which often requires a non-stationary policy. A stationary (time-invariant) policy depends only on the current state of the system. Stationary policies are typically used in applications where data from different time steps can be similar. For example, in a racing car simulation [Abbeel and Ng, 2004, Ross et al., 2011], steering right when about to drive left off the road is a useful action independent of the time this occurs. In another instance, simple motion for approaching an object can be also learned as a stationary policy [Khansari-Zadeh and Billard, 2011].

#### 2.5.4.2 Deterministic Policy

A deterministic policy for trajectory planning determines a unique trajectory $\boldsymbol{\tau}$ for a given initial state $\boldsymbol{x}_0$ and/or context $\boldsymbol{s}$ as

$$\boldsymbol{\tau} = \boldsymbol{\pi}(\boldsymbol{x}_0, \boldsymbol{s}). \tag{2.9}$$

Behavior cloning methods such as dynamic movement primitives [Ijspeert et al., 2013, Schaal et al., 2004] can be interpreted as deterministic policy representations for trajectory planning.

A deterministic policy in action-state space determines a unique control input $\boldsymbol{u}$ for a given state $\boldsymbol{x}$ and/or context $\boldsymbol{s}$ as

$$\boldsymbol{u} = \boldsymbol{\pi}(\boldsymbol{x}, \boldsymbol{s}). \tag{2.10}$$

In this case, $\boldsymbol{\pi}$ represents a deterministic function of $\boldsymbol{x}$. When a deterministic policy is used and both states and actions are fully observable, the distribution of the trajectory $\boldsymbol{\tau} = [\boldsymbol{x}_0, \boldsymbol{u}_0, \ldots, \boldsymbol{x}_T, \boldsymbol{u}_T]$ is given as

$$p(\boldsymbol{\tau}) = p(\boldsymbol{x}_0) \prod_{t=1}^{T} p(\boldsymbol{x}_{t+1} | \boldsymbol{x}_t, \pi_t(\boldsymbol{x}_t)). \tag{2.11}$$

Commonly, in non-adversarial sequential decision making problems, such as Markov decision processes, the optimal policy for accomplishing the objective for a given model is deterministic. Inverse reinforcement learning methods such as MMP [Ratliff et al., 2006b] and LEARCH [Ratliff et al., 2009, Zucker et al., 2011] employ a deterministic policy derived from a reward/cost function recovered from demonstrations. Behavior cloning methods such as [Pomerleau, 1988, Khansari-Zadeh and Billard, 2011] also employ deterministic policies.

### 2.5.4.3 Stochastic Policy

A stochastic policy in action-state space draws a control input $\boldsymbol{u}$ according to a probability distribution for a given state $\boldsymbol{x}$ and/or context $\boldsymbol{s}$ as

$$\boldsymbol{u} \sim \pi(\boldsymbol{u}|\boldsymbol{x}, \boldsymbol{s}). \tag{2.12}$$

In this case, $\pi$ represents a conditional distribution of the control input $\boldsymbol{u}$ given $\boldsymbol{x}$ and $\boldsymbol{s}$. If the probability distribution is given as a delta function, the policy is deterministic. When a stochastic policy is used and both states and actions are fully observable, the distribution of the trajectory $\boldsymbol{\tau} = [\boldsymbol{x}_0, \boldsymbol{u}_0, \ldots, \boldsymbol{x}_T, \boldsymbol{u}_T]$ is given as

$$p(\boldsymbol{\tau}) = p(\boldsymbol{x}_0) \prod_{t=1}^{T} p(\boldsymbol{x}_{t+1}|\boldsymbol{x}_t, \boldsymbol{u}_t) \pi(\boldsymbol{u}_t|\boldsymbol{x}_t). \tag{2.13}$$

A stochastic policy is useful to model the stochastic behavior of the expert. Inverse reinforcement learning methods such as [Ziebart et al., 2008, Boularias et al., 2011] employ a stochastic policy to represent such stochastic behavior. Stochastic policies introduce uncertainty, which is useful for exploring the parameter space of the policy in iterative methods. Model-based behavior cloning methods such as [Englert et al., 2013] and inverse reinforcement learning methods such as [Finn et al., 2016b] employ a stochastic policy and learn system dynamics through iterative learning.

## 2.6 Behavior Descriptors

In imitation learning, it is essential to quantify the behavior and measure the difference between the expert's behavior and the learner's behavior. For this purpose, we need to consider "*what should be matched between the expert and the learner?*" In the following, we list descriptors of behavior, which can be matched between the expert and the learner in imitation learning.

### 2.6.1 State-action Distribution

Given a dataset $\mathcal{D} = \{(\boldsymbol{x}, \boldsymbol{u})\}$ that consists of state-control input pairs, we can model the joint distribution of the state and the control input $p(\boldsymbol{x}, \boldsymbol{u})$ or the conditional distribution of the control input given the state $p(\boldsymbol{u}|\boldsymbol{x})$. Early imitation learning approaches [Chambers and Michie, 1969, Widrow and Smith, 1964, Pomerleau, 1988] learned a policy by modeling state-action distributions using supervised learning methods. However, since a state-action distribution only describes the short term behavior, matching only the state-action distribution can lead to a mismatch with long term behavior.

### 2.6.2 Trajectory Feature Expectation

To match the behavior between the expert and the learner over a long horizon, it is necessary to consider trajectory features. Since both a trajectory itself and observations of it are often stochastic and noisy, the expectation of trajectory features (an expectation has less noise compared to individual instances) is often used to describe the behavior of the expert and the learner. The expectation of the trajectory features with respect to the learner's policy is given by

$$\mathbb{E}_{p(\boldsymbol{\tau})}[\boldsymbol{\phi}(\boldsymbol{\tau})] = \int p(\boldsymbol{\tau})\boldsymbol{\phi}(\boldsymbol{\tau})d\boldsymbol{\tau}, \qquad (2.14)$$

where $p(\boldsymbol{\tau})$ is the trajectory distribution induced by the learner's policy and $\boldsymbol{\phi}(\boldsymbol{\tau})$ is the feature vector of the trajectory $\boldsymbol{\tau}$. The expectation of the trajectory $\mathbb{E}[\boldsymbol{\tau}]$ can be interpreted as a special case of the feature expectation. When a dataset of trajectories $\mathcal{D} = \{\boldsymbol{\tau}_i^{\text{demo}}\}_{i=1}^N$ is available, the expectation of the trajectory feature can be approximated

as

$$\mathbb{E}_{p(\boldsymbol{\tau})}[\boldsymbol{\phi}(\boldsymbol{\tau})] \simeq \frac{1}{N} \sum_{i=1}^{N} \boldsymbol{\phi}(\tau_i^{\text{demo}}). \tag{2.15}$$

Feature expectation matching appears both in behavior cloning [Ijspeert et al., 2002a, Osa et al., 2014] and inverse reinforcement learning [Abbeel and Ng, 2004, Ratliff et al., 2006b, Ziebart et al., 2008].

### 2.6.3 Trajectory Feature Distribution

A distribution over trajectory features $p(\boldsymbol{\phi}(\boldsymbol{\tau}))$ is also often used for matching the behavior between the expert and the learner. We can match not only the first order moment (mean) of the distribution but also higher order moments. The trajectory distribution $p(\boldsymbol{\tau})$ can be considered as a special case of the feature distribution. Behavior cloning methods such as [Paraschos et al., 2013, Englert et al., 2013] and inverse reinforcement learning methods such as [Arenz et al., 2016] use feature distributions.

## 2.7 Information Theoretic Understanding of Feature Matching in Imitation Learning

As we discussed in § 1.6, imitation learning can be formulated as a problem of finding a policy that minimizes the difference between demonstrated and learned behavior. For this purpose, many imitation learning methods perform a "projection" of demonstrated behavior into a parameterized policy space. Projecting demonstrations onto a manifold of a parameterized policy requires considering the relationship between the distribution of the demonstrations and the distribution of the parameterized policy. Information theory provides a principled way of assessing this relationship.

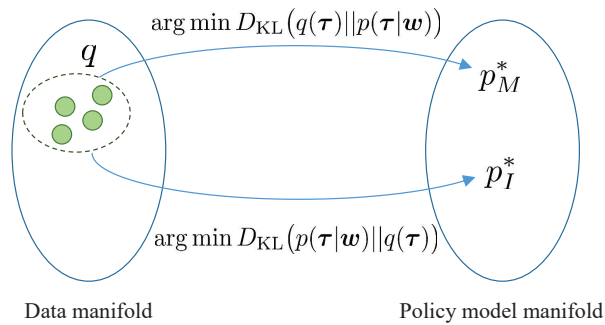

**Figure 2.4:** Illustration of M- and I- projections from the data manifold onto the policy model manifold. The solutions of M- and I- projections are different since the KL divergence is not symmetric.

### 2.7.1 Information Theoretic Understanding of Imitation Learning Algorithms for Trajectory Learning

Consider a trajectory distribution $p(\boldsymbol{\tau}|\boldsymbol{w})$ induced by a policy $\pi$ with a parameter vector $\boldsymbol{w}$. Supervised learning methods often obtain a solution based on the maximum likelihood of the given training data. As is well known Bishop [2006], maximizing the (causal) likelihood is equivalent to minimizing the KL divergence

$$D_{\mathrm{KL}}\left(q(\boldsymbol{\tau})||p(\boldsymbol{\tau}|\boldsymbol{w})\right) = \int q(\boldsymbol{\tau}) \ln \frac{q(\boldsymbol{\tau})}{p(\boldsymbol{\tau}|\boldsymbol{w})} d\boldsymbol{\tau}, \qquad (2.16)$$

where $q(\boldsymbol{\tau})$ is the empirical distribution over trajectories induced by the expert's policy and $\boldsymbol{\tau}$ is a trajectory. This equation can be interpreted as a projection from the data manifold to the policy model manifold [Amari, 2016]. On the other hand, as the KL divergence is not symmetric, minimizing $D_{\mathrm{KL}}\left(q(\boldsymbol{\tau})||p(\boldsymbol{\tau}|\boldsymbol{w})\right)$ is not equivalent to minimizing

$$D_{\mathrm{KL}}\left(p(\boldsymbol{\tau}|\boldsymbol{w})||q(\boldsymbol{\tau})\right) = \int p(\boldsymbol{\tau}|\boldsymbol{w}) \ln \frac{p(\boldsymbol{\tau}|\boldsymbol{w})}{q(\boldsymbol{\tau})} d\boldsymbol{\tau}, \qquad (2.17)$$

which represents the projection from the policy model manifold to the data manifold. We discuss a few more details of minimizing the different projections in the following.

First, we consider imitation learning methods for trajectory learning based on the M-projection defined in (1.4). The goal of imitation learning in this case is to learn a parameter vector $\boldsymbol{w}$, such that the M-projection is minimized, i.e.,

$$\boldsymbol{w}^* = \arg\min_{\boldsymbol{w}} D_{\mathrm{KL}}\left(q(\boldsymbol{\tau})||p(\boldsymbol{\tau}|\boldsymbol{w})\right). \tag{2.18}$$

The resulting objective function can also be written as

$$\mathcal{L}_{\mathrm{M}} = D_{\mathrm{KL}}\left(q(\boldsymbol{\tau})||p(\boldsymbol{\tau}|\boldsymbol{w})\right) = \int q(\boldsymbol{\tau}) \ln \frac{q(\boldsymbol{\tau})}{p(\boldsymbol{\tau}|\boldsymbol{w})} \mathrm{d}\boldsymbol{\tau} \tag{2.19}$$

$$= \mathbb{E}_q[\ln q(\boldsymbol{\tau})] - \mathbb{E}_q[\ln p(\boldsymbol{\tau}|\boldsymbol{w})], \tag{2.20}$$

where $\mathbb{E}_q[\cdot]$ is the expectation with respect to $q(\boldsymbol{\tau})$ [Bishop, 2006, Sugiyama, 2015]. The expectations $\mathbb{E}_q[\cdot]$ in (2.20) can be estimated using the demonstrated trajectories drawn from $q(\boldsymbol{\tau})$. Since the first term in (2.20) is independent from $\boldsymbol{w}$, $D_{\mathrm{KL}}\left(q(\boldsymbol{\tau})||p(\boldsymbol{\tau}|\boldsymbol{w})\right)$ can be minimized by maximizing the expected log likelihood $\mathbb{E}_q[\ln p(\boldsymbol{\tau}|\boldsymbol{w})]$. Hence, imitation learning based on simple supervised learning can be seen as a special case of computing the M-projection as these algorithms essentially perform a likelihood maximization. Examples of such algorithms are the least square solution for DMPs, expectation maximization (EM) for ProMPs, and EM for SEDS, which minimize $D_{\mathrm{KL}}\left(q(\boldsymbol{\tau})||p(\boldsymbol{\tau}|\boldsymbol{w})\right)$ with different parameterizations [Ijspeert et al., 2013, Paraschos et al., 2013, Khansari-Zadeh and Billard, 2011].

It is informative to note that there is a close relation between the maximum likelihood solution and the solution obtained from the principle of maximum entropy. Consider, for instance, average feature constraints

$$\mathbb{E}_p[\boldsymbol{\phi}(\boldsymbol{\tau})] = \boldsymbol{a}. \tag{2.21}$$

If we chose subject to the feature matching constraint the distribution that results in maximum entropy, we cover the exponential family parametrization of $p(\boldsymbol{\tau}|\boldsymbol{w})$ Amari [2016]:

$$p(\boldsymbol{\tau}|\boldsymbol{w}) = \frac{\exp\left(\boldsymbol{w}^\top \boldsymbol{\phi}(\boldsymbol{\tau})\right)}{Z}. \tag{2.22}$$

Substituting the resulting form $p(\boldsymbol{\tau}|\boldsymbol{w})$ with (2.22) into the original maximum entropy problem ignoring terms which do not depend on the parameters $\boldsymbol{w}$, the resulting dual objective function (or equivalently the one in (2.20)) yields

$$\mathcal{L}_{\mathrm{M}} = \mathbb{E}_q[\boldsymbol{w}^\top \boldsymbol{\phi}(\boldsymbol{\tau})] - \ln Z. \tag{2.23}$$

Differentiating (2.23) w.r.t. $\boldsymbol{w}$ yields the following gradient which can be used for optimization of the parameters:

$$\frac{d\mathcal{L}_{\mathrm{M}}}{d\boldsymbol{w}} = \mathbb{E}_q[\boldsymbol{\phi}(\boldsymbol{\tau})] - 1/Z \int \exp\left(\boldsymbol{w}^\top \boldsymbol{\phi}(\boldsymbol{\tau})\right) \boldsymbol{\phi}(\boldsymbol{\tau}) d\boldsymbol{\tau} \tag{2.24}$$

$$= \mathbb{E}_q[\boldsymbol{\phi}(\boldsymbol{\tau})] - \mathbb{E}_p[\boldsymbol{\phi}(\boldsymbol{\tau})]. \tag{2.25}$$

Note that setting the gradient to 0 in order to obtain the optimum yields the optimality condition required to hold in the primal, that is that feature expectations match:

$$\mathbb{E}_p[\boldsymbol{\phi}(\boldsymbol{\tau})] = \mathbb{E}_q[\boldsymbol{\phi}(\boldsymbol{\tau})]. \tag{2.26}$$

From (2.26), we can conclude that maximum likelihood on an assumed exponential family form is also a solution to finding the maximum entropy distribution (2.22) which respects the average feature constraint (2.26). The latter viewpoint allows us to reason about, for instance, cost function matching in Inverse Reinforcement Learning and to automatically construct an appropriate form for policies.

This observation is called the maximum likelihood / maximum entropy duality Dudík and Schapire [2006]. Furthermore, as the M-projection yields the same solution as maximizing the likelihood, we can conclude that the M-projection solution for an exponential family of trajectory distributions is equivalent to the maximum entropy one.

It is often useful to consider the maximum entropy principle in its regularized form [Ziebart et al., 2013] [Boularias et al., 2011], that is, instead of finding a maximum entropy distribution we want to find a distribution with the minimal KL divergence relative to a "prior" distribution $p_0(\boldsymbol{\tau})$ while matching the features of the demonstrator, that is,

$$\arg\min_w D_{\mathrm{KL}}\left(p(\boldsymbol{\tau}) || p_0(\boldsymbol{\tau})\right) \tag{2.27}$$

$$\text{s.t.:} \ \mathbb{E}_p[\boldsymbol{\phi}(\boldsymbol{\tau})] = \mathbb{E}_q[\boldsymbol{\phi}(\boldsymbol{\tau})]. \tag{2.28}$$

The solution to this problem can again be obtained by the method of Lagrangian multipliers

$$p(\boldsymbol{\tau}|\boldsymbol{w}) = \frac{p_0(\boldsymbol{\tau})\exp\left(\boldsymbol{w}^\top \boldsymbol{\phi}(\boldsymbol{\tau})\right)}{Z} \qquad (2.29)$$

with $p_0(\boldsymbol{\tau}) = \exp\left(\boldsymbol{w}_0^\top \boldsymbol{\phi}(\boldsymbol{\tau})\right)/Z_0$.

A particularly elegant result due to [Dudík and Schapire, 2006] demonstrates that if we have bounds on the accuracy with which our feature matching constraints hold, the resulting maximum entropy problem gives rise via duality to a regularized likelihood equivalent to a maximum a-posterior estimate with a prior on the dual parameters.

It is, however, important to note that such maximum entropy principles should not to be confused with the I-projection, which computes

$$\arg\min_{\boldsymbol{w}} D_{\mathrm{KL}}\left(p(\boldsymbol{\tau}|\boldsymbol{w})||q(\boldsymbol{\tau})\right).$$

Here, the data is induced via the distribution $q(\boldsymbol{\tau})$ on the right-hand side of the KL, while in the maximum entropy principle, the data is induced by the feature averages and $p_0(\boldsymbol{\tau})$ on the right-hand side of the KL is just a prior. The I-projection does not match features of the demonstrator. Whenever an algorithm matches average features, it is an instance of an M-projection based algorithm. Since $\ln q(\boldsymbol{\tau})$ is unknown and hard to evaluate in practice, it is challenging to perform the I-projection in the context of imitation learning. To the best of our knowledge, there is no existing imitation learning method that performs the I-projection exactly.

As we have seen from our discussion above, many imitation learning methods can be seen as related to the M-projection and to the principle of maximum entropy. This is true for most model-free and model-based methods. Model-free methods based on standard supervised learning [Ijspeert et al., 2013, Khansari-Zadeh and Billard, 2011] do not require access to the system dynamics or iterative data acquisition.

In contrast, model-based imitation learning methods often try to match features of the state distribution so as to satisfy $\mathbb{E}_p[\boldsymbol{\phi}(\boldsymbol{\tau})] = \mathbb{E}_q[\boldsymbol{\phi}(\boldsymbol{\tau})]$. In order to do so, we either need access to the system dynamics [Ziebart et al., 2008, Ziebart, 2010] or require iterative data acquisition [Boularias et al., 2011].

### 2.7.2 Information Theoretic Understanding of Imitation Learning Algorithms in Action-State Space

In this section, we have a look at imitation learning in action-state space from an information theoretic point of view. In a Markov model, the probability distribution over trajectories $p(\boldsymbol{\tau})$ can be decomposed as a sequence of states and actions

$$p(\boldsymbol{\tau}) = p(\boldsymbol{x}_0) \prod_{t=0}^{T} p(\boldsymbol{x}_{t+1}|\boldsymbol{x}_t, \boldsymbol{u}_t)\pi(\boldsymbol{u}_t|\boldsymbol{x}_t) , \qquad (2.30)$$

where the policy $\pi(\boldsymbol{u}_t|\boldsymbol{x}_t)$ maps from the states of the system to the control inputs. Let us consider the trajectory distribution $p(\boldsymbol{\tau})$ induced by the learner's policy and the trajectory distribution $q(\boldsymbol{\tau})$ induced by the expert's policy. If the embodiments of the learner and the expert are equivalent and stationary, that is, $q(\boldsymbol{x}_{t+1}|\boldsymbol{x}_t, \boldsymbol{u}_t) = p(\boldsymbol{x}_{t+1}|\boldsymbol{x}_t, \boldsymbol{u}_t) = p(\boldsymbol{x}_t|\boldsymbol{x}_{t-1}, \boldsymbol{u}_{t-1})$, the relation of $p(\boldsymbol{\tau})$ and $q(\boldsymbol{\tau})$ is given by

$$\frac{p(\boldsymbol{\tau})}{q(\boldsymbol{\tau})} = \frac{\prod_{t=0}^{T} \pi^{\mathrm{L}}(\boldsymbol{u}_t|\boldsymbol{x}_t)}{\prod_{t=0}^{T} \pi^{\mathrm{E}}(\boldsymbol{u}_t|\boldsymbol{x}_t)}, \qquad (2.31)$$

where $\pi^{\mathrm{L}}$ is the learner's policy and $\pi^{\mathrm{E}}$ is the expert's policy. In this case, imitation learning methods based on the M-projection minimize

$$D_{\mathrm{KL}}(q(\boldsymbol{\tau})||p(\boldsymbol{\tau})) = \int q(\boldsymbol{\tau}) \sum_{t=0}^{T} \ln \frac{\pi^{\mathrm{E}}(\boldsymbol{u}_t|\boldsymbol{x}_t)}{\pi^{\mathrm{L}}(\boldsymbol{u}_t|\boldsymbol{x}_t)} \mathrm{d}\boldsymbol{\tau} \qquad (2.32)$$

$$= \int q(\boldsymbol{x}, \boldsymbol{u}) \ln \frac{\pi^{\mathrm{E}}(\boldsymbol{u}|\boldsymbol{x})}{\pi^{\mathrm{L}}(\boldsymbol{u}|\boldsymbol{x})} \mathrm{d}\boldsymbol{x}\mathrm{d}\boldsymbol{u} \qquad (2.33)$$

$$= \mathbb{E}_{q(\boldsymbol{x},\boldsymbol{u})}[\ln \pi^{\mathrm{E}}(\boldsymbol{u}|\boldsymbol{x}) - \ln \pi^{\mathrm{L}}(\boldsymbol{u}|\boldsymbol{x})], \qquad (2.34)$$

where $q(\boldsymbol{x}, \boldsymbol{u})$ is the state action distribution induced by the trajectory distribution $q(\boldsymbol{\tau})$ of the expert. Since $\mathbb{E}_q[\cdot]$ can be approximated using the trajectories drawn from $q(\boldsymbol{\tau})$, minimization of the KL divergence in (2.34) can be solved using only the demonstrated trajectories. Early studies on imitation learning such as [Widrow and Smith, 1964, Pomerleau, 1988] are based on this kind of supervised learning. However, these methods may not work well in many applications as indicated by [Ross et al., 2011, Bagnell, 2015].

On the contrary, we could try to base imitation learning techniques on an I-projection [Amari, 2016] that minimizes

$$D_{\mathrm{KL}}\left(p(\boldsymbol{\tau})||q(\boldsymbol{\tau})\right) = \int p(\boldsymbol{x}, \boldsymbol{u}) \ln \frac{\pi^{\mathrm{L}}(\boldsymbol{u}|\boldsymbol{x})}{\pi^{\mathrm{E}}(\boldsymbol{u}|\boldsymbol{x})} \mathrm{d}\boldsymbol{x}\mathrm{d}\boldsymbol{u} \tag{2.35}$$

$$= \mathbb{E}_p[\ln \pi^{\mathrm{L}}(\boldsymbol{u}|\boldsymbol{x}) - \ln \pi^{\mathrm{E}}(\boldsymbol{u}|\boldsymbol{x})]. \tag{2.36}$$

However, it is hard to minimize $D_{\mathrm{KL}}\left(p(\boldsymbol{\tau})||q(\boldsymbol{\tau})\right)$ in practice as we can not evaluate $\ln \pi^{\mathrm{E}}(\boldsymbol{u}|\boldsymbol{x})$, and there is no prior work on imitation learning methods that minimize $D_{\mathrm{KL}}\left(p(\boldsymbol{\tau})||q(\boldsymbol{\tau})\right)$ to the best of our knowledge. Exploring imitation learning methods based on I-projection will be an interesting research direction. Intuitively, the solution obtained by DAGGER [Ross et al., 2011] may result in a smaller KL-divergence under the I-projection than the one obtained by ordinary supervised learning as DAGGER attempts to achieve good performance under the learner's own data distribution.

# 3

# Behavioral Cloning

In this chapter, we review behavioral cloning (BC) methods. BC methods learn a direct mapping from states/contexts to trajectories/actions without recovering the reward function. Behavioral cloning can be an efficient way to reproduce the demonstrated behavior when such direct mapping is the most parsimonious way to represent the desired behavior.

We start by reviewing model-free BC methods and continue by reviewing model-based BC methods, which leverage information about system dynamics.

## 3.1  Problem Statement

A controller for a robotic system often has a hierarchical structure. Figure 3.1 shows a control diagram of a robotic system with imitation learning. The upper-level controller plans the desired trajectory based on a given context and/or observations. Meanwhile, the lower-level controller determines the control input to achieve the desired trajectory. The main target of imitation learning for robotic systems is to learn these controllers.

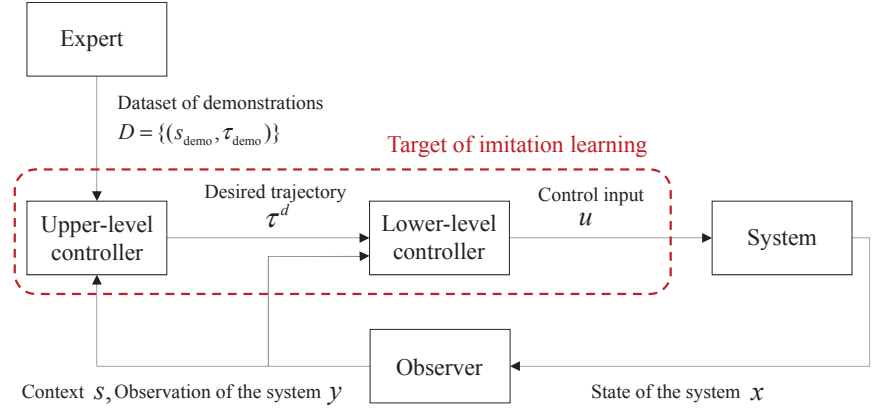

**Figure 3.1:** Control diagram of a robotic system with imitation learning. An expert demonstrates the desired behavior generating a dataset $D$. Based on $D$ and observations about the current context and system state an upper-level controller generates the desired trajectory $\boldsymbol{\tau}^d$. A lower-level feedback controller tries to follow $\boldsymbol{\tau}^d$ using observation feedback to generate a control $\boldsymbol{u}$ which causes a change to the system state $\boldsymbol{x}$ and a new observation. In imitation learning, the controllers are tuned to imitate the expert demonstrations.

When learning trajectories, the aim of imitation learning is to learn a policy that generates a desired trajectory

$$\boldsymbol{\tau}^d = \pi(\boldsymbol{s}) \tag{3.1}$$

for a given context $\boldsymbol{s}$. The context $\boldsymbol{s}$ can be the initial state of the robotic manipulator $\boldsymbol{x}_0$ or the state of objects relevant to a given task. In action-state space learning, the aim is to learn a policy that generates a control input $\boldsymbol{u}_t$ for a given state $\boldsymbol{x}_t$ and/or context $\boldsymbol{s}$,

$$\boldsymbol{u}_t = \pi(\boldsymbol{x}_t, \boldsymbol{s}). \tag{3.2}$$

In imitation learning, we assume that a dataset of experts' demonstrations is available. When learning trajectories, the dataset consists usually of a set of trajectories and contexts $\mathcal{D} = \{(\boldsymbol{\tau}_i, \boldsymbol{s}_i)\}_{i=1}^N$. In action-state space learning, the dataset will be given as a set of control inputs and states $\mathcal{D} = \{(\boldsymbol{u}_i, \boldsymbol{x}_i)\}_{i=1}^N$. Using such datasets, a policy can be learned as the direct mapping from the context to the trajectory or from the state to the control input. This learning problem can be formulated as a supervised learning problem in which a policy can be

---
**Algorithm 1** Abstract of behavioral cloning
---
 Collect a set of trajectories demonstrated by the expert $\mathcal{D}$
 Select a policy representation $\pi_{\boldsymbol{\theta}}$
 Select an objective function $\mathcal{L}$
 Optimize $\mathcal{L}$ w.r.t. the policy parameter $\boldsymbol{\theta}$ using $\mathcal{D}$
 **return** optimized policy parameters $\boldsymbol{\theta}$
---

obtained by solving a simple regression problem. We call this approach "*behavioral cloning*". Algorithm 1 abstracts the procedure of BC methods. The first step of BC is to record a set of expert demonstrations $\mathcal{D}$ which are usually given as a set of trajectories. Thereafter, we need to select a policy representation $\pi_{\boldsymbol{\theta}}$ appropriate for a given application. In addition, we need to select an objective function $\mathcal{L}$ that represents the similarity between the demonstrated behaviors and the learner's policy. The policy parameters $\boldsymbol{\theta}$ are then optimized using the collected dataset of demonstrations.

## 3.2 Design Choices for Behavioral Cloning

In addition to the design choices we described in Chapter 2, we list here some essential design choices for BC methods.

1. **What surrogate loss function should be used to represent the difference in demonstrated and produced behavior?** BC methods require a surrogate loss function which quantifies the difference between the demonstrated behavior and the learned policy. The choice of the surrogate loss function influences strongly how to train the policy, and we need to select the appropriate surrogate loss function to achieve efficient learning.

2. **What regression method should be used to represent the policy?** To obtain satisfactory system performance, it is essential to select the appropriate regression method. The regression model should be sufficiently expressive to represent the desired behavior but simple enough to allow for efficient training of the model. For efficient learning the regression method should be chosen together with the surrogate loss function.

### 3.2.1 Choice of Surrogate Loss Functions for Behavioral Cloning

We discuss some options for surrogate loss functions in this section.

#### 3.2.1.1 Quadratic Loss Function

The quadratic loss function is the most common choice for the loss function. Given two vectors, $\boldsymbol{x}_1$ and $\boldsymbol{x}_1$, a quadratic loss function is given by

$$\ell_{\text{quad}}(\boldsymbol{x}_1, \boldsymbol{x}_2) = (\boldsymbol{x}_1 - \boldsymbol{x}_2)^\top (\boldsymbol{x}_1 - \boldsymbol{x}_2). \tag{3.3}$$

For example, the difference between the state $\boldsymbol{x}^{\text{L}}$ induced by the learner's policy and the state $\boldsymbol{x}^{\text{demo}}$ demonstrated by the expert can be formulated as

$$\ell_{\text{quad}}(\boldsymbol{x}^{\text{L}}, \boldsymbol{x}^{\text{demo}}) = (\boldsymbol{x}^{\text{L}} - \boldsymbol{x}^{\text{demo}})^\top (\boldsymbol{x}^{\text{L}} - \boldsymbol{x}^{\text{demo}}). \tag{3.4}$$

The quadratic loss function is also called the $\ell_2$-loss function, and regression with minimizing the quadratic loss function is often called *least squares (LS) regression* or *$\ell_2$-loss minimization* Sugiyama [2015].

Minimizing the quadratic loss function is closely related to maximizing the expected log likelihood under the Gaussian distribution assumption. Let us consider the regression function $f_{\boldsymbol{\theta}}(\boldsymbol{x})$ parameterized by $\boldsymbol{\theta}$. Suppose that the target variable $y$ follows the equation

$$y = f_{\boldsymbol{\theta}}(\boldsymbol{x}) + \epsilon, \tag{3.5}$$

where $\epsilon$ is drawn from the Gaussian distribution as $\epsilon \sim \mathcal{N}(0, \sigma)$. In this model, the probability distribution of $y$ is given by

$$p(y|\boldsymbol{x}, \boldsymbol{\theta}) = \frac{1}{\sqrt{2\pi\sigma}} \exp\left(-\frac{(y - f_{\boldsymbol{\theta}}(\boldsymbol{x}))^2}{2\sigma}\right). \tag{3.6}$$

Finding the model $f_{\boldsymbol{\theta}}(\boldsymbol{x})$ that maximizes the expected log likelihood can be formulated as

$$\operatorname*{argmax}_{\theta} \mathbb{E}[\log p] = \operatorname*{argmax}_{\theta} \mathbb{E}\left[\log \exp\left(-\frac{(y - f_{\boldsymbol{\theta}}(\boldsymbol{x}))^2}{2\sigma}\right)^2\right] \quad (3.7)$$

$$= \operatorname*{arg\,min}_{\theta} \mathbb{E}[(y - f_{\boldsymbol{\theta}}(\boldsymbol{x}))^2] \quad (3.8)$$

$$\approx \operatorname*{arg\,min}_{\theta} \frac{1}{N} \sum_i (y - f_{\boldsymbol{\theta}}(\boldsymbol{x}))^2. \quad (3.9)$$

Therefore, minimizing the quadratic loss function is equivalent to maximizing the expected log likelihood under the Gaussian distribution. BC methods such as DMP [Schaal et al., 2004, Ijspeert et al., 2013] and ProMP [Paraschos et al., 2013, Maeda et al., 2016] learn a trajectory representation by minimizing quadratic loss functions.

Additionally, one can also use a weighted quadratic loss function

$$\ell_{\text{wquad}}(\boldsymbol{x}_1, \boldsymbol{x}_2, \boldsymbol{W}) = (\boldsymbol{x}_1 - \boldsymbol{x}_2)^\top \boldsymbol{W}(\boldsymbol{x}_1 - \boldsymbol{x}_2) \quad (3.10)$$

when an appropriate weight $\boldsymbol{W}$ is known. For example, Mahalanobis distance [Mahalanobis, 1936] given by

$$\ell_{\text{Mahal}}(\boldsymbol{x}_1, \boldsymbol{x}_2) = (\boldsymbol{x}_1 - \boldsymbol{x}_2)^\top \boldsymbol{\Sigma}^{-1}(\boldsymbol{x}_1 - \boldsymbol{x}_2), \quad (3.11)$$

where $\boldsymbol{\Sigma}$ is the covariance matrix of a distribution of interest, is often used in the literature [Rozo et al., 2016, Osa et al., 2017a].

### 3.2.1.2    $\ell_1$-Loss Function

The $\ell_1$-loss function is often employed for regression. The $\ell_1$-loss function is given by

$$\ell_{\text{abs}}(\boldsymbol{x}_1, \boldsymbol{x}_2) = \sum_i |x_{1,i} - x_{2,i}|, \quad (3.12)$$

where $x_{1,i}$ and $x_{2,i}$ are the $i$th element of the vectors $\boldsymbol{x}_1$ and $\boldsymbol{x}_2$, respectively. The $\ell_1$-loss function is also called the absolute loss function, and regression with minimization of $\ell_1$-loss is called *least absolute deviations regression* or *$\ell_1$-loss minimization* Sugiyama [2015]. Usually, $\ell_1$-loss minimization is more robust to outliers than $\ell_2$-loss minimization. This robustness can be attributed to the property of $\ell_1$-loss minimization, which gives the median of training samples, while $\ell_2$-loss minimization gives the mean of training samples. Effectively, in $\ell_2$-loss minimization a few large outliers can influence the mean significantly while in $\ell_1$-loss minimization the median can be largely unaffected by a few large outliers. In addition, unlike $\ell_2$-loss minimization, $\ell_1$-loss minimization results in a sparse solution, which can be computationally efficient. Although, in imitation learning, there are not many prior studies on using $\ell_1$-loss minimization, the discussed properties of the $\ell_1$-loss could be beneficial.

### 3.2.1.3 Log Loss Function

The log loss function is defined by

$$\ell_{\log}(q, p) = -\sum_i q_i \ln p_i, \tag{3.13}$$

where $q$ is the true probability and $p$ is the predicted probability. In binary classification, the log loss function is given by

$$\ell_{\log}(q, p) = -q \log p + (1 - q) \log(1 - p). \tag{3.14}$$

Since the log loss is equivalent to the cross entropy, the log loss is also called the cross-entropy loss [Sugiyama, 2015].

In binary classification (in imitation learning classification can be used to learn a discrete control policy from expert demonstrations), minimizing the log loss function is equivalent to maximizing the log likelihood in logistic regression. In more detail, suppose that we want to learn a binary classification where the probability follows the Bernoulli distribution

$$p(y = 1|\boldsymbol{x}, \boldsymbol{\theta}) = f_{\boldsymbol{\theta}}(\boldsymbol{x}), \quad p(y = 0|\boldsymbol{x}, \boldsymbol{\theta}) = 1 - f_{\boldsymbol{\theta}}(\boldsymbol{x}), \tag{3.15}$$

which can be more compactly written as

$$p(y|\boldsymbol{x}, \boldsymbol{\theta}) = (f_{\boldsymbol{\theta}}(\boldsymbol{x}))^y (1 - f_{\boldsymbol{\theta}}(\boldsymbol{x}))^{1-y}. \tag{3.16}$$

Maximizing the expected log likelihood $\mathbb{E}[\log p]$ of Bernoulli distributed data follows then as

$$\begin{aligned} \max \mathbb{E}[\log p] &= \max \mathbb{E}[y \log f_{\boldsymbol{\theta}}(\boldsymbol{x}) + (1-y)\log(1 - f_{\boldsymbol{\theta}}(\boldsymbol{x}))] \\ &= \max \frac{1}{N} \sum \left( y \log f_{\boldsymbol{\theta}}(\boldsymbol{x}) + (1-y)\log(1 - f_{\boldsymbol{\theta}}(\boldsymbol{x})) \right) \\ &= \min \ell_{\log}(y, f_{\boldsymbol{\theta}}(\boldsymbol{x})). \end{aligned} \quad (3.17)$$

Therefore, in binary classification, minimizing the log loss function is equivalent to maximizing the expected log likelihood under the Bernoulli distribution.

### 3.2.1.4 Hinge Loss Function

Hinge loss is a loss function often used for maximum margin optimization in classifiers such as support vector machines (SVMs) [Cortes and Vapnik, 1995]. Given two scalar variables, $x_1$ and $x_2$, the hinge loss can be defined as

$$\ell_{\text{hinge}}(x_1, x_2) = \max\left(0, 1 - x_1 x_2\right). \quad (3.18)$$

Intuitively, the hinge loss assigns zero costs if a classification is correct: $\ell_{\text{hinge}}(x_1, x_2) = 0$. For "wrong" classifications the cost is linear w.r.t. the parameters. This also explains the term "hinge"; in a visual illustration of the cost function one can imagine a hinge at $x_1 x_2 = 1$. While hinge loss is discontinuous at the "hinge" location $x_1 x_2 = 1$, optimization solutions still exist in practice. Moreover, since the hinge loss function is convex, it can be optimized efficiently with various convex optimizers.

### 3.2.1.5 Kullback-Leibler Divergence

In the field of information geometry, Kullback-Leibler (KL) divergence is used to quantify the difference between two probability distributions [Kullback and Leibler, 1951]

$$D_{\text{KL}}\left(p(x) || q(x)\right) = \int p(x) \ln \frac{p(x)}{q(x)} dx. \quad (3.19)$$

Since the KL divergence measures the difference between two probability distributions, it is useful when learning stochastic policies.

Please note that the KL divergence is not symmetric, therefore $D_{\text{KL}}(p(x)||q(x)) = D_{\text{KL}}(q(x)||p(x))$ does not hold in general. BC methods such as [Englert et al., 2013] use the KL divergence as the loss function.

### 3.2.2   Choice of Regression Methods for Behavioral Cloning

When applying behavioral cloning, an appropriate regression method must be chosen. Table 3.1 lists regression methods found in the literature. As discussed by Bishop [2006], one must choose a model that has appropriate complexity. Simple models which can be trained using linear regression are easy to train, but may not be sufficiently informative. Complex models such as neural networks can represent highly nonlinear mappings. However, training such complex models requires a large amount of training data. In addition, it is important to note that imitation learning cannot be addressed as simple supervised learning in many applications as we discussed in §2.7. We discuss an approach for *reducing* imitation learning to supervised learning with interaction in §3.4.3.

## 3.3   Model-Free and Model-Based Behavioral Cloning Methods

As discussed in §2.3, BC methods can be categorized into model-free and model-based methods. Table 3.2 shows advantages and disadvantages of both model-free and model-based BC methods.

Model-free BC methods learn a policy that reproduces the expert's behavior without learning/estimating system dynamics nor recovering the reward function. Since model-free BC methods do not require learning of system dynamics, model-free BC methods often do not require iterative learning and are relatively simple to implement compared to model-based BC methods. However, in trajectory learning, model-free BC methods do not ensure that the resulting trajectory is feasible in a given system. For this reason, it is hard to apply model-free methods to underactuated systems in which the set of reachable states is limited.

Contrary to model-free BC methods, model-based BC methods learn a policy using information about the system dynamics. By learning forward dynamics, it is possible to plan a feasible trajectory close to the expert's behavior even if a robotic system is underactuated. However, in many applications, learning a forward model is a non-trivial problem. In addition, model-based BC methods often require iterative learning, which is usually time-consuming compared with learning with model-free BC methods.

## 3.4 Model-Free Behavioral Cloning Methods in Action-State space

In this section we discuss behavioral cloning methods in action-state space. Although it seems that simple supervised learning can work in imitation learning, such a naive approach does not work in many applications. We will identify potential problems encountered when applying

**Table 3.1:** Regression methods in model-free behavioral cloning for both trajectory and action-state space learning. The output trajectory in trajectory learning consists of a long high dimensional sequence of variables while in action-state space learning the output is a single action. Therefore, some methods such as look-up tables have not been applied to trajectory learning. For modeling uncertainty in demonstrations, regression methods need to have explicit support for variance. Gaussian model, GMM and GPR methods model uncertainty explicitly.

| | | |
|---|---|---|
| Trajectory Learning | Gaussian Model | [Paraschos et al., 2013, Maeda et al., 2016] |
| | GMR | [Calinon and Billard, 2009, Gribovskaya et al., 2011, Khansari-Zadeh and Billard, 2014, Calinon, 2016] |
| | LWR | [Schaal and Atkeson, 1998, Mülling et al., 2013, Osa et al., 2017a] |
| | LWPR | [Vijayakumar et al., 2005] |
| | GPR | [Osa et al., 2017b] |
| Action-State Space | Look-Up Table | [Chambers and Michie, 1969] |
| | Linear Regression | [Widrow and Smith, 1964] |
| | Neural Network | [Pomerleau, 1988, LeCun et al., 2006, Stadie et al., 2017, Duan et al., 2017] |
| | Decision Tree | [Sammut et al., 1992] |
| | LWR | [Atkeson and Schaal, 1997] |
| | LWPR | [Vijayakumar and Schaal, 2000] |

supervised learning to imitation learning and discuss how we can alleviate these problems.

### 3.4.1 Model-Free Behavioral Cloning as Supervised Learning

Early studies on imitation learning such as [Widrow and Smith, 1964, Chambers and Michie, 1969, Pomerleau, 1988] employed supervised learning methods for imitation learning in action-state space. Among such early studies, in the seminal work ALVINN (Autonomous Land Vehicle In a Neural Network), Pomerleau [1988] developed an autonomous driving system using imitation learning. Pomerleau [1988] collected pairs of camera images and steering angles and trained a a neural network that modeled a direct mapping from camera images to steering angles. However, this simple approach can fail in practice and the autonomous car drives off the road quickly. As Bagnell [2015] indicated, learning errors *cascade* in sequential decision making, which makes the learner encounter unknown states that the expert never encounters in her/his successful demonstrations. Pomerleau [1988] described "*If the network is not presented with sufficient variability in its training exem-*

**Table 3.2:** A main choice when doing behavior cloning is whether to use a model-based or a model-free method. Model-free methods can directly learn a policy from data without learning a dynamics model. Direct learning also usually means that the learning algorithm does not need to iterate between trajectory and behavior generation. However, model-free methods are hard to apply to underactuated systems since without a model predicting desired behavior is hard. Model-based methods may work in underactuated systems but learning the actual model can be in many cases difficult.

|  | **Model-free** | **Model-based** |
|---|---|---|
| **Advantages** | A policy can be usually learned without iterative learning. | Applicable to underactuated systems. |
| **Disadvantages** | Hard to apply to underactuated systems. Hard to predict future states. | Model learning can be very difficult. An iterative learning process is often required. |

*plars to cover the conditions it is likely to encounter when it takes over driving from the human operator, it will not develop a sufficiently robust representation and will perform poorly. In addition, the network must not solely be shown examples of accurate driving, but also how to recover (i.e. return to the road center) once a mistake has been made."* That is, the distribution of the states that the learner encounters is different from the distribution of the states in the given demonstration data. Supervised learning is usually based on the assumption that training data samples are independent and identically distributed. However, this assumption is often violated in an imitation learning problem, especially when a policy for sequential decision making needs to be learned. To address this issue, Ross and Bagnell [2010], Ross et al. [2011] proposed an approach which reduces imitation learning to supervised learning with interaction, which we discuss in § 3.4.3.

### 3.4.2 Imitation as Supervised Learning with Neural Networks

Using neural networks for learning has attracted great interest in various fields. Supervised learning of neural networks can be also used for imitation learning: the desired neural network policy can be learned from the dataset generated/demonstrated by the expert. In this section, we shortly highlight some recent imitation learning successes with neural networks.

#### 3.4.2.1 Recent Successes of Imitation Leaning with Neural Networks

Recently, using neural networks for imitation learning has shown impressive results in certain applications such as learning to play Go [Silver et al., 2016], generating handwriting [Chung et al., 2015], generating natural language [Wen et al., 2015], or generating image captions [Karpathy and Fei-Fei, 2015]. Moreover, supervised learning of neural networks has been used as a building block for example for learning the policy or the cost function in inverse reinforcement learning (please see §4.4.6 for more details).

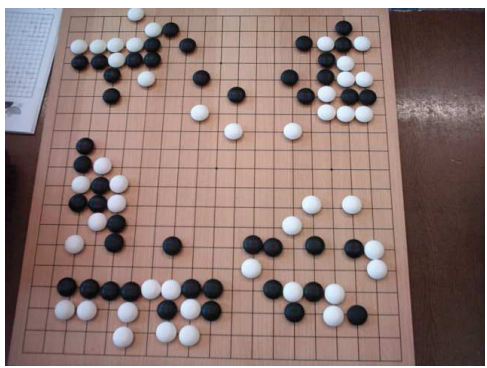

**Figure 3.2:** The game of Go is played on a 19x19 board. Even though the total number of possible board configurations exceeds $10^{170}$ and thus the training data can not cover all possible plays, the simple imitation learning approach in [Silver et al., 2016] was able to learn a competitive policy from demonstrations and improve the policy using self-play. [Figure from https://commons.wikimedia.org/wiki/File:Tuchola_026.jpg. CC license.]

Supervised imitation learning can be challenging when demonstrations do not cover the states that the learner encounters. For some applications, such as board games where the state space is known in advance, demonstrations could in principle be made to cover the state space. However, for example in the game of Go shown in Figure 3.2, the set of possible states is too large to cover completely and the supervised training approach needs to be able to generalize from training data. AlphaGo, an algorithm which was able to beat a human Go master [Silver et al., 2016], succeeded to learn a competitive Go policy using supervised imitation learning and then improve the policy using reinforcement learning.

AlphaGo trains a value network, which approximates the value function to predict the expected outcome of the game, and a policy network, which outputs actions using a representation of the image input of the board. The policy network is initialized by supervision using a large set of expert demonstrations, in total 30 million positions from the KGS Go Server. The value and policy networks are improved using data collected through self-play. AlphaGo selects actions by evaluating them with the policy and value networks.

The trained policy is a 13-layer deep neural network with alternating convolutional layers and rectifier nonlinearity layers, and the output is a soft-max layer resulting in a probability distribution over actions. The neural network receives as input a representation of the board state. For supervised training of the policy AlphaGo uses stochastic gradient ascent to maximize the likelihood of expert demonstrations w.r.t. parameters $\theta$: $\Delta\theta \propto \frac{\partial \log \pi_\theta(\boldsymbol{u}_t|\boldsymbol{x}_t)}{\partial \theta}$, where $\Delta\theta$ is the change in parameters, $\boldsymbol{u}_t$ is the expert action and $\boldsymbol{x}_t$ is the state. AlphaGo also utilizes a smaller, less accurate, but faster policy for predicting the expected outcome of actions.

### 3.4.2.2  Learning with Recurrent Neural Networks

In many applications, supervised learning of recurrent neural networks has made imitation learning of complex time series predictions possible. Wen et al. [2015] show how to generate human like natural language using a special form of the long short-term memory (LSTM) [Hochreiter and Schmidhuber, 1997] network. Wen et al. [2015] train their system using data collected from a spoken dialogue system. Table 3.3 shows an example of natural language generated by the trained neural network.

**Table 3.3:** Natural language generated by the semantically controlled LSTM (SC-LSTM) cell neural network proposed in [Wen et al., 2015]. The table shows an example dialogue act and related natural language samples from [Wen et al., 2015]. The neural network generates natural language learned from human demonstrations. The neural network is conditioned on the dialogue act which limits the generated sentences to specific meanings.

| *Dialogue act:* |
| --- |
| inform(name="red door cafe", goodformeal="breakfast", area="cathedral hill", kidsallowed="no") |
| *Generated samples:* |
| red door cafe is a good restaurant for breakfast in the area of cathedral hill and does not allow children . |
| red door cafe is a good restaurant for breakfast in the cathedral hill area and does not allow children . |
| red door cafe is a good restaurant for breakfast in the cathedral hill area and does not allow kids . |
| red door cafe is good for breakfast and is in the area of cathedral hill and does not allow children . |
| red door cafe does not allow kids and is in the cathedral hill area and is good for breakfast . |

As is common when designing neural network based systems, the neural network architecture in [Wen et al., 2015] is adapted to the task at hand. Moreover, neural network approaches need to take problems such as vanishing gradients, co-adaptation, and overfitting into account. Vanishing gradients can be a problem especially for recurrent neural networks due to the high optimization depth. The neural network architecture in [Wen et al., 2015] includes skip connections [Graves et al., 2013] to soften vanishing gradients and Wen et al. [2015] utilize dropout [Srivastava et al., 2014], a technique which randomly deactivates connections in the neural network during training, to reduce co-adaptation and overfitting.

Learning recurrent neural networks from demonstrations has been shown to work also for other kinds of data. Karpathy and Fei-Fei [2015] show how to learn to generate annotations for image regions from demonstrations. The approach of [Karpathy and Fei-Fei, 2015] learns from a combination of image and language data to generate natural language descriptions of images. Chung et al. [2015] show how to learn to generate handwriting and natural speech from demonstrations. Chung et al. [2015] propose a new type of recurrent neural network with hidden random variables and argue that random variables are needed to model variability in data with complex correlations between different time steps, for example, in natural speech.

### 3.4.3 Teacher-Student Interaction during Behavioral Cloning

Although the goal of imitation learning is to learn a policy that reproduces the expert's behavior, any learned policy will inevitably make at least occasional mistakes. As a result small error may cascade [Bagnell, 2015]: a small error at an early time-step may lead the learner to a state which deviates from expert demonstrations. Consequently, the learner will make further mistakes, leading to poor performance.

This highlights a central difference between imitation learning and the traditional setting of supervised learning, where we typically assume the input distribution to be independent and identically distributed [Shalev-Shwartz and Ben-David, 2014]. Instead, in imitation learning, the features/states in a dataset of demonstrations are not drawn from the distribution of the features which the learner will encounter using their own policy. This means that the assumption of independent and identically distributed (i.i.d.) data is often violated in imitation learning. Crudely speaking, a policy for recovering from mistakes needs to be learned as suggested by Pomerleau [1988].

However, in even modest-scale imitation learning problems it is infeasible to collect demonstrations under all possible situations, and instead we must focus corrections to the most relevant scenarios. Instead, a policy can be iteratively learned by alternating between policy updates and requesting additional demonstrations for the current state distribution [Ross and Bagnell, 2010, Ross et al., 2011, Bagnell, 2015]. We review methods that address this problem in the following section.

### 3.4.3.1 Reduction of Structured Prediction to Iterative Learning of Simple Classification

The task of learning a function that maps inputs $x$ to structured outputs $y$ (for example, parse trees, trajectories, matchings, etc. [Taskar, 2005]) is referred to as structured prediction [Tsochantaridis et al., 2005, BakIr et al., 2007]. Problems of imitation learning can often profitably be phrased as structured prediction [Ratliff et al., 2006b,a, 2009], and has led to developments of some techniques we cover extensively in this survey in § 4.4.2.

Conversely, Search-based structured prediction (SEARN) proposed by Daumé III et al. [2009], is a seminal work that demonstrated that one can also reduce structured prediction to a kind of imitation learning. In particular, SEARN crafts a series of reductions from structured prediction to simple classification. In SEARN, structured prediction is formulated as a search process over the components $y_t$ of the structured output $y$, where the *t*th decision is dependent on the preceding $t-1$ decisions. Therefore, the training process of a classifier in SEARN is dependent on the classifier itself.

SEARN learns a multiclass cost-sensitive classifier, e.g. [Zadrozny et al., 2003], for each state in the dataset through an iterative process. By performing the prediction using the current classifier $\pi$, SEARN creates new cost-sensitive samples. These cost-sensitive samples are used to learn a new classifier $\pi'$ which SEARN combines with the current classifier $h$ in a stochastic manner. Daumé III et al. [2009] show that the performance of SEARN is competitive with other methods such as structured SVM [Tsochantaridis et al., 2005] and Conditional Random Field [Lafferty et al., 2001], while often being tremendously faster to learn. Modern, high performance, implementations of such search based structured prediction use online learning methods of DAGGER [Ross et al., 2011, 2013], AggreVaTe [Ross and Bagnell, 2014, Sun et al., 2017], or LOLS [Chang et al., 2015, Daumé III and Langford, 2015].

However, for each time step, simple implementations of these search based structured prediction require a state reset and an expert demonstration. Such a reset is often infeasible in the physical world, and even if possible, the expert may need to provide a prohibitively large number of demonstrations. For these reasons, SEARN, AggreVaTe and LOLS require substantial care to implement efficiently, using *e.g.* bandit methods or value regression, and deal with resets [Chang et al., 2015].

#### 3.4.3.2 Confidence-Based Approach

Chernova and Veloso [2009] proposed a method that learns a policy by requesting additional expert demonstrations based on the confidence of a given state. In this method, the learner learns how to select the action from a finite set of action primitives by using classifiers that return selection confidence, e.g. Gaussian mixture models. When the confidence is lower than a threshold, additional expert demonstrations are requested. In addition, when the expert observes incorrect actions by the learner, the expert corrects the action and the corrected action is added to the training dataset. By requesting additional demonstrations, this method also tries to empirically learn a policy under the state distribution induced by the learner's policy.

---

**Algorithm 2** Confidence-based autonomy algorithm: confident execution and corrective demonstration [Chernova and Veloso, 2009]

---

**Input:** Demonstration of the action-state pairs $\mathcal{D} = \{(\boldsymbol{x}_i, \boldsymbol{u}_i)\}_{i=1}^{N}$, confidence threshold $c_0$
Initialize the policy $\pi$
**repeat**
  Observe the state $\boldsymbol{x}$
  Compute the confidence $c(\boldsymbol{x})$
  Plan action $\boldsymbol{u}^{\text{L}}$
  **if** $c(\boldsymbol{x}) < c_0$ **or** Corrective demonstration is necessary **then**
    Receive the demonstration data $\mathcal{D}_{\text{new}} = \{(\boldsymbol{x}^{\text{new}}, \boldsymbol{u}^{\text{new}})\}$
    Update the dataset $\mathcal{D} \leftarrow \mathcal{D} \cup \mathcal{D}_{\text{new}}$
    Update the policy $\pi^{\text{L}}$
  **end if**
**until** the task learned

---

#### 3.4.3.3 Data Aggregation Approach: DAGGER

Ross et al. [2011] proposed an meta-algorithm called DAGGER, which attempts to collect expert demonstrations under the state distribution induced by the learned policy. It can be seen most naturally as an *on-policy* approach [1] [Sutton and Barto, 1998] to imitation learning: the expert provides the correct actions to take, but the input distribution of examples comes from the learner's own behavior.

Figure 3.3 shows an overview of the DAGGER approach to imitation learning. The simplest form of DAGGER proceeds as follows. At the first iteration, the policy is initialized by behavioral cloning of the expert demonstrations, resulting in policy $\pi_1^{\text{L}}$. Subsequently, the policy is used to collect a dataset of trajectories, and those newly obtained trajectories and the demonstrated trajectories are aggregated into a dataset $\mathcal{D}$, which is used to train a policy $\pi_2^{\text{L}}$. At iteration $n$, a policy $\pi_n^{\text{L}}$ is used to collect more trajectories, and those trajectories are added to the dataset $\mathcal{D}$. The next policy $\pi_{n+1}^{L}$ is trained so that $\pi_{n+1}^{L}$ mimics the expert on the whole dataset $\mathcal{D}$. To leverage the presence of the expert, DAGGER queries partial expert demonstrations $\pi^{E}$ in the learning phase, and the policy $\pi_i = \beta_i \pi^{E} + (1 - \beta_i)\pi_i^{L}$— a stochastic mixing of expert and learner– is used to collect the next dataset. In other words, partial expert demonstrations are requested under the states induced by the learned policy $\pi_i^{L}$. Thus, DAGGER learns a policy from the expert demonstrations under the state distribution induced by the learned policy. Algorithm 3 shows the details of the general DAGGER algorithm.

---

[1] The first using of the phrasing of *on-policy*, which nicely evokes the closely related approaches and issues in Reinforcement Learning is due to [Laskey et al., 2017].

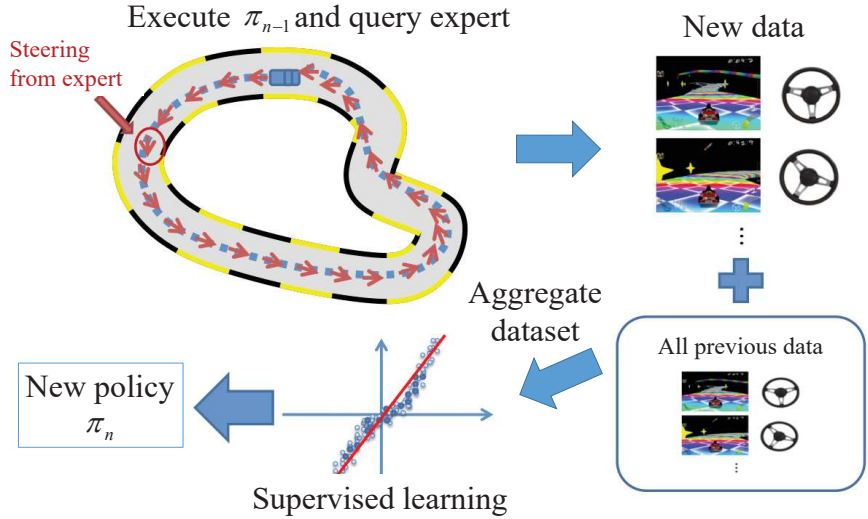

**Figure 3.3:** An overview of DAGGER from [Bagnell, 2015]. In each iteration, DAGGER generates new examples using the current policy with corrections (labels) provided by the experts, adds the new demonstrations to a demonstration dataset and computes a new policy to optimize performance in aggregate over that dataset. The figure illustrates a single iteration of DAGGER . The basic version of DAGGER initializes the demonstration dataset from a single set of expert demonstrations and then interleaves policy optimization and data generation to grow the dataset. More generally, there is nothing special about aggregating data– any method, like gradient descent or weighted majority that is sufficiently stable in its policy generation and does well on average over the iterations (or more broadly, all *no-regret* algorithm run over each iterations dataset) will achieve the same guarantees, and maybe strongly preferred for computational reasons.

---
**Algorithm 3** DAGGER [Ross et al., 2011]
---
    **Input:** initial dataset of demonstrations $\mathcal{D} = \{(\boldsymbol{x}, \boldsymbol{u})\}$, $\{\beta_i\}$ such that $\frac{1}{N}\sum_{i=1}^{N}\beta_i \to 0$
    Initialize: $\pi_1^{\mathrm{L}}$
    **for** $i = 1$ **to** $N$ **do**
        Let $\pi_i = \beta_i \pi^{\mathrm{E}} + (1-\beta_i)\pi_i^{\mathrm{L}}$.
        Sample trajectories $\boldsymbol{\tau} = [\boldsymbol{x}_0, \boldsymbol{u}_0, ..., \boldsymbol{x}_T, \boldsymbol{u}_T]$ using $\pi_i$
        Get dataset $\mathcal{D}_i$ of visited states by $\pi_i$ and actions given by expert.
        Aggregate datasets: $\mathcal{D} \leftarrow \mathcal{D} \cup \mathcal{D}_i$
        Train the policy $\pi_{i+1}^{\mathrm{L}}$ on $\mathcal{D}$.
    **end for**
    **return** best $\pi_i^{\mathrm{L}}$ on validation.
---

By collecting the expert demonstrations under the state which the learner encountered, DAGGER alleviates the problem that the state distribution induced by the learner's policy is different from the state distribution in the initial demonstration data. This approach significantly reduces the size of the training dataset necessary to obtain satisfactory performance[Ross et al., 2011], and often achieves much better performance even asymptotically. DAGGER can be interpreted as a *reduction* of imitation learning to supervised learning with interaction Bagnell [2015].

Crucially, the DAGGER approach is not limited to naive aggregation of all previous data: in fact, any algorithm (like gradient descent, some variants of newton's method, the exponentiated gradient approach, etc.) that enjoys the property of being *no-regret* can be used to learn iteratively on each newly collected data-set, and achieve the related formal guarantees. In practice, for instance, training complex policies with substantial training data is often based on online learning approaches like gradient descent.[2] We can think crudely of no-regret algorithms as the class of methods whose predictions are asymptoti-cally good on average over the data-sets they are presented, and are sufficiently stable between iterations [Hazan, 2016].
**Data as Demonstrator**: Venkatraman et al. [2015] extended DAGGER and proposed a framework called *Data as Demonstrator* (DaD) where the problem of multi-step prediction is formulated as imitation learning. Prediction errors will cascade over time in multi-step prediction as in the case of learning a policy, and this prediction error can also be improved by a data aggregation approach. Recent work shows the efficacy of DaD in control problems [Venkatraman et al., 2016].

---
[2]Note it is not technically correct to refer to these as *Stochastic Gradient Descent* (SGD) methods because the data being generated is not independent and identically distributed. Instead, the more general analysis of *Online Gradient Descent* [Hazan, 2016] is required.

## 3.5 Model-Free Behavioral Cloning for Learning Trajectories

In this section, we review approaches to learn trajectories from demonstrations. In robotic manipulation, trajectory planning is one of the most significant problems. If we assume that the system is (nearly) fully actuated and that a low-level controller to achieve the desired state is available, a trajectory for a given task can be learned without explicitly estimating the system dynamics. Since many commercialized robotic manipulators usually have such low-level controllers, this model-free BC approach has been dominant in imitation learning research for robotic manipulator trajectory planning. Next, we show how the choice of the trajectory representation influences trajectory learning and how the representation needs to fit to the application at hand.

### 3.5.1 Trajectory Representation

In order to learn trajectories we first need to define how to represent a trajectory. The choice of trajectory representation determines the parameterized space where demonstrated trajectories are projected. Therefore, it is essential to figure out the most parsimonious representation for a given application.

For planning a desired trajectory, we need a policy that generates a trajectory $\boldsymbol{\tau} \in \mathcal{T}$. The trajectory is given by a sequence of desired states and/or control inputs based on a given context $\boldsymbol{s} \in \mathcal{S}$. Given a set of demonstrated trajectories $\mathcal{D} = \{(\bm{s}_i, \bm{\tau}_i)\}_{i=1}^{N}$, we can use a supervised learning method to learn a policy which directly maps from contexts to trajectories

$$\pi : \mathcal{S} \mapsto \mathcal{T}. \tag{3.20}$$

For this purpose, we can use various regression methods developed in the field of machine learning. For example, Calinon et al. [2007] employed Gaussian mixture regression to model a mapping from time to states, and Osa et al. [2017b] used Gaussian Process regression for learning a mapping from contexts to trajectories. For learning such policies the choice of methods is usually not limited to specific regression methods, and we can also employ various machine learning techniques such as dimensionality reduction [Sugiyama, 2015] in order to alleviate the challenges of trajectory learning.

However, when planning a desired trajectory for a robotic system we need to ensure that the planned trajectory is physically feasible and a naive application of regression may not be the best choice. It is often necessary to impose some constraints on the planned trajectory, such as smooth convergence to the goal state. Such constraints may be implicitly satisfied when regression methods are used to learn a policy, but it is often convenient to use a policy that explicitly satisfies some constraints. Dynamic movement primitives (DMPs) [Schaal et al., 2004, Ijspeert et al., 2013] and the stable estimator of dynamical systems (SEDS) approach [Khansari-Zadeh and Billard, 2011] are representations that explicitly satisfy the condition of smooth convergence to the goal state. For learning these policies, regression methods are used in specific ways such that the desired constraints are satisfied. In the following, we discuss the details of different trajectory representations.

### 3.5.1.1 Keyframe/Via-Point Based Approaches

One obvious way to represent trajectories is as a sequence of keyframes or via-points. In the field of computer graphics, the term "keyframe" is used to express important states which are needed for accomplishing a given task [Parent, 2002]. In a keyframe-based approach, a task trajectory is represented as a sequence of keyframes. In robotic motion planning literature, the term "via-point" is used similarly to the term keyframe [Pastor et al., 2009, Paraschos et al., 2013, Zucker et al., 2013]. Instead of using the terms "keyframe" and "via-point", several articles describe a trajectory as consisting of a sequence of discrete states [Lee and Nakamura, 2009, Takano and Nakamura, 2015].

A keyframe-based trajectory representation appears in several imitation learning applications. Nakaoka et al. [2007] developed a humanoid system that learns dancing from human expert demonstration using a keyframe-based approach. The motion of the human expert was captured by a 3D motion tracking system, and the keyframes were subsequently extracted. By modifying the keyframes according to the dynamics of the humanoid, the humanoid was able to perform the demonstrated dance properly. Okamoto et al. [2014] developed a system that can perform a dance synchronously to music with difference rhythms by learning the correspondence between the music and the dancing motion.

Trajectories can be represented using discrete states. For discrete states one natural dynamics and observation model representation is the hidden Markov model which we will discuss next.

### 3.5.1.2 Representation with Hidden Markov Models

A hidden Markov model (HMM) is often used to model the probabilistic transition between discrete states [Inamura et al., 2004, Kulić et al., 2008, Lee and Nakamura, 2009, Takano and Nakamura, 2015]. A discrete HMM consists of a finite set of latent states $\boldsymbol{X}$, a finite set of observation labels $\boldsymbol{Y}$, a state transition matrix $A = \{a_{ij}\}$, an output probability matrix $B = \{b_{ij}\}$, and an initial distribution vector $\boldsymbol{d}_i$. When an HMM is used to represent motion, the latent state often represents the phase of the motion, and the observation represents the kinematic state of the system. Given a set of observation sequences and the set of states, $A$ and $B$ can be obtained by the Baum-Welch algorithm, which is a variant of the Expectation-Maximization (EM) algorithm. Once $A$ and $B$ are trained, a motion sequence can be estimated for a given initial state.

One of the benefits of an HMM representation is the ability to recognize the current system state based on the learned probabilistic model. HMMs have been used in classical speech recognition [Rabiner, 1989], and motion recognition can be performed in the same manner using HMMs [Inamura et al., 2004, Takano and Nakamura, 2015]. Given an HMM $\lambda = (A, B)$ and an observation sequence $\boldsymbol{Y}'$, the likelihood of observing a given sequence $p(\boldsymbol{Y}'|\lambda)$ can be computed. Therefore, the observed motion can be recognized as

$$\lambda^* = \arg \max_{\lambda} p\left(\boldsymbol{Y}'|\lambda\right). \tag{3.21}$$

In the framework in [Inamura et al., 2004], HMMs are used to represent primitive motions. The library of primitive motions are represented by a set of HMMs, and the motion is recognized based on the likelihood as in (3.21). This framework is extended to clustering and segmentation of demonstrated trajectories in [Kulić et al., 2008, Lee and Nakamura, 2009, Lee et al., 2010, Takano and Nakamura, 2015, 2016].

On the other hand, one of the drawbacks of the HMM representation is discreteness. Recognition with HMMs works well when the number of states is relatively low [Kulić et al., 2008]. However, HMMs with too few states may not be capable of reproducing a motion sequence. In robotic applications, HMMs are often used to represent the discrete high-level state of the system, assuming a low-level controller to achieve the desired state is available. However, it is non-trivial to plan smooth and feasible trajectories in many robotic systems.

To overcome the discreteness of HMMs, recent work uses other techniques in combination with HMMs, such as state specific Gaussian models [Calinon et al., 2010] to represent continuous values such as velocity, spatial position, or force [Racca et al., 2016]. Recent work also uses Hidden Semi-Markov Models (HSMM) [Yu, 2010] to model more complex state duration distributions [Calinon et al., 2011]. The work by Rozo et al. [2016] employs an LQR controller to address the problem of optimizing a trajectory retrieved from an HSMM. Additionally, Takano and Nakamura [2017] recently proposed an HMM-based method for planning joint torques to control the contact force.

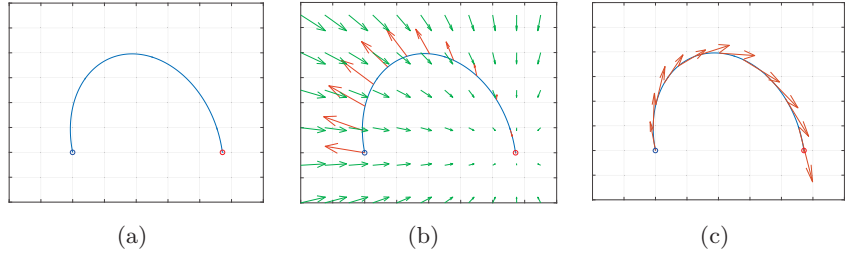

(a) (b) (c)

**Figure 3.4:** Schematic illustration of DMP. DMPs represent the demonstrated motion as a combination of a nonlinear force term and an attractor force term. Blue and red points represent the start and goal positions, respectively. Suppose that the trajectory shown in (a) is given as a demonstrated trajectory. The nonlinear force term along the trajectory, which is dependent on the phase of the motion, is shown as orange vectors, and in (b) green vectors represent the attractor force term, which is stationary and dependent on the state of the system. The dynamics of the demonstrated motion is learned as a sum of these terms shown in (c).

### 3.5.1.3 Dynamic Movement Primitives

Dynamic Movement Primitives (DMPs) were introduced by Ijspeert et al. [2002a,b], Schaal et al. [2004], Ijspeert et al. [2013]. DMPs are motivated by differential equations of well-defined attractor dynamics. Representation with DMPs ensures the smoothness and continuity of the trajectory. In addition, a DMP is able to represent nonlinear movements without losing the stability of the behavior. Figure 3.4 shows a schematic illustration of DMPs. DMPs represent demonstrated motion with a combination of a nonlinear force term and an attractor force term. The nonlinear force term enables expressing complex motions. Since the nonlinear force term decays in time, the goal attractor force term is dominant in the end of the motion and a path planned by a DMP smoothly converges to the goal state.

We describe details of DMPs in the following. In a DMP, the demonstrated motion with one degree of freedom (DoF) is modeled as a spring-damper system

$$\tau^2 \ddot{x} = \alpha_x \left( \beta_x (g - x) - \tau \dot{x} \right) + f, \quad (3.22)$$

where $\boldsymbol{x}$ is the state of the system, $f$ is the forcing function that determines the nonlinear behavior, $\alpha_x$ and $\beta_x$ are constants that determine the damping and spring behavior, respectively. $\tau$ is a constant that determines the temporal behavior, and $g$ denotes the goal state. In the example shown in Figure 3.4, the forcing function $f$ and the goal attractor term $\alpha_x \beta_x (g - x)$ are visualized in Figure 3.4(b). In imitation learning with DMPs, we can often assume that the final state $x_{\text{demo}}(T)$ of the demonstrated motion is the goal state $g = x_{\text{demo}}(T)$.

One significant feature of DMPs is time modulation by using a phase variable. By choosing the appropriate form of the basis function of the forcing function and the phase variable, DMPs can represent various movements with different execution speeds [Ijspeert et al., 2013]. Let us denote by $z$ a phase variable. For a striking movement, one can introduce the phase variable that follows the first-order linear dynamics as

$$\tau \dot{z} = -\alpha_z z, \tag{3.23}$$

where $\alpha_z$ is a constant. Ijspeert et al. [2013] called this equation the *canonical system* because it models the generic behavior of the system. In this case, the phase variable $z$ is given by a function of time $t$

$$z = z_0 \exp\left(-\frac{\alpha_z}{\tau} t\right), \tag{3.24}$$

where $z_0$ is the initial value of $z$. The phase variable $z$ exponentially converges to zero from an arbitrary initial state. Typically, the phase variable $z$ is used as $z \in [0, 1]$ for a striking movement.

The forcing function that models the nonlinear behavior is learned as a function of the phase variable $z$. Using a Gaussian basis function with this phase variable $z$, the forcing function can be formulated as

$$f(z) = (g - x_0) \sum_{i=1}^{M} \psi_i(z) w_i z, \tag{3.25}$$

where $x_0$ denotes the initial position and $M$ the number of the basis functions. The Gaussian basis function $\psi_i(z)$ is given by

$$\psi_i(z) = \frac{\exp\left(-h_i (z - c_i)^2\right)}{\sum_{j=1}^{N} \exp\left(-h_j (z - c_j)^2\right)}, \tag{3.26}$$

where $h_i$ and $c_i$ are constants that determine the width and centers of the basis functions, respectively. This system represents stable attractor

---

**Algorithm 4** Learning dynamic movement primitives [Schaal et al., 2004, Ijspeert et al., 2013]

---

**Input:** demonstrated trajectory $\boldsymbol{\tau}^{\text{demo}}$, parameters $\alpha_x, \beta_x, \tau, \alpha_z, \omega_z$
Choose a system of a phase variable $z$, e.g., (3.23)
Choose a basis function $\psi$ of the forcing function $f$
Compute the forcing function at each time step using $\boldsymbol{\tau}^{\text{demo}}$ with (3.27)
Find a weight vector $\boldsymbol{w}$ that minimize $\mathcal{L}_{\text{DMP}}$ in (3.28) using a least-square solution (3.29)

---

dynamics with nonlinear behavior. DMPs can be also used to represent rhythmic movements by using periodic basis functions [Schaal et al., 2004, Ijspeert et al., 2013].

If we assume that a demonstrated trajectory $\boldsymbol{\tau}^{\text{demo}}$ is given, the weight vector $\boldsymbol{w}$ can be learned as a supervised learning problem [Schaal et al., 2004, Ijspeert et al., 2013]. From the given trajectory, we compute the position, velocity and acceleration at each time step. To obtain the weight parameters in a DMP, we compute the target value of the forcing function from the given trajectory as

$$f_{\text{target}}(t) = \tau^2 \ddot{x}^{\text{demo}}(t) - \alpha_x \left( \beta_x(g - x^{\text{demo}}(t)) - \tau \dot{x}^{\text{demo}}(t) \right), \quad (3.27)$$

where $x^{\text{demo}}(t), \dot{x}^{\text{demo}}(t), \ddot{x}^{\text{demo}}(t)$ are the position, velocity and acceleration at the time $t$, respectively. Subsequently, we can find the weight vector $\boldsymbol{w}$ that minimizes the sum of the squared error

$$\mathcal{L}_{\text{DMP}} = \sum_{t=0}^{T} (f_{\text{target}}(t) - \xi(t)\boldsymbol{\Psi}\boldsymbol{w})^2, \quad (3.28)$$

where $\xi(t) = (g - x_0)z(t)$ for the discrete system and $\xi(t) = 1$ for the rhythmic system, and the entry of $\boldsymbol{\Psi}$ is computed as $\Psi_{ij} = \psi_i(t_j)$ with (3.25). The weight vector $\boldsymbol{w}$ can be obtained by a least-square solution

$$\boldsymbol{w} = \left( \boldsymbol{\Psi}^\top \boldsymbol{\Psi} \right)^{-1} \boldsymbol{\Psi}^\top \boldsymbol{F}. \quad (3.29)$$

For the attractor dynamics in (3.25), $\boldsymbol{F}$ is given by

$$\boldsymbol{F} = \left[ \frac{f_{\text{target}}(0)}{(g-x_0)z(0)}, \ldots, \frac{f_{\text{target}}(t)}{(g-x_0)z(t)}, \ldots, \frac{f_{\text{target}}(T)}{(g-x_0)z(T)} \right]^\top, \quad (3.30)$$

where $T$ is the number of the total time steps. Algorithm 4 summarizes the procedure for learning DMPs. Since DMPs are primarily designed for learning a motion for a single degree of freedom, multiple DMPs need to be learned for each dimension when learning motions with multiple dimensions.

**Variants of Dynamic Movement Primitives**: Since DMPs have been proposed, numerous variants of DMPs have been developed. Hoffmann et al. [2009] proposed an extended version of DMPs for obstacle avoidance and real-time goal adaptation. Deniša et al. [2016] developed Compliant Movement Primitives (CMPs) for learning compliant motions that require physical interaction between a robot and objects. For learning coupled motions, several variants of DMPs have been proposed by Kober et al. [2008], Gams et al. [2014], Amor et al. [2014]. Mülling et al. [2013] proposed a Mixture of Movement Primitives (MoMPs), which generalize the movement primitives to new contexts by mixing a set of learned movement primitives. DMPs have been applied to various robotic tasks and recognized as one standard representation of robotic motions.

**Relation to Hilbert Norm Minimization:** Dragan et al. [2015] revealed the relation between DMP-like methods and trajectory optimization based on Hilbert norm minimization such as CHOMP Zucker et al. [2013]. Dragan et al. [2015] formulated the problem of adapting a demonstrated trajectory $\boldsymbol{\tau}^{\text{demo}}$ to new start and goal states as minimization of the distance between the demonstration and the new trajectory subject to the new start and goal point constraints:

$$\boldsymbol{\tau}^* = \arg\min \left\| \boldsymbol{\tau}^{\text{demo}} - \boldsymbol{\tau} \right\|_{\boldsymbol{M}}^2 \quad (3.31)$$

$$\text{s.t. } \boldsymbol{x}(0) = \boldsymbol{x}_{\text{s}}^{\text{new}} \quad (3.32)$$

$$\boldsymbol{x}(T) = \boldsymbol{x}_{\text{g}}^{\text{new}} \quad (3.33)$$

where $\boldsymbol{x}_{\text{s}}^{\text{new}}$ and $\boldsymbol{x}_{\text{g}}^{\text{new}}$ are the new start and goal states, $\boldsymbol{M}$ is a linear operator that defines the inner product in the Hilbert space. When time is discrete, $M$ is a matrix, and the norm is given by $\|\boldsymbol{\tau}\|_M^2 = \boldsymbol{\tau}^\top M \boldsymbol{\tau}$. This formulation can be generalized to arbitrary norms, and Dragan et al. [2015] prove that trajectory adaptation with DMPs performs this norm minimization with a particular choice of the Hilbert norm, which is the same as the norm often used in trajectory optimization algorithms such as CHOMP Zucker et al. [2013].

### 3.5.1.4 Probabilistic Movement Primitives

While DMPs represent the movement in a deterministic way, demonstration performed by human experts is often stochastic. Such probabilistic behavior cannot be represented by DMPs. Probabilistic Movement Primitives (ProMPs) proposed by Paraschos et al. [2013] represent movement as a distribution over trajectories. In ProMP, the trajectory is parameterized as a linear combination of basis functions $\psi(t)$. The state of the system $\boldsymbol{x}(t)$ at time $t$ is expressed as

$$\boldsymbol{x}(t) = \begin{bmatrix} q(t) \\ \dot{q}(t) \end{bmatrix} = \boldsymbol{\Psi}(t)^\top \boldsymbol{\omega} + \boldsymbol{\epsilon}_x, \quad (3.34)$$

where $\boldsymbol{\Psi}(t)$ is a $M \times 2$ dimensional time-dependent basis matrix defined as

$$\boldsymbol{\Psi}(t) = \begin{bmatrix} \psi_1(t) & \dot{\psi}_1(t) \\ \vdots & \vdots \\ \psi_M(t) & \dot{\psi}_M(t) \end{bmatrix}, \quad (3.35)$$

$\boldsymbol{\omega}$ is a weight vector, and $\boldsymbol{\epsilon}_x \sim \mathcal{N}(0, \boldsymbol{\Sigma}_x)$ is zero-mean i.i.d. Gaussian noise. Here, the probability of observing the state $\boldsymbol{x}(t)$ is expressed as

$$p(\boldsymbol{x}(t)|\boldsymbol{\omega}) = \mathcal{N}(\boldsymbol{x}(t)|\boldsymbol{\Psi}(t)^\top \boldsymbol{\omega}, \boldsymbol{\Sigma}_x). \quad (3.36)$$

Thus, the probability of observing the whole trajectory $\boldsymbol{\tau} = [\boldsymbol{x}(0), \ldots, \boldsymbol{x}(T)]$ is written as

$$p(\boldsymbol{\tau}|\boldsymbol{\omega}) = \prod_t \mathcal{N}\left(\boldsymbol{x}(t)|\boldsymbol{\Psi}(t)^\top \boldsymbol{\omega}, \boldsymbol{\Sigma}_x\right). \quad (3.37)$$

By introducing a phase variable $z(t)$, we can achieve temporal modulation in ProMP. The phase variable is defined as $z(0) = 0$ at the beginning of the movement and as $z(T) = 1$ in the end. The basis function directly depends on the phase variable by replacing $\psi(t)$ with $\psi(z(t))$ and $\dot{\psi}(t) = \frac{d\psi}{dz}\frac{dz(t)}{dt}$.

The basis function should be selected according to the type of the movement as in DMPs. For point-to-point movements, one typical choice is a Gaussian function $b^{\mathrm{G}}$, that is,

$$b_i^{\mathrm{G}}(z(t)) = \exp\left(-\frac{(z(t)-c_i)^2}{2h}\right), \tag{3.38}$$

where $h$ defines the width of the basis function and $c_i$ is the center for the $i$th basis function. For rhythmic movements, the Von-Mises function can be used to model periodicity.

For imitation learning, the weight vectors $\boldsymbol{\omega}$ and the covariance matrix $\boldsymbol{\Sigma}_y$ need to be learned from the demonstrated trajectories. This problem can be formulated as a simple supervised learning problem. Let us assume that the trajectories demonstrated by experts are given as $D = [\tau^1, \ldots, \tau^N]$. If we assume that the demonstrated trajectories are aligned properly in the time domain, a weight vector $\boldsymbol{w}^i$ for the $i$th demonstrated trajectory can be obtained by minimizing the sum of squared errors

$$\mathcal{L}_{\mathrm{ProMP}} = \sum_{t=0}^{T}\left\|\boldsymbol{x}(t) - \boldsymbol{\Psi}(t)^{\top}\boldsymbol{w}\right\|^2, \tag{3.39}$$

where $\boldsymbol{x}(t) = [q(t)\ \dot{q}(t)]^{\top}$. The solution is given by a least squares solution

$$\boldsymbol{\omega}^i = \left(\boldsymbol{\Gamma}\boldsymbol{\Gamma}^{\top}\right)^{-1}\boldsymbol{\Gamma}\begin{bmatrix} q^i(0) \\ \dot{q}^i(0) \\ \vdots \\ q^i(T) \\ \dot{q}^i(T) \end{bmatrix}, \tag{3.40}$$

where the basis function matrix $\boldsymbol{\Gamma}$ is given by

$$\boldsymbol{\Gamma} = \begin{bmatrix} \psi_1(0) & \dot{\psi}_1(0) & \cdots & \psi_1(T) & \dot{\psi}_1(T) \\ \vdots & & \ddots & & \vdots \\ \psi_M(0) & \dot{\psi}_M(0) & \ldots & \psi_M(T) & \dot{\psi}_M(T) \end{bmatrix}. \tag{3.41}$$

For each demonstrated trajectory, we obtain a weight vector and for the whole set of demonstrated trajectories $D$ we obtain a set of weight

**Algorithm 5** Learning probabilistic movement primitives [Paraschos et al., 2013]

---
**Input:** Multiple demonstrated trajectories $\mathcal{D} = \{\boldsymbol{\tau}_i^{\text{demo}}\}_{i=1}^N$
Choose a basis function $\psi$ and the number of the basis function $M$
Compute the basis function matrix $\boldsymbol{\Psi}(t)$
**for** each demonstrated trajectory **do**
　　Obtain $\boldsymbol{\omega}$ by computing (3.40)
**end for**
Compute $p(\boldsymbol{\omega}) \sim \mathcal{N}(\boldsymbol{\mu_\omega}, \boldsymbol{\Sigma_\omega})$

---

vectors $\boldsymbol{\Omega} = [\boldsymbol{\omega}^1, \ldots, \boldsymbol{\omega}^N]$. From the set of weight vectors $\boldsymbol{\Omega}$ we can estimate a distribution over the weight vectors $p(\boldsymbol{\omega}) \sim \mathcal{N}(\boldsymbol{\mu_\omega}, \boldsymbol{\Sigma_\omega})$. The distribution of the state at time $t$ can be modeled as

$$p(\boldsymbol{x}(t)) = \mathcal{N}\left(\boldsymbol{x}(t) \left| \boldsymbol{\Psi}(t)^\top \boldsymbol{\mu_\omega}, \boldsymbol{\Psi}(t)^\top \boldsymbol{\Sigma} \boldsymbol{\Psi}(t) + \boldsymbol{\Sigma_x} \right.\right). \tag{3.42}$$

Algorithm 5 summarizes the procedure for learning ProMPs.

One of the characteristic features of ProMPs is the conditional distribution of the weight conditioned on a sequence of states $\boldsymbol{x}^* = [\boldsymbol{x}(t), \ldots, \boldsymbol{x}(t')]$. When $\boldsymbol{x}^*$ is specified as via-points, the distribution of the weight vector conditioned on $\boldsymbol{x}^*(t)$ is given as a Gaussian with mean and variance

$$\begin{aligned}\boldsymbol{\mu}_{\boldsymbol{\omega}}^+ &= \boldsymbol{\mu_\omega} + \boldsymbol{K}\left(\boldsymbol{x}^* - \boldsymbol{\Psi}(t)\boldsymbol{\mu_\omega}\right), \\ \boldsymbol{\Sigma}_{\boldsymbol{\omega}}^+ &= \boldsymbol{\Sigma_\omega} - \boldsymbol{K}\boldsymbol{H}^\top(t)\boldsymbol{\Sigma_\omega}.\end{aligned} \tag{3.43}$$

where $\boldsymbol{K} = \boldsymbol{\Sigma_\omega}\boldsymbol{H}^\top(t)\left(\boldsymbol{\Sigma_x} + \boldsymbol{H}^\top(t)\boldsymbol{\Sigma_\omega}\boldsymbol{H}(t)\right)^{-1}$ and $\boldsymbol{H}$ is the observation matrix defined as $\boldsymbol{H} = [\boldsymbol{\Psi}(t), \ldots, \boldsymbol{\Psi}(t')]^\top$. By using this conditioning, ProMPs can deal with modulation of via-points, final positions, or velocities. Figure 3.5 visualizes the conditioning of the trajectory distribution on the target position as an example.

### 3.5.1.5 Trajectory Representation with Time-Invariant Dynamical Systems

Khansari-Zadeh and Billard [2011] developed a framework to represent task trajectories as a time-invariant dynamical system (DS)

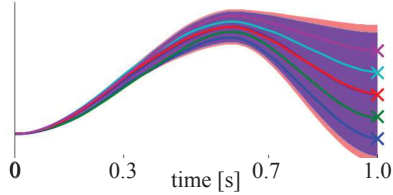

**Figure 3.5:** Conditioning of the learned distribution on the target position [Paraschos et al., 2013].

[Gribovskaya et al., 2011, Khansari-Zadeh and Billard, 2014]. While DMPs model the attractor dynamics and nonlinear behavior as separate terms, this framework models demonstrated movements as a single nonlinear dynamical system. The trajectory generated from this time-invariant DS is stably attracted to the given target position in the Lyapunov sense. The time-invariant DS representation cannot represent time-variant behavior by its nature.

[Khansari-Zadeh and Billard, 2011, Gribovskaya et al., 2011] modeled demonstrated trajectories as an *autonomous system* [Khalil, 1996], which follows time-invariant dynamics as

$$\dot{\boldsymbol{x}} = \boldsymbol{f}(\boldsymbol{x}), \tag{3.44}$$

where $\boldsymbol{x}$ is the system state, and $\boldsymbol{f}$ is a function that governs the behavior of the system. Khansari-Zadeh and Billard [2011], Gribovskaya et al. [2011] learn the function $\boldsymbol{f}$ as a GMM.

Let us define $\boldsymbol{x}$ as the state vector of the system. When a set of demonstrated trajectories is given, the joint distribution of $\boldsymbol{x}$ and $\dot{\boldsymbol{x}}$ can be estimated from the observations using a GMM. The $k$th component of the GMM models the distribution $p(\boldsymbol{x}, \dot{\boldsymbol{x}}|k)$ as

$$p(\boldsymbol{x}, \dot{\boldsymbol{x}}|k) \sim \mathcal{N}\left(\left[\begin{array}{c} \boldsymbol{x} \\ \dot{\boldsymbol{x}} \end{array}\right] \middle| \left[\begin{array}{c} \boldsymbol{\mu}_{\boldsymbol{x}} \\ \boldsymbol{\mu}_{\dot{\boldsymbol{x}}} \end{array}\right], \left[\begin{array}{cc} \boldsymbol{\Sigma}_{\boldsymbol{x},k} & \boldsymbol{\Sigma}_{\boldsymbol{x}\dot{\boldsymbol{x}},k} \\ \boldsymbol{\Sigma}_{\dot{\boldsymbol{x}}\boldsymbol{x},k} & \boldsymbol{\Sigma}_{\dot{\boldsymbol{x}},k} \end{array}\right]\right). \tag{3.45}$$

The estimated dynamics function $\hat{\boldsymbol{f}}$ is learned as

$$\hat{\boldsymbol{f}} = \sum_{k=1}^{K} h_k(\boldsymbol{x}) \left(\boldsymbol{\mu}_{\dot{\boldsymbol{x}}} + \boldsymbol{\Sigma}_{\dot{\boldsymbol{x}}\boldsymbol{x},k} \boldsymbol{\Sigma}_{\boldsymbol{x},k}^{-1}(\boldsymbol{x} - \boldsymbol{\mu}_{\boldsymbol{x},k})\right), \tag{3.46}$$

where

$$h_k(\boldsymbol{x}) = \frac{p(k)p(\boldsymbol{x}|k)}{\sum_{i=1}^{K} p(i)p(\boldsymbol{x}|i)} = \frac{\pi_x \mathcal{N}(\boldsymbol{x}|\boldsymbol{\mu}_{\boldsymbol{x},k}, \boldsymbol{\Sigma}_{\boldsymbol{x},k})}{\sum_{i=1}^{K} \pi_i \mathcal{N}(\boldsymbol{x}|\boldsymbol{\mu}_{\boldsymbol{x},i}, \boldsymbol{\Sigma}_{\boldsymbol{x},i})}, \quad (3.47)$$

where $\pi_k$ is the prior of the $k$th Gaussian component.

The study by Khansari-Zadeh and Billard [2011] showed that the system described by (3.46) is globally asymptotically stable at the target $\boldsymbol{x}^*$ if the condition

$$\begin{cases} \boldsymbol{A}^k + (\boldsymbol{A}^k)^\top \text{is negative definite,} \\ -\boldsymbol{A}^k \boldsymbol{x}^* = \boldsymbol{\mu}_{\dot{\boldsymbol{x}},k} - \boldsymbol{A}^k \boldsymbol{\mu}_{\boldsymbol{x},k}, \end{cases} \quad (3.48)$$

is satisfied for all $k = 1, \ldots, K$ where $\boldsymbol{A}^k = \boldsymbol{\Sigma}_{\dot{\boldsymbol{x}}\boldsymbol{x},k}(\boldsymbol{\Sigma}_{\boldsymbol{x},k})^{-1}$.

Khansari-Zadeh and Billard [2011] proved that (3.48) is the sufficient condition to show that the system is globally asymptotically stable in the sense of Lyapunov. For the details of the proof, we refer to the original paper [Khansari-Zadeh and Billard, 2011]. Khansari-Zadeh and Billard [2011] call this time-invariant DS represented by GMMs with constraints of (3.48) stable estimator of dynamical systems (SEDS).

This representation with time-invariant DS is nonparametric, and models the correlation of movements in multiple DoFs. In addition, this approach can be also used to learn second-order dynamics as $\ddot{\boldsymbol{x}} = \boldsymbol{g}(\boldsymbol{x}, \dot{\boldsymbol{x}})$ (please refer to [Khansari-Zadeh and Billard, 2011] for more details). The approaches with DS have been applied to various applications, such as learning coupled movements and learning stiffness [Shukla and Billard, 2012, Lukic et al., 2014, Kim et al., 2014].

The limitation of this approach is that the time-invariant representations cannot represent time-variant behaviors by its nature. In addition, due to the constraint of (3.48), SEDS can handle only models in which the dimensions of the input and output are equal [Shukla and Billard, 2012].

### 3.5.2 Comparison of Trajectory Representations

We show a comparison of different trajectory representations in Table 3.4. As can be seen from Table 3.4, every representation has strengths and weaknesses.

When choosing a trajectory representation, it is essential to consider the most parsimonious description for the desired trajectories and select a representation with a model complexity appropriate for the desired behavior. For example, SEDS in [Khansari-Zadeh and Billard, 2011, 2014] represents the motion as a time-invariant dynamical system. Although SEDS may be insufficient to model time-dependent motions, SEDS works well for some tasks such as catching a flying object [Kim et al., 2014]. With regard to stable attraction to a target position, global asymptotic stability is guaranteed in the sense of Lyapunov for SEDS [Khansari-Zadeh and Billard, 2011]. This property is useful for planning a stable behavior to approach a target position.

DMP is a good option for learning a point-to-point motion since motions can be easily generalized to different start and goal positions. In addition, bounded-input bounded-output (BIBO) stability is guar-anteed with regard to stable attraction to a target position. For this reason, DMP is often used to represent primitive motions in task-level motion planning Kroemer et al. [2015], Niekum et al. [2014], Manschitz et al. [2015]. On the other hand, stochasticity of the demonstrated trajectories cannot be encoded by DMPs, and multi-dimensional motion needs to be modeled by separate DMPs. ProMPs can address these problems. However, unlike DMP and SEDS, ProMPs do not guarantee stability of planned trajectories.

**Table 3.4:** Comparison of trajectory representations. Time dependence means here that the learned policy differs for each time step. With regard to stable attraction to a target position, bounded-input bounded-output (BIBO) stability is guaranteed for DMPs [Ijspeert et al., 2013], and global asymptotic stability is guaranteed in the sense of Lyapunov for SEDS [Khansari-Zadeh and Billard, 2011]. Stochasticity of trajectories means that a method takes uncertainty into account when modeling behavior. Encoding spatial coordination means here that a method can explicitly model the coordination of multi-dimensional motions.

|   | Time dependence | Stable attraction to a target position | Stochasticity of trajectories | Encoding spatial coordination patterns |
|---|---|---|---|---|
| Way points / Keyframe [Abbeel et al., 2010, Nakaoka et al., 2007] | ✓ | - | - | - |
| HMMs [Inamura et al., 2004, Takano and Nakamura, 2015] | (✓) | - | ✓ | ✓ |
| DMP [Schaal et al., 2004, Ijspeert et al., 2013] | ✓ | ✓ | - | - |
| ProMP [Paraschos et al., 2013, Maeda et al., 2016] | ✓ | - | ✓ | ✓ |
| SEDS [Khansari-Zadeh and Billard, 2011, 2014] | - | ✓ | - | ✓ |

In this section we presented several different trajectory representations and gave some suggestions how to choose them based on the different properties of the representations. However, the way to choose among the trajectory representations is still an interesting open question. Although efforts for benchmarking these different techniques have been made, e.g. [Lemme et al., 2015], it is necessary to establish metrics and benchmarks for comparing existing methods.

### 3.5.3 Generalization of Demonstrated Trajectories

Generalization of the demonstrated trajectories is one of the most important problems in imitation learning. The parameterization of trajectories enables generalizing the movements to new scenes. For example, a movement represented as a DMP can be adapted to a new scene by changing parameters such as goal and start positions. A popular approach for generalizing a parametrized motion is conditioning Gaussian distributions. This approach appears in several frameworks such as ProMP and SEDS. However, generalization with conditioning on Gaussian distributions is limited to situations where feature vectors with fixed length are available. Therefore, these methods often require manually selected feature vectors which are sufficiently informative. Another way to generalize demonstrated skills is to leverage geometrical warping from a demonstrated scene to a new scene. Recent work such as [Schulman et al., 2013, Lee et al., 2015a,b, Huang et al., 2015] propose methods for generalizing skills to new scenes based on non-rigid registration of point clouds, which does not rely on feature vectors of fixed length. In the following, we describe a short overview of generalization of demonstrated behaviors using different representations.

**Motion Generalization with DMP:** A trajectory represented with DMPs can be generalized to different start and goal positions [Schaal et al., 2004, Ijspeert et al., 2013]. For generalization according to additional features, some extensions are required. For example, Amor et al. [2014] proposed to model the joint distribution of DMP parameters and generalize learned motion in human-robot interaction scenarios.

**Motion Generalization with ProMP:** ProMP learns the distribution of the demonstrated trajectory in a parameter space. By conditioning the learned distribution, we can generalize the demonstrated trajectories to new start and goal positions or via-points [Paraschos et al., 2013, Maeda et al., 2016]. Maeda et al. [2016] show how to adapt learned ProMP skills in the context of human-robot interaction.

**Motion Generalization with SEDS:** Since the SEDS approach learns the joint distribution of the state and motion of the system, the demonstrated motion can be generalized to new states [Khansari-Zadeh and Billard, 2011].

**Trajectory Transfer with Geometrical Warping:** Although conditioning on Gaussian distributions are popular methods for generalizing skills, such methods are limited to the generalization with feature vectors with a fixed length. Another way to generalize demonstrated skills is to leverage Geometrical warping of the demonstrated scene to a new scene. Recently, Schulman et al. [2013] proposed a method to generalize the demonstrated trajectories based on non-rigid registration. In the non-rigid registration problem, one tries to find a correspondence between two point-sets and determine a good non-rigid transformation that can map one point-set onto the other [Chui and Rangarajan, 2003]. Thus far, non-rigid registration has been applied to for example template matching in OCR, motion generation in animation, or image registration in medical image analysis. Schulman et al. [2013] used non-rigid registration in order to transfer the demonstrated trajectories to new contexts as shown in Figure 3.6.

The trajectory transfer method consists of three steps: 1) find a transformation from the training scene to the test scene using a non-rigid registration method, 2) apply the transformation to the demonstrated end-effector trajectory in task space, and 3) convert the end-effector trajectory in task space into a joint space.

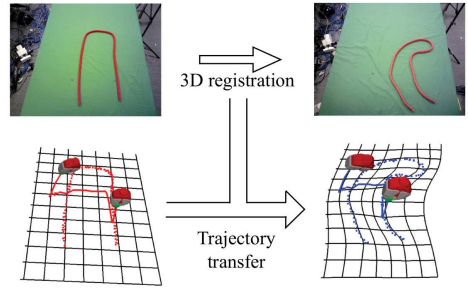

**Figure 3.6:** Trajectory transfer using non-rigid registration [Schulman et al., 2013].

This method has been extended in various ways [Lee et al., 2015a,b, Huang et al., 2015]. Trajectory transfer with non-rigid registration can be used to generalize both spatial motion and force profiles [Lee et al., 2015a]. Although the original work on trajectory transfer with non-rigid registration employed the thin plate spline robust point matching (TPS-RPM) approach proposed in [Chui and Rangarajan, 2003], the framework is not limited to specific non-rigid registration methods. The recent work by Lee et al. [2015b] shows that the use of the coherent point drift (CPD) algorithm improves trajectory transfer performance.

Unlike methods such as ProMPs or the dynamical systems approach, non-rigid registration based trajectory transfer works directly on point clouds and can generalize demonstrated trajectories to new scenes without modeling the distribution over demonstrated trajectories. However, non-rigid trajectory transfer requires that system dynamics are approximately invariant between source and target scenarios [Schulman et al., 2013]. In order to plan a trajectory in a new scene, one must select demonstrations performed in scenes with covariant system dynamics. For thousands of stored demonstrations, this search for an appropriate demonstration is a time-consuming process.

We discussed generalizing policies to new demonstrated trajectories. Table 3.5 shows a comparison of methods for generalizing demonstrated trajectories. DMPs allow stable convergence to arbitrary goal positions, but DMPs' generalization capability is relatively limited compared to other methods. ProMPs can generalize trajectories by Gaussian condi-tioning, but there is no guarantee of stable behavior. SEDS can generalize the trajectories with a guarantee of stable behavior, but cannot model the time dependence of movements. Trajectory transfer using non-rigid registration can achieve complex generalization, but does not incorporate stochasticity in demonstrations and there is no guarantee of stable behavior.

In addition to methods discussed above, there are numerous studies on generalizing demonstrated trajectories. Calinon [2015] proposed task-parameterized Gaussian mixture model (TP-GMM), which encodes the context information in its trajectory model. The approach based on TP-GMM has been recently employed in several studies [Calinon, 2016, Rozo et al., 2016]. The recent work by Osa et al. [2017a] proposed a trajectory optimization method for collision avoidance, which incorporates the distribution of the demonstrated trajectories. In ad-dition, although we focused on the trajectory-based approach, recent work such as [Finn et al., 2017b, Nair et al., 2017, Liu et al., 2017, Rahmatizadeh et al., 2017] addressed the problem of generalizing skills based on visual information by using a deep learning approach, which is a promising way to deal with complex environments.

**Table 3.5:** Generalization of skills using existing methods. DMPs enable stable convergence to arbitrary goal positions. ProMPs can generalize trajectories by Gaussian conditioning, but there is no guarantee of stable behavior. SEDS can generalize trajectories while guaranteeing stable behavior, but cannot model time dependence of movements. Trajectory transfer using non-rigid registration can achieve complex generalization, but does not incorporate stochasticity of demonstrations and there is no guarantee of stable behavior.

| Method | Generalizable context | Advantages | Disadvantages |
|---|---|---|---|
| DMP [Schaal et al., 2004, Ijspeert et al., 2013] | Start and goal positions | Guarantee of stable behavior | Limited generalization capabilities |
| ProMP [Paraschos et al., 2013, Maeda et al., 2016] | Any subset of the observations of the system | Generalization based on stochasticity of demonstrations | No guarantee of stable behavior |
| SEDS [Khansari-Zadeh and Billard, 2011, 2014] | State of the system with fixed dimensionality | Generalization with guarantee of stable behavior | No time-dependence |
| Way points with non-rigid registration [Schulman et al., 2013] | A point cloud of the given scene | Generalization based on point clouds of a given scene | Stochasticity of demonstrations is not considered |

### 3.5.4 Information Theoretic Understanding of Model-Free BC

Trajectory representations such as DMP, ProMP, and SEDS parameterize the trajectories as $p(\boldsymbol{\tau}|\boldsymbol{w})$ by solving linear equations using a least-squares method. Solving linear equations by minimizing a sum-of-squares error function is equivalent to maximizing the likelihood for the given dataset of demonstrations $\mathcal{D} = \{\boldsymbol{\tau}_i^{\text{demo}}\}_{i=1}^N$ under the assumption that the noise is drawn from a Gaussian distribution. This solution can be interpreted from an information theoretic point of view.

According to information theory, the entropy is a quantity that represents the amount of information, and the KL divergence can be obtained as a Bregman divergence derived from the entropy [Amari, 2016]. As described in [Bishop, 2006], finding parameters that maximize the likelihood $p(\boldsymbol{\tau}|\boldsymbol{w})$ for the given dataset is equivalent to minimizing the KL divergence given by

$$D_{\text{KL}}\left(q(\boldsymbol{\tau})||p(\boldsymbol{\tau}|\boldsymbol{w})\right) = \int q(\boldsymbol{\tau}) \ln \frac{q(\boldsymbol{\tau})}{p(\boldsymbol{\tau}|\boldsymbol{w})} \mathrm{d}\boldsymbol{\tau}.$$

where $q(\boldsymbol{\tau})$ is the distribution of the trajectory induced by the experts' policy. A sample of the demonstrated trajectories $\boldsymbol{\tau}^{\text{demo}}$ is drawn from the distribution $q(\boldsymbol{\tau})$ induced by the experts' policy. Therefore, the expectation with respect to $q(\boldsymbol{\tau})$ can be approximated as

$$D_{\text{KL}}\left(q(\boldsymbol{\tau})||p(\boldsymbol{\tau}|\boldsymbol{w})\right) \simeq \frac{1}{N} \sum_{i=1}^N \left( -\ln p(\boldsymbol{\tau}_i^{\text{demo}}|\boldsymbol{w}) + \ln q(\boldsymbol{\tau}_i^{\text{demo}}) \right). \quad (3.49)$$

Since $\ln q(\boldsymbol{\tau})$ is independent from $\boldsymbol{w}$, minimizing $D_{\text{KL}}\left(q(\boldsymbol{\tau})||p(\boldsymbol{\tau}|\boldsymbol{w})\right)$ is equivalent to maximizing the likelihood $\ln p(\boldsymbol{\tau}|\boldsymbol{w})$ for the given dataset $\mathcal{D}$.

Therefore, a policy obtained by model-free BC methods based on maximizing the likelihood under the Gaussian noise assumption can be

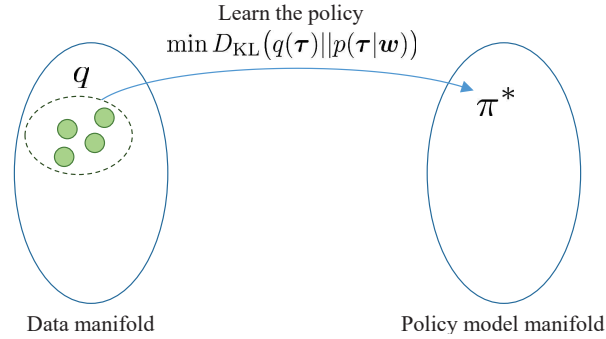

**Figure 3.7:** Schematic illustration of model-free BC methods. Model-free BC methods can be often interpreted as an M-projection onto the policy model manifold.

regarded as the policy that minimizes the KL divergence as

$$\pi^* = \arg\min_{\pi} D_{\mathrm{KL}}\left(q(\boldsymbol{\tau}) || p(\boldsymbol{\tau}|\boldsymbol{w})\right).$$

Thus, we can see that model-free methods discussed in the previous section parameterize the demonstrated behaviors by minimizing the KL divergence in a different parameter space as shown in Figure 3.7.

It is important to note that these model-free methods can suffer from the problem of *covariate shift* where the distribution of the test condition is different from the distribution of the demonstrated conditions. In other words, the learned skill may not work when the test condition is too different from the demonstrated condition. To cope with this problem, we will need incremental learning methods, which are discussed in § 3.5.7.

### 3.5.5 Time Alignment of Multiple Demonstrations

When the expert demonstrates the task trajectory multiple times, the execution speeds are different for each demonstration. Therefore, when a task trajectory is learned from multiple demonstrations, the time alignments of the demonstrated trajectories often need to be normalized if a time-dependent trajectory representation is used.

For this purpose, dynamic time warping (DTW) proposed by Sakoe and Chiba [1978] is often employed. Although DTW is originally devel-

---

**Algorithm 6** Estimate the latent trajectory and the time alignments of multiple demonstrations [van den Berg et al., 2010]

---

**Initialize:** $R^j = I$, and $z_t^j = z\frac{T^j}{T_{\text{ave}}}$
**repeat**
  $\boldsymbol{\xi} \leftarrow \text{KalmanSmoother}(\boldsymbol{y}, \boldsymbol{R}, z)$
  $\boldsymbol{R} \leftarrow \arg\max_R \mathbb{E}_{\boldsymbol{\xi}}\left(l(\boldsymbol{R}|\boldsymbol{\xi}, \boldsymbol{y})\right)$
  $z^j \leftarrow \arg\max_z \mathbb{E}_{\boldsymbol{\xi}}\left(l(z^j|\boldsymbol{\xi}, \boldsymbol{y})\right)$
**until** convergence

---

oped for speech recognition, DTW is frequently used to deal with the time alignment of trajectories in robotics. The original formulation of DTW finds the best time alignment of two data sequences. However, we often obtain more than two demonstrations, and we need to align all of them appropriately in the time domain.

In the field of imitation learning, Coates et al. [2008] proposed a method to normalize the time alignment of multiple demonstrated trajectories. Similar approaches appear in applications such as autonomous helicopter flight [Abbeel et al., 2010] and automation of robotic surgery [van den Berg et al., 2010, Osa et al., 2014]. Here, we review the method employed by van den Berg et al. [2010].

van den Berg et al. [2010] regarded the demonstrated trajectories as noisy '*observations*' of the '*reference*' trajectories. The reference trajectory and the time mapping from the reference trajectory to the demonstrated trajectory are computed using the EM (Expectation Maximization)-algorithm.

The linear system is described as

$$\boldsymbol{\xi}(t+1) = \begin{bmatrix} \boldsymbol{A} & \boldsymbol{B} \\ 0 & \boldsymbol{I} \end{bmatrix} \boldsymbol{\xi}(t) + \boldsymbol{w}(t),\ \boldsymbol{w}(t) \sim \mathcal{N}\left(0, \begin{bmatrix} \boldsymbol{P} & 0 \\ 0 & \boldsymbol{Q} \end{bmatrix}\right) \quad (3.50)$$

where $\boldsymbol{\xi}(t) = [\boldsymbol{x}^\top(t), \boldsymbol{u}^\top(t)]^\top$ is the state and the control input of the system at time $t$, $A$ and $B$ are the state matrix and the input matrix, respectively. $\boldsymbol{w}(t)$ is the noise that follows the zero-mean Gaussian distribution. $P$ and $Q$ are the covariance matrices of process noise and observation noise, respectively. If we assume that the $j$th demonstrated trajectory $\boldsymbol{\tau}^j$ is given by $\boldsymbol{\tau}^j = [\boldsymbol{x}^j(0), \boldsymbol{u}^j(0), \cdots, \boldsymbol{x}^j(T^j), \boldsymbol{u}^j(T)^j]$, the relation between the reference trajectory and the observed trajectories is represented as

$$\begin{bmatrix} \boldsymbol{\tau}^1(z_t^1) \\ \vdots \\ \boldsymbol{\tau}^N(z_t^N) \end{bmatrix} = \begin{bmatrix} \boldsymbol{I} \\ \vdots \\ \boldsymbol{I} \end{bmatrix} \boldsymbol{\xi}(t) + \boldsymbol{v}(t), \ \boldsymbol{v}(t) \sim \mathcal{N}\left(0, \begin{bmatrix} \boldsymbol{R}^1 & 0 & 0 \\ 0 & \ddots & 0 \\ 0 & 0 & \boldsymbol{R}^N \end{bmatrix}\right), \quad (3.51)$$

where $\boldsymbol{v}$ is the noise that follows a zero-mean Gaussian distribution, and $z_t^j$ is the mapping of time $t$ in the reference trajectory $\boldsymbol{\xi}$ to the corresponding time in trajectory $\boldsymbol{\tau}^j$. The covariance matrices $\boldsymbol{R}^j$ behave as weights on the $j$th demonstrated trajectory $\boldsymbol{\tau}^j$ for estimating the reference trajectory $\boldsymbol{\xi}$.

The reference trajectory $\boldsymbol{\xi}$, covariance matrices $\boldsymbol{R}$ and the time-mapping $\tau$ are estimated using the EM algorithm. In the E-step, the reference trajectory $z$ can be estimated using a Kalman smoother based on the model in (3.50). In the M-step, the time mapping $\tau$ and the covariance matrices $\boldsymbol{R}$ are updated by maximizing the likelihood with respect to the estimated $\boldsymbol{z}$. DTW is used to update the time mapping $\tau$ in [Abbeel et al., 2010, van den Berg et al., 2010]. This procedure is summarized in Algorithm 6.

### 3.5.6 Learning Coupled Movements

It is often necessary to learn the correlation of movements between multiple DoFs or multiple agents. For example, in human-robot interaction, an autonomous agent needs to know how to react to a human operator's movements. In such a case, the human movement and the robot reaction can be considered as coupled movements. In this section, we review how to learn such correlations of movements with multiple DoFs or agents. One typical approach is modeling the joint distribution of the parameterized trajectories in multiple DoFs with a Gaussian (or a mixture of Gaussians) distribution. When partial observations of the coupled movements are given, the rest of movements are estimated by computing the conditional distribution on the partial observation. We will see in the following section that the choice of the trajectory representation plays an important role in modeling the trajectory distribution.

#### 3.5.6.1 Learning Coupled Movements with DMPs

DMPs have been used to learn both perceptual coupling and coupling for human-robot collaborative motion [Kober et al., 2008, Amor et al., 2014]. In robotic applications, a movement is often represented as trajectories in multiple spaces. For example, a position of an end effector can be measured using a vision system in Cartesian space, while a trajectory of a robotic manipulator is often controlled in joint space. When DMP is used, trajectories in different spaces are often learned as separate DMPs. However, it is essential to learn the coupling between the trajectories in different spaces. Kober et al. [2008] proposed to learn such perceptual coupling for motor skills with DMPs. Instead of using the forcing function shown in (3.25), the perceptual coupling is modeled using the modified forcing function

$$\hat{f} = \sum_{i=1}^{M} \psi_i(z)\hat{\boldsymbol{w}}z + \sum_{j=1}^{M_c} \hat{\psi}_j(z) \left( \boldsymbol{\kappa}_j^\top (\boldsymbol{y} - \bar{\boldsymbol{y}}) + \boldsymbol{\delta}_j^\top (\dot{\boldsymbol{y}} - \dot{\bar{\boldsymbol{y}}}) \right), \qquad (3.52)$$

where $\boldsymbol{y}$ denotes the state of the external variable, $\bar{\boldsymbol{y}}$ is the expected state of the external variable, $\boldsymbol{\kappa}$ and $\boldsymbol{\delta}$ are the coupling factors that act as the gains on difference between the desired and actual behaviors of the external variable. $M_c$ is the number of the basis function for modeling the coupled behavior. While the weight vectors $\boldsymbol{w}$ and $\hat{\boldsymbol{w}}$ can be learned from a single demonstration, the coupling factors $\boldsymbol{\kappa}$ and $\boldsymbol{\delta}$ cannot be learned from demonstrations since the deviation from the nominal behavior is necessary for learning these parameters. For this reason, Kober et al. [2008] used a reinforcement learning method for learning $\boldsymbol{\kappa}$ and $\boldsymbol{\delta}$ through trial and error.

#### 3.5.6.2 Learning Coupled Movements with Gaussian Conditioning

Statistical machine learning methods offer ways to model correlation of variables. For example, Gaussian conditioning is a simple way to model such correlations. Coupled motion in robotic applications can be learned using such statistical methods. Amor et al. [2014] represented the motions of two agents using DMPs and learned the correlations of the distribution of the motion parameters. When one agent's motion is observed, the motion of the other agent can be predicted based on Gaussian conditioning.

Likewise, ProMPs have also been used to learn the correlation of multiple agents' motion. Maeda et al. [2016] developed an imitation learning framework called Interaction ProMP to learn coupled motions in human-robot collaboration. In the framework of Interaction ProMP, correlated movements are learned as a distribution of the correlated weight vectors of ProMPs. Using a partial observation of the movement, unobserved movements are estimated as a conditional distribution of the weight vectors on the given partial observation.

Here, we describe details of Interaction ProMP. Suppose demonstrations of human robot collaborative movements are given. Here, we define the state vector as a concatenation of the $P$ DoFs executed by the human, followed by the $Q$ DoFs executed by the robot

$$\boldsymbol{x}(t) = \left[ \begin{array}{c} \boldsymbol{x}^h(t) \\ \boldsymbol{x}^r(t) \end{array} \right], \tag{3.53}$$

where $\boldsymbol{x}^h(t)$ is a $P \times 1$ dimensional vector that represents the state of the human, and $\boldsymbol{x}^r(t)$ is a $Q \times 1$ dimensional vector that represents the state of the robot at time $t$. The distribution of the trajectory is parameterized as

$$p(\boldsymbol{x}|\boldsymbol{\omega}) = \mathcal{N}(\boldsymbol{x}|\boldsymbol{H}^\top(t)\boldsymbol{\omega}, \boldsymbol{\Sigma}_y), \tag{3.54}$$

where

$$\boldsymbol{H}^\top(t) = \text{diag}(\boldsymbol{\Psi}^\top(t), \ldots, \boldsymbol{\Psi}^\top(t)), \tag{3.55}$$

$\boldsymbol{\Psi}^\top(t)$ is a $M \times 2$ matrix defined as (3.35) and $M$ is the number of basis functions. When a trajectory of a human-robot collaborative movement is demonstrated, the weight vector $\boldsymbol{\omega}$ can be learned as

$$\bar{\boldsymbol{\omega}} = [(\boldsymbol{\omega}_1^h)^\top, \ldots, (\boldsymbol{\omega}_P^h)^\top, (\boldsymbol{\omega}_1^r)^\top, \ldots, (\boldsymbol{\omega}_Q^r)^\top]^\top. \tag{3.56}$$

By learning from multiple demonstrations, we can obtain the distribution of the weight vector $p(\bar{\boldsymbol{\omega}}) \sim \mathcal{N}(\boldsymbol{\mu}_{\bar{\boldsymbol{\omega}}}, \boldsymbol{\Sigma}_{\bar{\boldsymbol{\omega}}})$ where $\boldsymbol{\mu}_{\bar{\boldsymbol{\omega}}} \in \mathbb{R}^{(P+Q)M \times 1}$ and $\boldsymbol{\Sigma}_{\bar{\boldsymbol{\omega}}} \in \mathbb{R}^{(P+Q)M \times (P+Q)M}$. After learning the distribution of the weight vector $p(\bar{\omega})$, the robot's reaction to an observed human movement can be planned as the conditional distribution of the weight vectors. When a sequence of the observations of the human movement $y^*$ is given, the conditional distribution of the ProMP parameters given the observation, $p(\bar{\omega}|y^*)$, can be computed by applying the Bayes theorem (3.43). By using a mixture of Interaction ProMPs, the non-Gaussian distribution $p(\bar{\omega})$ can be represented as a mixture of Gaussians [Ewerton et al., 2015, Maeda et al., 2016]. The framework of Interaction ProMPs is summarized in Figure 3.8.

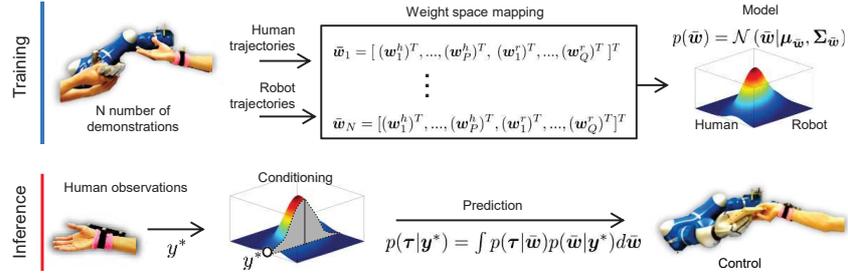

**Figure 3.8:** Overview of Interaction ProMPs in [Maeda et al., 2016]. In the interaction ProMP framework, correlated movements are learned as the joint distribution of weight vectors of ProMPs. Thanks to the probabilistic modeling of the trajectory distribution, the interaction ProMP framework works with noisy observations of trajectories [Maeda et al., 2016]. In this figure, $\bar{\omega}$ represents the weight vector that contains movements of all DoFs controlled by the robot and the human operator as defined in (3.56).

In the Interaction ProMP framework, correlated movements are learned as correlated weight vectors of ProMPs. Thanks to the probabilistic modeling of the trajectory distribution, the interaction ProMP framework works with noisy observations of trajectories [Maeda et al., 2016].

### 3.5.6.3 Learning Coupled Movements with Time-Invariant Dynamical Systems

The Time-invariant dynamical system (DS) approach in [Khansari-Zadeh and Billard, 2011] can be also used to learn coupled move-ments [Shukla and Billard, 2012, Lukic et al., 2014, Kim et al., 2014]. Shukla and Billard [2012] developed a framework for learning coupled movement based on DS, which they call the Coupled Dynamical System (CDS) model. The idea of CDS is to model the correlation between two agents using statistical models.

Let us assume two agents, which we call the *master* and *slave*, perform a coupled motion. The correlation of the movement of the master $\boldsymbol{x}_s$ and the movement of the slave $\boldsymbol{x}_s$ can be modeled with CDS. In CDS, three GMMs are trained to model three joint distributions:
1) the joint distribution of the master movement $p(\boldsymbol{x}_m, \dot{\boldsymbol{x}}_m)$
2) the joint distribution of the states of the master and the desired state of the slave $p\left(\Phi(\boldsymbol{x}_m), \boldsymbol{x}_s^d,\right)$
3) the joint distribution of the slave movement $p(\tilde{\boldsymbol{x}}_s, \dot{\boldsymbol{x}}_s)$
where $\tilde{\boldsymbol{x}}_s = \boldsymbol{x}_s - \boldsymbol{x}_s^d$ and $\boldsymbol{x}_s^d$ is the desired state of the slave. To ensure the stability of the system, SEDS is used to model these three joint distributions [Khansari-Zadeh and Billard, 2011]. The function $\Phi(\cdot)$ maps $\boldsymbol{x}_m$ to the same dimensionality of $\boldsymbol{x}_s$. This mapping is necessary because SEDS can handle only models in which the inputs and outputs have the same dimensionality [Shukla and Billard, 2012].

The reproduction of learned motions is performed by repeating three steps: First, the movement of the master is planned using $p(\boldsymbol{x}_m, \dot{\boldsymbol{x}}_m)$. Subsequently, the state of the slave is estimated based on $p\left(\boldsymbol{x}_s^d | \Phi(\boldsymbol{x}_m)\right)$. Third, the motion of the slave is planned based on $p(\boldsymbol{x}_s, \dot{\boldsymbol{x}}_s)$. These steps are repeated until the system converges to the goal position. The CDS approach has been applied to learn the correlation between the arm and fingers [Shukla and Billard, 2012, Kim et al., 2014], or the eye and arm [Lukic et al., 2014].

### 3.5.7 Incremental Trajectory Learning

Demonstrations by human experts are not always optimal for the learner, and the performance of the learner can be unsatisfactory after learning from demonstrations. In such cases, corrective actions can be used to improve the performance of the learner.

The study by Calinon and Billard [2007] extended the framework of statistical trajectory learning in [Calinon et al., 2007] to incremen-

---

**Algorithm 7** Incremental gesture learning [Calinon and Billard, 2007]
  **repeat**
      Record the demonstrated trajectories
      Project demonstrated trajectories onto the latent space with PCA
      Recognize the motion
      Train GMMs
      Plan a trajectory in the latent space using the updated GMMs
      Re-project the planned trajectory onto the joint space
      Execute/simulate the trajectory
  **until** task learned

---

tal learning. In [Calinon and Billard, 2007], GMMs are initialized with trajectories demonstrated by a human wearing a motion sensor. Subsequently, the motion of the humanoid robot is modified through kinesthetic teaching by a human coach. Through this iterative process, the model of the trajectory distribution is improved incrementally. The method in [Calinon and Billard, 2007] is summarized in Algorithm 7. The method in [Lee and Ott, 2011] used a similar representation by combining GMMs with HMMs. In the framework of [Lee and Ott, 2011], the compliance of a robot manipulator is controlled in order to represent an area where motion refinement is allowed. However, the method in [Calinon and Billard, 2007] does not address the context of the task. Therefore, the generalization of the demonstrated trajectories to new situations is not concerned. Recent follow-up work [Havoutis and Calinon, 2017] addressed the online learning and the adaptation of the skill to new contexts by combining an optimal control approach and TP-GMM in [Calinon, 2015].

Ewerton et al. [2016] used ProMPs for incremental imitation with generalization to different contexts. Ewerton et al. [2016] parameterizes trajectories with ProMPs as $p(\boldsymbol{\tau}|\boldsymbol{w})$. To generalize the demonstrated trajectories to new contexts, the joint distribution of trajectory parameters and the Gaussian context $p(\boldsymbol{w}, \boldsymbol{s})$ is incrementally learned under the supervision of a human. Given a new context $\boldsymbol{s}^{\text{new}}$, the trajectory is planned as a conditional distribution $p(\boldsymbol{\tau}|\boldsymbol{s}^{\text{new}})$. The method in [Ewerton et al., 2016] which is suitable for incremental learning of

---

**Algorithm 8** Incremental imitation learning of context-dependent motor skills [Ewerton et al., 2016]

   **Input:** demonstrated trajectories and the contexts $\mathcal{D} = \{\boldsymbol{\tau}, \boldsymbol{s}\}$
   Initialize $p(\boldsymbol{w}, \boldsymbol{s})$ with $\mathcal{D}$
   **for** each new context $\boldsymbol{s}$ **do**
     Compute $\boldsymbol{\mu}_{\boldsymbol{w}|\boldsymbol{s}}$ and $\boldsymbol{\Sigma}_{\boldsymbol{w}|\boldsymbol{s}}$
     Compute $\boldsymbol{\mu}_{\boldsymbol{\tau}|\boldsymbol{s}}$ and $\boldsymbol{\Sigma}_{\boldsymbol{\tau}|\boldsymbol{s}}$
     **repeat**
       Plan the trajectory based on $p(\boldsymbol{\tau}|\boldsymbol{s})$
       Execute the trajectory with human intervention
       Record the context and the executed trajectory $\boldsymbol{\tau}_{\text{new}}$, $\boldsymbol{s}_{\text{new}}$
     **until** human decides to stop
     Compute the weight vector $\boldsymbol{w}_{\text{new}}$ for $\boldsymbol{\tau}_{\text{new}}$
     Update $p(\boldsymbol{w}, \boldsymbol{s})$ using $\boldsymbol{w}_{\text{new}}$ and $\boldsymbol{s}_{\text{new}}$
   **end for**

---

time-dependent trajectories is summarized in Algorithm 8. Recently, an incremental learning method which combines DMPs and Gaussian Processes (GPs) was proposed by Maeda et al. [2017]. By modeling the conditional trajectory distribution with GPs, the system can generalize the trajectories to new scenes and request additional demonstrations when the prediction uncertainty is large. In addition, the convergence to the desired point can be ensured by DMPs.

Kronander et al. [2015] proposed incremental trajectory learning using a local modulation in a time-invariant dynamical system. The concept of local modulation is applicable to various vector fields. We describe some details of the framework in the following. Let $M(\boldsymbol{x})$ be the local modulation function. The velocity for the state $\boldsymbol{x}$ is given by

$$\dot{\boldsymbol{x}}_{\text{mod}} = M(\boldsymbol{x})\dot{\boldsymbol{x}}_{\text{ini}} \qquad (3.57)$$

where $\dot{\boldsymbol{x}}_{\text{mod}}$ is the velocity with the local modulation and $\dot{\boldsymbol{x}}_{\text{ini}}$ is the velocity given by the initial dynamical system. The local modulation is represented by scaling and rotation of the original dynamics in the framework of [Kronander et al., 2015]. Therefore, the modulation func-

tion is given by

$$M(\boldsymbol{x}) = (1 + \kappa(\boldsymbol{x}))R(\boldsymbol{x}) \qquad (3.58)$$

where $\kappa$ is a scaling factor and $R$ is a rotation matrix. For 2D motion $R$ is parameterized by a rotation angle $\phi$. For 3D motion $R$ is parameterized by a rotation angle $\phi$ and the rotation vector $\boldsymbol{\mu}_R$. When local additional demonstrations are given, the nonlinear local dynamics is modeled with a GP.

While a GP was used to model the local modulation, the framework in [Kronander et al., 2015] is not limited to a specific regression method. For movement which can be represented as a vector field, the method in [Kronander et al., 2015] is considered a reasonable option for incremental learning.

### 3.5.8 Combining Multiple Expert Policies

When multiple movement primitives can be learned, it is possible to combine movement primitives to generalize them to new situations. Jacobs et al. [1991] proposed the concept of *mixture of experts*, which generates a policy by mixing multiple experts' policies. Given multiple experts' policies $\{\pi_i\}_{i=1}^{M}$, the policy can be obtained as a mixture of these policies

$$\pi(\boldsymbol{x}) = \frac{\sum_{i=1}^{M} o_i \pi_i(\boldsymbol{x})}{\sum_{i=1}^{M} o_i}, \qquad (3.59)$$

where $o_i$ is the weight on each expert policy.

Another way of combining multiple experts' policies is *products of experts* proposed by Hinton [2002]. The policy can be obtained as a product of multiple experts' policies

$$\pi(\boldsymbol{x}) = \frac{\prod_{i=1}^{M} \pi_i(\boldsymbol{x})}{\int \prod_{i=1}^{M} \pi_i(\boldsymbol{x})d\boldsymbol{x}}. \qquad (3.60)$$

In imitation learning literature, the concept of mixture of experts has been applied to multiple DMPs [Mülling et al., 2013]. Mülling et al. [2013] learned a library of DMP based movement primitives for hitting a table tennis ball. In [Mülling et al., 2013], given a new ball coming, a mixture of learned policies generates a striking movement. In addition to initializing policies by learning from demonstration, Mülling et al. [2013] used a reinforcement learning method to improve the performance.

Likewise, Ewerton et al. [2015] learned human-robot collaborative motions as a mixture of ProMPs. Ewerton et al. [2015] learned various interaction patterns as Gaussian Mixture models of ProMP weight vectors. This method can also be interpreted as a variant of mixture of experts.

Haruno et al. [2001] proposed the modular selection and identification for control (MOSAIC) model, which learns multiple modules of forward and inverse dynamics models. In the MOSAIC model, each module learns local models, and the control input is determined by a mixture of multiple modules.

Although the concept of products of experts has been used in reinforcement learning, it has not been popular in imitation learning so far. An interesting direction of future work could be using products of experts for combining multiple expert policies in imitation learning.

## 3.6 Model-Free Behavioral Cloning for Task-Level Planning

When a task requires a complex motion, it is often necessary to plan the motion as a sequence of primitive motions. This kind of high level motion planning is known as task-level planning [Lozano-Perez et al., 1989, Ekvall and Kragic, 2008, Cambon et al., 2009, Lagriffoul et al., 2014]. In this section, we review model-free behavioral cloning methods for task-level planning.

### 3.6.1 Segmentation and Clustering for Task-Level Planning

Although model-free methods for trajectory learning often implicitly assume that each demonstrated trajectory contains a single motion, a demonstrated trajectory may consist of a sequence of different types of primitive motions in practice. Therefore, in order to learn each primitive motion, it is necessary to segment the demonstrated trajectory.

In addition, after the segmentation of trajectories, it is often necessary to cluster the segmented motions in order to learn multiple types of primitive motions. However, manual segmentation and clustering of trajectories is often time-consuming. For this reason, methods for segmenting and clustering the demonstrated trajectories have been investigated in the field of imitation learning. The development of methods for trajectory segmentation is closely related to the theoretical advances in clustering in machine learning. Although theories for segmentation and clustering are out of our scope, we shortly review methods for segmentation and clustering in imitation learning.

Kohlmorgen and Lemm [2001] developed an online segmentation method based on HMMs. By computing the "distance" between nearby data windows, Kohlmorgen and Lemm [2001] segments human motion data using unsupervised learning. Kulić et al. [2008] proposed a method for segmenting and clustering whole body motions by using factorized HMMs. In their method, the distances between HMMs are computed, and segments of the observed motion are clustered into a tree structure. Fearnhead and Liu [2007] proposed an online direct simulation algorithm for online inference in change-point problems (problems where the probability distribution changes at "change-points"). Konidaris et al. [2011] extended the approach in [Fearnhead and Liu, 2007] to learning skill trees. The beta process autoregressive HMM (BP-AR-HMM) developed by Fox et al. [2009] is a Bayesian nonparametric approach, which finds dynamic features in time-series data. The BP-AR-HMM is also employed by Niekum et al. [2014] for learning primitive motion sequences in robotics. As seen from these previous studies, advances in trajectory segmentation in imitation learning [Kulić et al., 2008, Konidaris et al., 2011, Niekum et al., 2014] are closely related to the methodological advances [Fearnhead and Liu, 2007, Fox et al., 2009] in the machine learning community.

### 3.6.2 Learning a Sequence of Primitive Motions

For learning a sequence of primitive motions, it is necessary to model the structure of the skill and learn the transition between primitive motions from the demonstrated behavior.

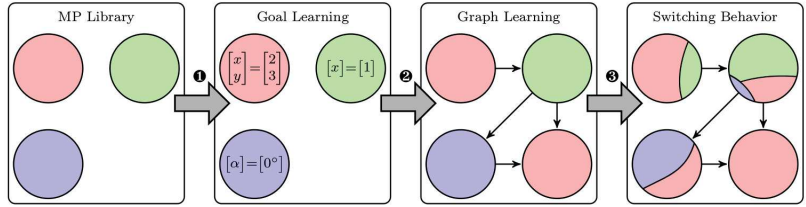

**Figure 3.9:** Learning a motion sequence in [Manschitz et al., 2015]. A library of movement primitives are learned from demonstrations, and transitions between movement primitives are modeled using SVMs.

One way of learning a sequence of movement primitives is to learn a tree-like structure of skills. Konidaris et al. [2011] proposed an online algorithm for constructing skill trees from demonstrations. Based on change point detection using MAP estimation [Fearnhead and Liu, 2007], a demonstrated trajectory is segmented into a skill chain. Multiple skill chains are merged into a skill tree by identifying similar skills in different skill chains. The method in [Konidaris et al., 2011] has been applied to path planning of a mobile robot.

Another way to sequence movements is to learn a transition model between different movement primitives. Manschitz et al. [2015] learns a library of movement primitives and uses a support vector machine (SVM) to compute the solution to the multi-class classification problem of choosing the next movement primitive for each current movement primitive. This results in a movement primitive graph structure as shown in Figure 3.9.

For learning a probabilistic transition model between movement primitives, HMM-based methods are often used. In the autoregressive hidden Markov Model (STARHMM) [Kroemer et al., 2014] the probability distribution over latent variables also depends on the observed state contrary to the classical auto-regressive hidden Markov model (AR-HMM) where the current state depends only on the previous state. STARHMM includes a latent phase variable that defines the current phase of the task. The framework in [Kroemer et al., 2015] uses STARHMM to represent a task as a sequence of DMPs [Ijspeert et al., 2013], where the phase variable corresponds to the currently active DMP. The model allows for a conditional movement primitive

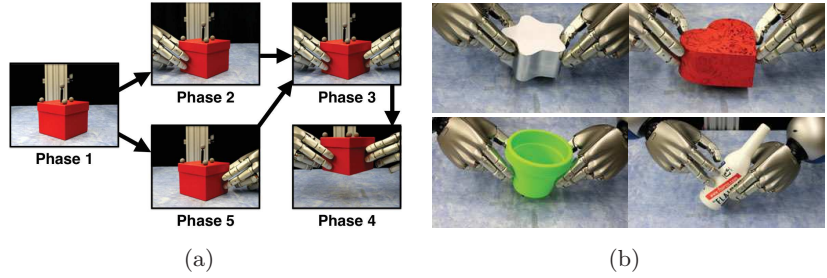

**Figure 3.10:** Learning a hierarchical skill in [Kroemer et al., 2015]. Left: A sequence of skills are modeled using a variant of HMM. Right: The learned DMPs can be adapted to different objects.

---

**Algorithm 9** Incremental semantically grounded learning from demonstration [Niekum et al., 2014]
---

**Input:** Demonstrated trajectories and object poses $\mathcal{D} = \{\boldsymbol{\tau}^{\text{demo}}, \boldsymbol{o}\}$
Segment the demonstrations with BP-AR-HMM
**for** each segment **do**
   Learn parameters of DMP
**end for**
Construct FSM
Replay the task based on the current observation
**if** correction is necessary **then**
   Collect interactive correction from users
**end if**

---

plan that switches from one DMP to another based on the observations. Kroemer et al. [2015] learn DMPs using imitation learning and optimize high-level policies using reinforcement learning. Kroemer et al. [2015] demonstrate the approach in robotic manipulation tasks as shown in Figure 3.10.

Although it is often assumed that a sufficient amount of demonstration data is available, this may not be the case in many applications. Incremental imitation learning for task-level planning proposed by Niekum et al. [2014] can address this issue. The framework in [Niekum et al., 2014] leverages unstructured demonstrations and corrective ac-tions by human experts. Niekum et al. [2014] segment the demonstrated task using a Beta Process Autoregressive Hidden Markov Model (BP-AR-HMM) [Fox et al., 2009], and model the transition between discrete primitives as a finite-state automaton (FSA). When a new situation is given, the learner uses the trained FSA to plan the task as a sequence of movement primitives. If an expert considers that refinement of the planned motion is necessary, she/he can stop the autonomous execution of the task and correct the motion through kinesthetic teaching. In this way, the learner improves the performance through interaction with experts. Algorithm 9 summarizes the procedure.

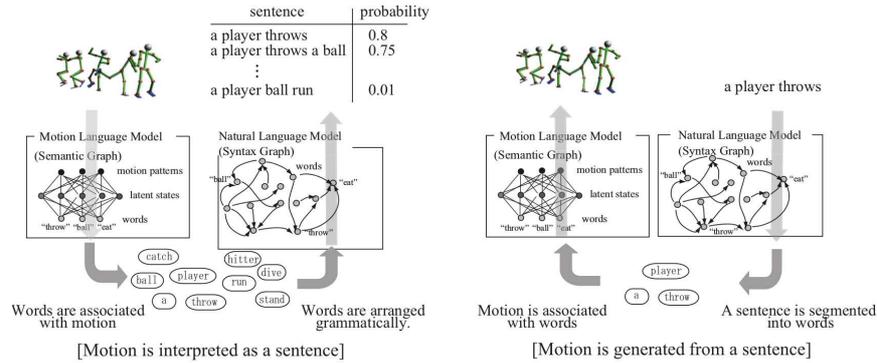

**Figure 3.11:** Mutual language model between motion and sequence in [Takano and Nakamura, 2015](Figure used with permission of Wataru Takano). Relevance between words and motion is learned using a probabilistic model. The approach can work in two directions: generating sentences from motion or generating motion from sentences. When motion is observed, a motion language semantic graph model generates words for the observed motion. A natural language model arranges the words then into sentences. When observing language the language is segmented into words using a natural language model and the words are then transformed into motion using a semantic graph.

One interesting approach for task-level planning is to leverage annotation of demonstrated motions. Recently, Takano and Nakamura [2015] developed methods for learning a mutual model between language and motions, which leverage a dataset of demonstrated motions and annotated sentences. In the framework of [Takano and Nakamura, 2015], the relationship between the motion symbols and words via latent variables is learned as a motion language model, and the sentence

---

**Algorithm 10** Motion language model [Takano and Nakamura, 2015]
___
    **Learning:**
    **Input:** demonstrated trajectories and sentences $\mathcal{D} = \{\boldsymbol{\tau}^{\mathrm{demo}}, \boldsymbol{y}\}$
    Train a set of HMMs that represent the primitive motions
    Train the motion language model and the natural language model
    **Prediction:**
    **Input:** a motion sequence or a sentence
    **if** the given input is a motion sequence **then**
        Recognize the motion symbol $\lambda^{in}$ using HMMs
        Predict words for the given motion
          $\boldsymbol{y}^* = \arg\max_{\boldsymbol{y}} p(\boldsymbol{y}|\lambda^{in})$
        Arrange the order of the words using the natural language model
        **return** sentence
    **end if**
    **if** the given input is a sentence **then**
        Predict a motion symbol corresponding to the given sentence $\boldsymbol{y}^{\mathrm{in}}$
          $\lambda^* = \arg\max_{\lambda \in \Lambda} p(\lambda|\boldsymbol{y}^{\mathrm{in}})$
        Predict the motion sequence from the motion symbol $\lambda^*$
        **return** motion sequence
    **end if**
___

structure is learned as a natural language model using an n-gram model. Figure 3.11 summarizes the framework of a mutual model between language and motion. HMMs are used to represent primitive motions, and the library of primitive motions are learned as a set of HMMs. In the motion language model, the probability $p(\lambda|\boldsymbol{y})$ and $p(\boldsymbol{y}|\lambda)$ are learned, where $\boldsymbol{y}$ is an annotated sentence and $\lambda$ is the motion symbol. This motion language model can be learned using an EM algorithm. Meanwhile, a natural language model learns the transition between two words $p(\boldsymbol{y}_i|\boldsymbol{y}_j)$. When a new motion $\boldsymbol{\tau}^{in}$ is observed, the corresponding motion symbol $\lambda^{in}$ is predicted using HMMs. Subsequently, words associated with the motion symbol are estimated as

$$\boldsymbol{y}^* = \arg\max_{\boldsymbol{y}} p(\boldsymbol{y}|\lambda^{in}), \tag{3.61}$$

where $\lambda^{\mathrm{in}}$ is the recognized motion symbol. Thereafter, the estimated words are arranged grammatically using the natural language model.

When a new sentence $\boldsymbol{y}^{\text{in}}$ is given, the motion symbol is selected so as to maximize the likelihood of observing $\boldsymbol{y}^{\text{in}}$

$$\lambda^* = \arg\max_{\lambda \in \Lambda} p(\lambda|\boldsymbol{y}^{\text{in}}), \qquad (3.62)$$

where $\Lambda$ is a set of learned motion symbols, and $\lambda^*$ is the predicted motion symbol. A motion sequence is then generated using the predicted motion symbol. The method is summarized in Algorithm 10. Leveraging the mutual model between language and motion will be an interesting research direction in imitation learning.

## 3.7 Model-Based Behavioral Cloning Methods

We discuss model-based behavioral cloning (BC) in this section. As we discussed in §2.3, model-based BC methods require an iterative learning process with access to a forward dynamics model. Next, we discuss model-based BC in more detail.

### 3.7.1 Model-Based Behavioral Cloning Methods with Forward Dynamics Models

In imitation learning, experts demonstrate behavior and an autonomous agent tries to imitate the demonstrations. However, the embodiment of the expert is often different from the embodiment of the learner. In such cases, the demonstrated trajectory needs to be adjusted for the embodiment of the learner. Otherwise, the learner fails to perform the intended task properly. This problem is known as the "*correspondence problem*" in imitation learning [Billard et al., 2008]. The correspondence problem frequently appears when we try to teach humanoids how to imitate human motion obtained e.g. from motion trackers [Ude et al., 2004, Nakaoka et al., 2007]. Due to the different embodiments between a human expert and a robot learner, it is essential to adapt the demonstrated trajectories to follow the constraints and dynamics of the learner.

Even when the embodiments of the demonstrator and learner match, we may face a similar correspondence problem when we try to execute a trajectory at a velocity differing from the original velocity [van den Berg et al., 2010, Englert et al., 2013]. Even if the desired configuration is kinematically feasible, the demonstrated/desired velocity may be infeasible due to the underactuation of the manipulator. In this case, it is also necessary to adjust the planned trajectory based on the system dynamics.

The straightforward way for solving the correspondence problem is to explicitly learn a forward dynamics model of the system

$$\boldsymbol{x}_{t+1} = f(\boldsymbol{x}_t, \boldsymbol{u}_t) \tag{3.63}$$

and then plan trajectories based on the learned forward model. Forward dynamics model learning can be framed as a regression problem. Table 3.6 lists different regression methods which have been utilized in model-based BC. Although locally weighted regression and Gaussian mixture regression were used in early studies of model-based methods, recent studies often employ Gaussian Processes. As we will review in §3.7.1.2, Gaussian Processes can incorporate inputs with uncertainty. This property is important for multi-step forward prediction since the uncertainty is propagated over time. However, due to the computational cost, Gaussian Process regression is not suitable for high-dimensional data. To deal with high-dimensional data such as raw images, a deep learning approach is employed for modeling a forward dynamics in the most recent studies [Oh et al., 2015, Finn et al., 2017a, Baram et al., 2017, Nair et al., 2017]. In the following sections, we review some of the model-based methods with explicit learning of a forward model.

**Table 3.6:** Model-based behavioral cloning methods using different regression methods. Early studies on model-based behavior cloning focused on locally weighted regression but later studies have moved to Gaussian mixture regression and even more recently to Gaussian processes. We expect that studies based on deep neural networks will be popular in the near future.

| Regression | Employed by ... |
|---|---|
| *Locally Weighted Regression* | [Atkeson et al., 1997, Schneider, 1997] |
| *Gaussian Mixture Regression* | [Grimes et al., 2006b, Grimes and Rao, 2009] |
| *Gaussian Process Regression* | [Grimes et al., 2006a, Englert et al., 2013, Deisenroth et al., 2014] |
| *Neural Networks* | [Baram et al., 2017, Nair et al., 2017] |

### 3.7.1.1 Imitation with a Gaussian Mixture Forward Model

We will now discuss details of the methods in [Grimes et al., 2006b, Grimes and Rao, 2009] as an example of learning forward dynamics with Gaussian Mixture Models (GMMs).

We can obtain a dataset of state $\boldsymbol{x}_t$ and action $\boldsymbol{u}_t$ trajectories $\mathcal{D} = \{\boldsymbol{\tau}^i = [\boldsymbol{x}_1^i, \boldsymbol{u}_1^i \cdots, \boldsymbol{x}_T^i, \boldsymbol{u}_T^i]\}_{i=1}^M$ from sensor readings. If we introduce $\boldsymbol{z}_t = [\boldsymbol{x}_t, \boldsymbol{u}_t]$, the joint distribution of $\boldsymbol{x}_{t+1}$ and $\boldsymbol{z}_t$ can be modeled as a mixture of Gaussian distributions as

$$p(\boldsymbol{x}_{t+1}, \boldsymbol{z}_t) = \sum_k p(k)\mathcal{N}(\boldsymbol{\mu}_k, \boldsymbol{\Sigma}_k), \tag{3.64}$$

where $p(k)$ is the prior and the $k$th Gaussian component is given by

$$p(\boldsymbol{x}_{t+1}, \boldsymbol{z}_t|k) = \mathcal{N}\left(\left[\begin{array}{c}\boldsymbol{z}_t \\ \boldsymbol{x}_{t+1}\end{array}\right] \middle| \left[\begin{array}{c}\boldsymbol{\mu}_{z,k} \\ \boldsymbol{\mu}_{x,k}\end{array}\right], \left[\begin{array}{cc}\boldsymbol{\Sigma}_{z,k} & \boldsymbol{\Sigma}_{zx,k} \\ \boldsymbol{\Sigma}_{xz,k} & \boldsymbol{\Sigma}_{x,k}\end{array}\right]\right). \tag{3.65}$$

The conditional distribution of $\boldsymbol{x}_{t+1}$ for a given $\boldsymbol{z}_t^*$ is a Gaussian distribution with the mean and variance given by

$$\begin{aligned}\boldsymbol{\mu}_{x|z} &= \sum w_k \boldsymbol{\mu}_{x|z,k}, \\ \boldsymbol{\Sigma}_{x|z} &= \sum_{k=1}^K w_k \left(\boldsymbol{\Sigma}_{x|z,k} + \boldsymbol{\mu}_{x|z,k}\boldsymbol{\mu}_{x|z,k}^\top\right) - \boldsymbol{\mu}_{x|z}\boldsymbol{\mu}_{x|z}^\top,\end{aligned} \tag{3.66}$$

where

$$\begin{aligned}\boldsymbol{\mu}_{x|z,k} &= \boldsymbol{\mu}_{x,k} + \boldsymbol{\Sigma}_{xz,k}(\boldsymbol{\Sigma}_{z,k})^{-1}(\boldsymbol{z}_t^* - \boldsymbol{\mu}_{z,k}), \\ \boldsymbol{\Sigma}_{x|z,k} &= \boldsymbol{\Sigma}_{x,k} - \boldsymbol{\Sigma}_{xz,k}(\boldsymbol{\Sigma}_{z,k})^{-1}\boldsymbol{\Sigma}_{zx,k}, \\ w_k &= \frac{p(k)\mathcal{N}(\boldsymbol{z}_t^*|\boldsymbol{\mu}_{z,k}, \boldsymbol{\Sigma}_{z,k})}{\sum_{k=1}^K p(k)\mathcal{N}(\boldsymbol{z}_t^*|\boldsymbol{\mu}_{z,k}, \boldsymbol{\Sigma}_{z,k})}.\end{aligned} \tag{3.67}$$

When a given input is drawn from a Gaussian distribution $\boldsymbol{z}_t^* \sim \mathcal{N}(\boldsymbol{\mu}^{\text{in}}, \boldsymbol{\Sigma}^{\text{in}})$, the marginal distribution $p(\boldsymbol{x}_{t+1}|\boldsymbol{\mu}^{\text{in}}, \boldsymbol{\Sigma}^{\text{in}})$ is a Gaussian distribution with the mean $\boldsymbol{\mu}_{t+1}$ and covariance $\boldsymbol{\Sigma}_{t+1}$ given by

$$\begin{aligned}\boldsymbol{\mu}_{x|z} &= \sum w_k \boldsymbol{\mu}_{x|z,k}, \\ \boldsymbol{\Sigma}_{x|z} &= \sum_{k=1}^K w_k \left(\boldsymbol{\Sigma}_{x|z,k} + \boldsymbol{\mu}_{x|z,k}\boldsymbol{\mu}_{x|z,k}^\top\right) - \boldsymbol{\mu}_{x|z}\boldsymbol{\mu}_{x|z}^\top,\end{aligned} \tag{3.68}$$

**Algorithm 11** Behavior acquisition via Bayesian inference and learning [Grimes and Rao, 2009]
  Observe an expert's demonstrations $[\boldsymbol{o}_1, \cdots, \boldsymbol{o}_T]$
  Estimate the kinematics of the expert
  Initialize the forward model $f$
  Infer bootstrap actions based on the forward model
  **repeat**
    Execute actions
    Learn/update the GMR forward model
    Infer constrained actions
  **until** task learned

where

$$\begin{aligned}
\boldsymbol{\mu}_{x|z,k} &= \boldsymbol{\mu}_{x,k} + \boldsymbol{\Sigma}_{xz,k} \left(\boldsymbol{\Sigma}_{z,k} + \boldsymbol{\Sigma}^{\text{in}}\right)^{-1} (\boldsymbol{z}_t^* - \boldsymbol{\mu}_{z,k}), \\
\boldsymbol{\Sigma}_{k,t+1} &= \boldsymbol{\Sigma}_{x,k} - \boldsymbol{\Sigma}_{xz,k} \left(\boldsymbol{\Sigma}_{z,k} + \boldsymbol{\Sigma}^{\text{in}}\right)^{-1} \boldsymbol{\Sigma}_{zx,k}, \\
w_k &= \frac{p(k)\mathcal{N}(\boldsymbol{z}_t^*|\boldsymbol{\mu}_{z,k}, \boldsymbol{\Sigma}_{z,k} + \boldsymbol{\Sigma}^{\text{in}})}{\sum_{k=1}^K p(k)\mathcal{N}(\boldsymbol{z}_t^*|\boldsymbol{\mu}_{z,k}, \boldsymbol{\Sigma}_{z,k} + \boldsymbol{\Sigma}^{\text{in}})}.
\end{aligned} \quad (3.69)$$

Grimes and Rao [2009] used this GMR for one-step prediction and recursively predicted learner's trajectories. Using the learned forward model, the action is selected so as to maximize the posterior likelihood as

$$\boldsymbol{u}_1^*, \cdots, \boldsymbol{u}_T^* = \arg\max_{\boldsymbol{u}_1, \cdots, \boldsymbol{u}_T} p(\boldsymbol{u}_1, \cdots, \boldsymbol{u}_T | \boldsymbol{o}_1, \cdots, \boldsymbol{o}_T, \boldsymbol{c}_1, \cdots, \boldsymbol{c}_T), \quad (3.70)$$

where $[\boldsymbol{o}_1, \cdots, \boldsymbol{o}_T]$ is a time series of the observed demonstrated states, and $[\boldsymbol{c}_1, \cdots, \boldsymbol{c}_T]$ is a time series of the feasible states of the learner under the kinematic and dynamic constraints. By repeating the execution of the planned trajectories, the estimation of the forward model improves. Algorithm 11 summarizes the procedure in [Grimes and Rao, 2009].

### 3.7.1.2  Imitation with a Gaussian Process Forward Model

Recent studies on model-based BC have employed Gaussian Processes (GPs) for modeling the forward dynamics of the system $f \sim \mathcal{GP}$ [Englert et al., 2013, Deisenroth et al., 2014]. Given a dataset $\mathcal{D} = \{x_{t+1}, z_t\}$ where $z = [x_t^\top, u_t^\top]^\top$, a GP models a mapping from the input $z_t$ to the output $x_{t+1} = f(z_t)$ as

$$f(z_t) \sim \mathcal{GP}\left(m(z_t), k(z_t, z_t')\right), \tag{3.71}$$

where $k(z, z')$ is the covariance function. A popular choice of a covariance function is the squared exponential covariance function given by

$$k(z, z') = \exp\left(-\frac{\|z - z'\|^2}{l^2}\right). \tag{3.72}$$

The joint distribution of the given target value and the function value $x_{t+1}$ at the test input $z_t^*$ can be written as

$$\begin{bmatrix} x_{t+1} \\ x_{t+1}^* \end{bmatrix} \sim \mathcal{N}\left(\mathbf{0}, \begin{bmatrix} K(Z, Z) + \sigma_n^2 I & K(Z, z_t^*) \\ K(z_t^*, Z) & K(z_t^*, z_t^*) \end{bmatrix}\right), \tag{3.73}$$

where $Z$ is a matrix in which the input vectors $z_t$ for all training samples are aggregated. The conditional distribution of $x_{t+1}^*$ given the test input $z_t^*$ is a Gaussian with mean and variance

$$\begin{aligned} \mu(z_t^*) &= k^\top K^{-1} x_{t+1}, \\ \sigma^2(z_t^*) &= k(z_t^*, z_t^*) - k^\top K^{-1} k, \end{aligned} \tag{3.74}$$

where $K = K(Z, Z) + \sigma_n^2 I$ and $k = K(z_t^*, Z)$.

As with GMR, propagation of uncertainty can be approximately modeled by GPs. If we assume that $z = [x_t^\top, u_t^\top]^\top$ is drawn from a Gaussian distribution $p(z_t|\mu_t, \Sigma_t)$, the predictive distribution of the state at time $t+1$ is given by

$$p(x_{t+1}|\mu_t, \Sigma_t) = \int p(f(z_t)|z_t, \mathcal{D}) p(z_t) dz_t, \tag{3.75}$$

where $p(f(x)|x, \mathcal{D})$ is a Gaussian distribution given by (3.74). The marginal distribution $p(x_{t+1}|\mu_t, \Sigma_t)$ in (3.75) can be approximated by a Gaussian distribution by following the results from [Deisenroth and Rasmussen, 2011, Deisenroth et al., 2013a].

Englert et al. [2013] used GPs for predicting the trajectory distribution, and the KL divergence was used to evaluate the similarity of

---

**Algorithm 12** Probabilistic model-based imitation learning [Englert et al., 2013]

**Input:** $n$ trajectories $\boldsymbol{\tau}_i$ demonstrated by the expert
Estimate the expert distribution over trajectories $q(\boldsymbol{\tau}^{\text{demo}})$
Record state-action parts of the robot through applying random control inputs
**repeat** $i = 1$ **to** $N$ **do**
   Learn/update probabilistic GP forward model
   Predict the new trajectory distribution $p(\boldsymbol{\tau})$
   Learn policy $\pi^{\text{L}} = \arg\min_\pi D_{\text{KL}}\left(q(\boldsymbol{\tau}^{\text{demo}})||p(\boldsymbol{\tau})\right)$
   Apply $\pi^{\text{L}}$ to the system and record data
**until** task learned

---

the demonstrated and learned behaviors. Englert et al. [2013] modeled trajectories as a Gaussian distribution

$$p(\boldsymbol{\tau}) \sim \prod_{t=1}^{T} p(\boldsymbol{x}(t)) = \prod_{t=1}^{T} \mathcal{N}(\boldsymbol{x}(t)|\boldsymbol{\mu}(t), \boldsymbol{\Sigma}(t)). \qquad (3.76)$$

For two given Gaussian distributions $p(\boldsymbol{x}(t)) \sim \mathcal{N}(\boldsymbol{x}|\boldsymbol{\mu}_p(t), \boldsymbol{\Sigma}_p(t))$ and $q(\boldsymbol{x}(t)) \sim \mathcal{N}(\boldsymbol{x}|\boldsymbol{\mu}_q(t), \boldsymbol{\Sigma}_q(t))$, the KL divergence of $q$ and $p$ can be computed in closed form. Using the factorization in (3.76), the KL divergence between the trajectory distribution induced by the expert policy $q(\boldsymbol{\tau})$ and the trajectory distribution induced by the learned policy $p(\boldsymbol{\tau})$ can be computed as

$$D_{\text{KL}}\left(q(\boldsymbol{\tau})||p(\boldsymbol{\tau})\right) = \sum_{t=1}^{T} D_{\text{KL}}(q(\boldsymbol{x}(t))||p(\boldsymbol{x}(t))). \qquad (3.77)$$

Englert et al. [2013] used this KL divergence to define the objective function to be minimized as $\mathcal{L}_{\text{KL}} = D_{\text{KL}}\left(q(\boldsymbol{\tau})||p(\boldsymbol{\tau})\right)$. To minimize $\mathcal{L}_{\text{KL}}$ we can compute the gradient analytically and use gradient descent [Deisenroth, 2010, Deisenroth and Rasmussen, 2011]. Algorithm 12 summarizes the procedure in [Englert et al., 2013]. The method in [Englert et al., 2013] matches the first and second moment of the trajectory distribution through iterative learning. Since the deriva-

---

**Algorithm 13** Iterative control learning [van den Berg et al., 2010]

   **Input:** desired trajectory $\boldsymbol{\tau}^d$, learning rate $\alpha$
   Initialize the target trajectory as $\boldsymbol{\tau} = \boldsymbol{\tau}^d$
   **repeat**
      Execute a controller with the target trajectory $\hat{\boldsymbol{\tau}}$
      Record the executed trajectory $\boldsymbol{\tau}$
      Update the target trajectory $\hat{\boldsymbol{\tau}} \leftarrow \hat{\boldsymbol{\tau}} - \alpha(\boldsymbol{\tau} - \boldsymbol{\tau}^d)$
   **until** $\boldsymbol{\tau} \approx \boldsymbol{\tau}^d$

---

tives can be analytically computed when using a GP forward dynamics model, imitation learning can be efficiently performed.

### 3.7.2 Imitation Learning through Iterative Learning Control

In order to develop a controller to achieve the desired trajectory, we can also use an iterative learning control approach without a forward dynamics model. Abbeel et al. [2010], van den Berg et al. [2010] learn a controller iteratively to reproduce a desired trajectory.

While van den Berg et al. [2010] uses a Linear Quadratic Regulator (LQR) [Anderson and Moore, 1990] for optimal control, the method is not limited to a specific controller. Algorithm 13 shows how iterative control learning in [van den Berg et al., 2010] works. Given a desired trajectory $\boldsymbol{\tau}^d$, LQR control is performed to track the target trajectory $\hat{\boldsymbol{\tau}}$. In the initial step, the target trajectory is initialized as $\hat{\boldsymbol{\tau}} = \boldsymbol{\tau}^d$. When the executed trajectory $\boldsymbol{\tau}$ deviates from the desired trajectory $\boldsymbol{\tau}^d$, the approach updates the target trajectory as $\hat{\boldsymbol{\tau}} \leftarrow \hat{\boldsymbol{\tau}} - \alpha(\boldsymbol{\tau} - \boldsymbol{\tau}^d)$, where $\alpha$ is the learning rate. By repeating this execution and update, a target trajectory $\boldsymbol{\tau} \approx \boldsymbol{\tau}^*$ can be obtained. Although this method is simple and easy to implement, the controller cannot be generalized to different desired trajectories.

When a given system is fully controllable, we can learn forward and inverse dynamics of the system. As indicated by [Nguyen-Tuong and Peters, 2011], various methods have been developed for model learning. However, it is often challenging to apply such approaches to not fully controllable systems. Iterative LQR (iLQR) is often employed to con-trol a system of which the dynamics is not accurately known [Todorov and Li, 2005, Abbeel et al., 2010, Tassa et al., 2012]. iLQR learns a linear feedback controller to follow a trajectory through an iterative learning process. Abbeel et al. [2010] learns from experts' demonstrations trajectories for acrobatic RC helicopter flights, and utilizes iLQR to reproduce the desired trajectory.

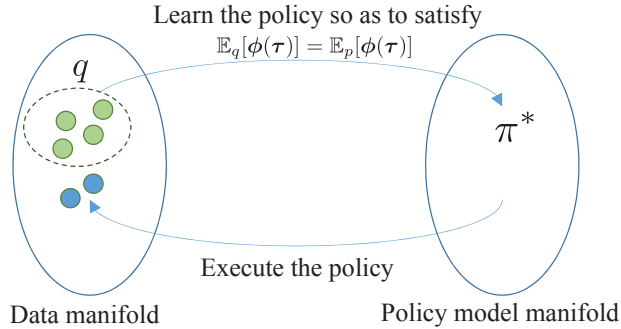

**Figure 3.12:** Schematic illustration of model-based BC methods. Model-based BC methods often iterate between policy updates and task execution so as to match the expected features as $\mathbb{E}_q[\boldsymbol{\phi}] \simeq \mathbb{E}_p[\boldsymbol{\phi}]$.

### 3.7.3 Information Theoretic Understandings of Model-Based Behavioral Cloning Methods

BC methods with forward dynamics models such as [Englert et al., 2013, Grimes and Rao, 2009] iteratively evaluate the learned policy $\pi^{\mathrm{L}}(\boldsymbol{u}|\boldsymbol{x})$ in order to reproduce trajectories close to the demonstrations. These methods evaluate the trajectory under the distribution induced by the learned policy and match its expected feature with that of the expert demonstrations. This approach can be interpreted as a process to empirically learn the policy $\pi^{\mathrm{L}}(\boldsymbol{u}|\boldsymbol{x})$ that satisfies

$$\mathbb{E}_p[\boldsymbol{\phi}(\boldsymbol{\tau})] \simeq \mathbb{E}_q[\boldsymbol{\phi}(\boldsymbol{\tau})], \quad (3.78)$$

where $q(\boldsymbol{\tau})$ is the expert trajectory distribution and $p(\boldsymbol{\tau})$ is the trajectory distribution induced by the learner's policy. The learning process of BC methods with forward dynamics can be illustrated as Figure 3.12. In addition, the method in [Englert et al., 2013] assumes that the tra-jectory distribution is Gaussian. As Park and Bera [2009] indicated, Gaussian distribution is one of the maximum entropy distributions. Therefore, matching the feature expectation as in (3.78) under the assumption of the Gaussian distribution can be interpreted as the M-projection onto the manifold of the maximum entropy distribution as we discussed in §2.7.1.

## 3.8 Robot Applications with Model-Free BC Methods

Robot Applications with Model-Free Behavioral Cloning Methods In this section, we show several examples of model-free behavioral cloning (BC) in robotic applications, to demonstrate the capability of model-free BC methods. Model-free BC methods have been utilized successfully in various applications, including autonomous RC helicopter flight, ball-hitting tasks, and robotic surgery. Abbeel et al. [2010] uses an iterative LQR controller in acrobatic helicopter flight to control the nonlinear system. [Osa et al., 2017b] performs knot-tying tasks using a standard PD controller on a surgical robot. From the following application examples, one can see that different applications require different controllers and learning methods.

### 3.8.1 Learning to Hit a Ball with DMP

Hitting a ball is a typical example of tasks that can be learned as a point-to-point motion. Ijspeert et al. [2002b] showed that a tennis swing can be learned with DMPs. The motion of a tennis swing was demonstrated by a human, and the motion was recorded using a motion capture suit, which can mechanically measure the joint angles of 35 DoFs of the human body at 100Hz. The recorded motion was reproduced in a humanoid robot with 30 DoFs. To accurately reproduce the trajectories, an inverse dynamics controller was employed in this experiment. The experimental results showed that the learned motion was generalized to different target positions.

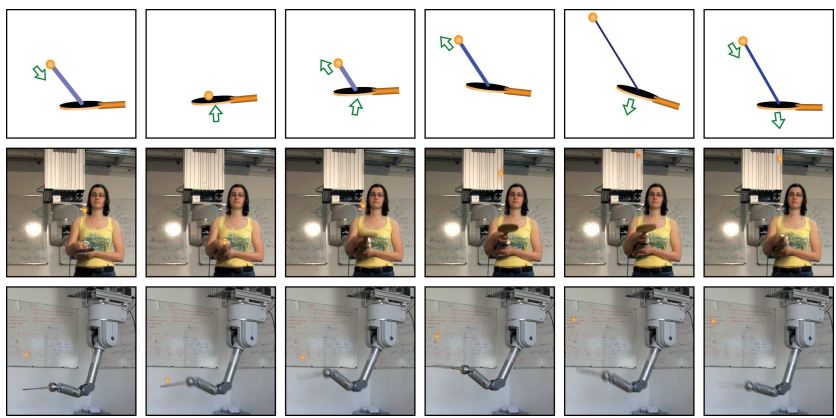

**Figure 3.13:** Learning rhythmic motions for the Ball-Paddling task in [Kober and Peters, 2009]. Kober and Peters [2009] used kinesthetic teaching to demonstrate periodic hitting motions in Ball-Paddling and trained rhythmic DMPs to reproduce the demonstrated periodic movements.

Kober and Peters [2009] learned a Ball-Paddling task shown in Figure 3.13 from demonstrations. The goal of this task is to have the ball repeatedly bouncing. Kober and Peters [2009] used the seven degrees of freedom Barrett WAM arm to demonstrate trajectories using kinesthetic teaching and learned periodic motion using rhythmic DMPs. In the experiments, ten basis functions per motor primitive were used to represent the task.

### 3.8.2 Learning Hand-Over Tasks with ProMPs

Motion planning in the context of human-robot collaboration often requires learning the coupled motions of a human operator and a robot. Maeda et al. [2016] shows that correlation of the two agents' motion can be modeled using ProMPs. Maeda et al. [2016] illustrates the approach in a hand-over motion: when a human extends her/his hand to receive a plate or screw, the robot grasps and gives it to the human operator. Maeda et al. [2016] used a KUKA LWR robot and kinesthetic teaching for demonstrating tasks, and the motion of a human operator was tracked using a 3D optical tracking system. The task was demonstrated 13-20 times. Demonstrated trajectories are shown

(b)

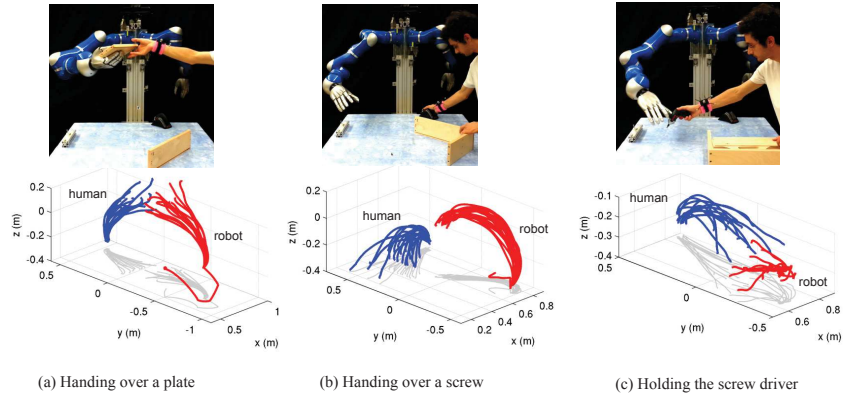

(a) Handing over a plate  (b) Handing over a screw  (c) Holding the screw driver

**Figure 3.14:** Learning human-robot collaborative motions in [Maeda et al., 2016]. Maeda et al. [2016] used kinesthetic teaching to demonstrate coupled movements, where both the human and robot need to move to perform a task. The demonstrations were used to train interaction ProMPs which take correlations between human and robot movement into account: the robot motion can be planned as conditional distribution given the human movement. The pictures show how the robot is able to adapt its movement in several tasks.

in Figure 3.14. The correlation of the robot's motion and the human operator's motion was learned with interaction ProMPs, which is an extension of ProMPs proposed by Paraschos et al. [2013]. To achieve the human-robot collaborative task, the robot motion was planned by conditioning the learned distribution on the observed motion of the human operator. Maeda et al. [2016] applied interaction ProMPs to several tasks as shown in Figure 3.14. The study by Maeda et al. [2016] showed that the reactive motions of the robot were successfully planned based on the observed motions of the human operator.

Recent work by Lioutikov et al. [2017] proposed a method for segmenting demonstrated trajectory in a probabilistic manner and learning a sequence of movement primitives represented by ProMPs. Tasks that emulate table tennis, writing and chair assembly are reported in [Lioutikov et al., 2017].

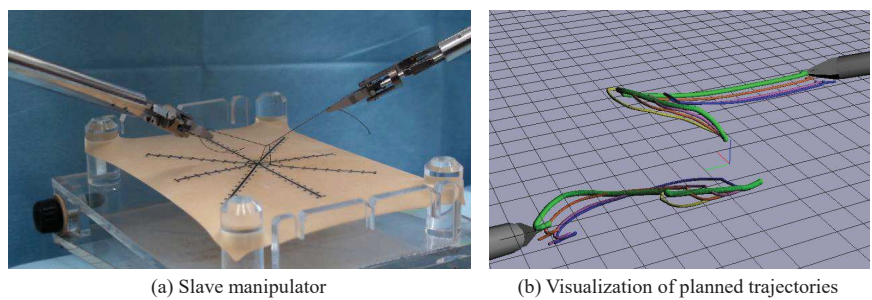

(a) Slave manipulator  (b) Visualization of planned trajectories

**Figure 3.15:** Autonomous knot-tying with a surgical robot [Osa et al., 2017b]. Left: Bimanual manipulation tasks were learned using a model-free BC method. Right: The trajectories can be updated in real time when the context is changing during task execution. The demonstration was performed under various contexts, and the trajectory distribution was modeled using a Gaussian Process. A force controller was build as an outer loop of the standard PD position controller.

### 3.8.3 Learning to Tie a Knot by Modeling the Trajectory Distribution with Gaussian Processes

Knot-tying in robotic surgery is one of the tasks that is hard to learn as a sequence of point-to-point motions. In a looping motion required for the knot-tying task, the topological shape of the entire trajectory is critical, although the start and goal positions of the trajectory is not critical to the success of the task. Osa et al. [2017b] applied a behavioral cloning method to this knot-tying task as shown in Figure 3.15. Osa et al. [2017b] learned a conditional distribution of the demonstrated trajectories given the context as a Gaussian Process allowing generalizing demonstrated trajectories to a new context in real time. Additionally, the learned trajectory distribution was used to plan and control the contact force between the surgical instruments and objects. Osa et al. [2014] employed Algorithm 6 for normalizing the time alignment of multiple demonstrated trajectories.

In experiments with a bimanual teleoperated master-slave system for robotic surgery shown in Figure 3.15, the system performed tasks that emulate tying a knot and cutting soft tissues. The task was demonstrated 9-20 times under various contexts. The experimental results show that the trajectories can be updated in real time.

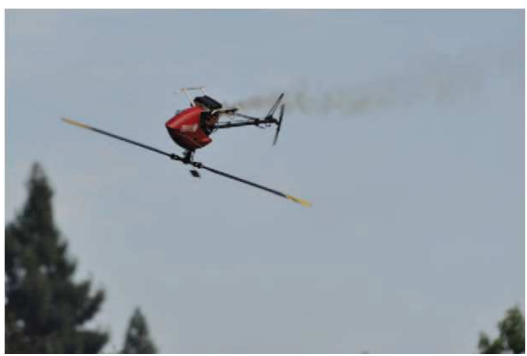

**Figure 3.16:** Learning autonomous helicopter maneuvers from expert demonstrations in [Abbeel et al., 2010]. Acrobatic flights were learned in a system that involves highly nonlinear dynamics. An iterative LQR controller is employed to execute the trajectory learned from demonstrations.

## 3.9 Robot Applications with Model-Based Behavioral Cloning Methods

We present applications of model-based BC methods in this section. Model-based BC methods can be used to control robotic systems with nonlinear dynamics. A remarkable application example of model-based BC methods is acrobatic helicopter flights [Abbeel et al., 2010]. Additionally, we discuss an application for learning from different embodiments. Subsequently, we show applications of planning in action-state space.

### 3.9.1 Learning Acrobatic Helicopter Flights

Autonomous flight of an RC helicopter involves nonlinear dynamics, making helicopter control non-trivial. Abbeel et al. [2010] showed how to learn acrobatic RC helicopter flight from experts' demonstrations. For modeling time-dependent trajectories, Abbeel et al. [2010] normalizes the temporal alignment of the demonstrated trajectories using an Expectation Maximization (EM)-like method, which we discussed in §3.5.5. Abbeel et al. [2010] learns acrobatic flight trajectories using a model-based behavioral cloning method. Due to the challenge of controlling the highly nonlinear helicopter dynamics, Abbeel et al. [2010] uses an iterative LQR controller. In the experiments, the heli-copter control system performs various maneuvers including in-place flips, in-place rolls, loops and hurricanes, and even auto-rotation landings, chaos and tic-toc. Figure 3.16 shows a snapshot of the acrobatic flight reported in [Abbeel et al., 2010]. Previously, these acrobatic maneuvers could only be performed by exceptional experts, but Abbeel et al. [2010] showed that such expert skills can be transferred to a robotic system by combining model-based BC and iterative controller learning.

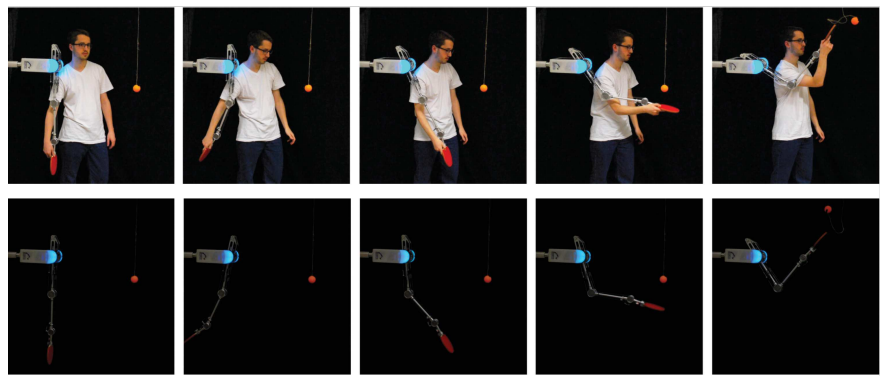

**Figure 3.17:** Learning to hit a ball with an underactuated manipulator in [Englert et al., 2013]. Englert et al. [2013] learned a forward model of the system using Gaussian Processes. Together with the forward model Englert et al. [2013] used PILCO Deisenroth and Rasmussen [2011], Deisenroth et al. [2013a] as the reinforcement learning method Englert et al. [2013] to train a policy to reproduce demonstrated trajectories.

### 3.9.2 Learning to Hit a Ball with an Underactuated Robot

Learning tasks with an underactuated robot is challenging since feasible trajectories are limited. Englert et al. [2013] learned ball hitting with an underactuated robot using a model-based imitation learning method. In the experiments, the trajectories were demonstrated by kinesthetic teaching, and the trajectory and the controller to achieve the task were learned from demonstrations. BioRob$^{TM}$ [Lens et al., 2010] robot, which is an underactuated and compliant manipulator, was used in the experiments. Figure 3.17 shows a task with the underactuated manipulator reported in [Englert et al., 2013]. Since Englert et al. [2013] learns a robot-specific controller, the controller is robust to the correspondence problem compared with model-free behavioral cloning methods. Learning a robot-specific policy is one of the benefits of model-based imitation learning. Although developing a controller for an underactuated robot with unknown nonlinear dynamics is not trivial, model-based behavioral cloning methods can address this problem by exploiting the learned forward dynamics model. This method requires an iterative learning process to obtain a policy that reproduces the expert's trajectory.

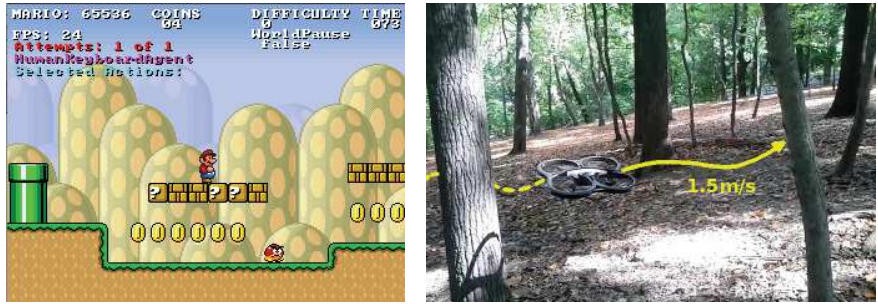

**Figure 3.18:** Applications of DAGGER [Ross et al., 2011]. Left: Learning to play a video game [Ross et al., 2011]. Right: Learning autonomous UAV flight [Ross et al., 2013]. The UAV flew autonomously in real forest environments. In DAGGER, the learner complements initial demonstrations by querying an expert online for demonstrations specifically for states induced by the learner's policy.

### 3.9.3 Learning to Control with DAGGER

Ross et al. [2011] demonstrated how the DAGGER algorithm learns to play a video game as shown in Figure 3.18. Visual features of 2D images were used as system state, and a policy linear to the visual features was learned using DAGGER. A human expert demonstrated the correct steering for observed game images. DAGGER has also been applied to control UAVs as shown in Figure 3.18 [Ross et al., 2013]. Ross et al. [2013] trained a controller that can avoid trees in natural environments using a small set of human demonstrations and performed autonomous flights in a real forest. In both examples, a small error at an early timestep may lead the learner to an unseen state which largely deviates from expert demonstrations. Since the learner encounters various states in which the expert did not demonstrate how to act, an online learning approach such as DAGGER is essential in these applications.

# 4
# Inverse Reinforcement Learning

In inverse reinforcement learning (IRL) [Russell, 1998], also called inverse optimal control [Kalman, 1964, Moylan and Anderson, 1973, Dvijotham and Todorov, 2010, Levine and Koltun, 2012], inverse planning [Baker et al., 2009], or structural estimation of MDPs Rust [1994] the learner tries to recover a reward function from a policy (or demonstrations of a policy). Recovering the reward function can be beneficial when the reward function is the most parsimonious way to describe the desired behavior.

We begin discussion of inverse reinforcement learning (IRL) with a definition of IRL in §4.1, discuss the critical assumption of linear vs. nonlinear reward functions in §4.2, continue with model-based IRL methods in §4.4 and model-free IRL methods in §4.5, give an information theoretic interpretation of IRL methods in §4.6, show how partial observability affects IRL in §4.7, and, finally finish with applications of IRL in §4.8.

## 4.1 Problem Statement

Russell defines the problem of IRL [Russell, 1998] as follows:

> **Given** 1) measurements of an agent's behavior over time, in a variety of circumstances, 2) measurements of the sensory inputs to that agent; 3) a model of the physical environment (including the agent's body).
> **Determine** the reward function that the agent is optimizing.

A common assumption in IRL is that the demonstrator utilizes a Markov decision process (MDP) for decision making. Formally, an MDP is a tuple $(\mathcal{X}, \mathcal{U}, \mathcal{P}, \gamma, D, R)$. $\mathcal{X}$ is a finite set of states; $\mathcal{U}$ is a set of control inputs; $\mathcal{P}$ is a set of state transitions probabilities; $\gamma \in [1, 0)$ is a discount factor; $D$ is the initial-state distribution from which the initial state $\boldsymbol{x}_0$ is drawn; and $R : \mathcal{X} \mapsto \mathbb{R}$ is the reward function. In addition, many IRL methods assume that there are vectors of features $\boldsymbol{\phi} : \mathcal{X} \mapsto [0, 1]^k$. IRL methods often estimate the reward function as a function of these features $\boldsymbol{\phi}$.

The goal of IRL is to recover the unknown reward function $R(\boldsymbol{\tau})$ from the expert's trajectories. However, since a policy can be optimal for multiple reward functions, the problem of determining the reward function is "ill-posed". To obtain the unique solution in IRL, many studies have proposed additional objective functions to be optimized, such as margin between the optimal policy and others [Ng and Russell, 2000, Abbeel and Ng, 2004, Ratliff et al., 2006b,a, 2009, Silver et al., 2010] and to maximize the entropy [Ziebart et al., 2008, Ziebart, 2010, Kitani et al., 2012, Shiarlis et al., 2016].

Many IRL methods usually require an iterative learning process (although see Ratliff et al. [2006b] for a description directly in terms of a quadratic program). Algorithm 14 summarizes a class of IRL methods that proceed by alternatingly solving an RL style problem and updating a cost function estimate. In order to obtain the performance equivalent to the expert's policy, state-action visitation frequency $\boldsymbol{\mu}$ needs to be matched between demonstrated trajectories and the trajectories

---

**Algorithm 14** Abstract version of feature matching inverse reinforcement learning

---

**Input:** Expert trajectories $\mathcal{D} = \{\boldsymbol{\tau}_i\}_{i=1}^{N}$
Initialize the reward function and policy parameters $\boldsymbol{w}, \boldsymbol{\theta}$
**repeat**
  Evaluate the state-action visitation frequency $\boldsymbol{\mu}$ of the current policy $\pi_\theta$
  Evaluate the objective function $\mathcal{L}$ and its derivative $\nabla_{\boldsymbol{w}}\mathcal{L}$ by comparing $\boldsymbol{\mu}$ and the state-action distribution implied by $\mathcal{D}$
  Update the reward function parameter $\boldsymbol{w}$
  Update the policy parameter $\boldsymbol{\theta}$ with a reinforcement learning method
**until**
**return** optimized policy parameters $\boldsymbol{\theta}$ and reward function parameter $\boldsymbol{w}$

---

induced by the learner's policy as indicated by Abbeel and Ng [2004], Ho and Ermon [2016]. The reward function parameter $\boldsymbol{w}$ is updated through optimizing the objective function under the expected feature matching constraint. This objective function is designed to estimate the reward function which makes the demonstrations appear more optimal than the current policy. The policy parameters $\boldsymbol{\theta}$ are then updated using an optimal control solution (i.e. reinforcement learning method) based on the current estimate of the reward function. For this purpose, inverse reinforcement learning methods often have a RL style procedure in an inner loop. By repeating this process, the policy and reward function parameters can be obtained.

Each IRL method has a different way of performing these steps. Model-based methods require the knowledge of system dynamics in order to evaluate the state-action visitation frequency. On the contrary, model-free methods often employ sampling-based methods for this purpose. In order to obtain an optimal policy based on the recovered reward function, various reinforcement learning methods can be used. Although MDP solvers can be used for the policy optimization in discrete state-action space as in [Abbeel and Ng, 2004, Ratliff et al., 2006a], recent policy search methods can be also used. For example, Finn et al. [2016b] employed guided policy search [Levine and Abbeel, 2014], and Ho and Ermon [2016] and Ho et al. [2016] employed trust region policy optimization [Schulman et al., 2015].

## 4.2 Model-Based and Model-Free Inverse Reinforcement Learning Methods

As with behavioral cloning methods, IRL methods can be categorized into two categories: model-based and model-free methods. Model-based IRL methods assume that the dynamics of the system, e.g. state transition probabilities, are known. The prior knowledge of the system dynamics is often used to evaluate and update the learned reward function and policy. These model-based IRL method are relatively simple to implement when the system dynamics are known. However, it is challenging to apply model-based IRL methods to applications with nonlinear dynamics, which are hard to estimate. On the other hand, model-free IRL methods do not require prior knowledge of the system dynamics. Model-free IRL methods evaluate and update the learned reward function and policy using sampling-based methods, which can be applied to systems with nonlinear dynamics. However, it is necessary to sample many trajectories to estimate the trajectory distribution, which can be time-consuming and computationally expensive. Table 4.1 summarizes the advantages and disadvantages of model-free and model-based IRL methods.

## 4.3 Design Choices for Inverse Reinforcement Learning Methods

In addition to design choices we described in Chapter 2, there are IRL specific design choices:

1. **What objective should be used to obtain the unique solution in IRL?** As discussed in §4.1, IRL itself is an ill-posed problem, and it is necessary to design the objective function so as to obtain the unique solution in IRL. Table 4.2 summarizes different objectives for learning reward functions. As shown, the maximum entropy principle is a popular choice in recent studies on IRL, although the concept of maximizing the margin between the optimal policy and others was popular in the early studies on IRL. The maximum entropy principle is well-founded in information theory, and we review the related IRL methods in §4.4.3.

2. **Should the reward function be linear or nonlinear to the features?** Although many IRL methods employ a reward function linear to the features, complex tasks in robotics require a nonlinear reward function. On the other hand, IRL with the reward function nonlinear to the features is more challenging than IRL with the linear reward functions. Therefore, we need to consider the most parsimonious representation of the reward function among sufficiently expressive ones.

**Table 4.1:** Advantages and disadvantages of model-based and model-free methods in inverse reinforcement learning. Model-based IRL methods can be more data-efficient compared to model-free methods. However, it is challenging to apply model-based IRL methods to systems with nonlinear dynamics. Model-free IRL methods have been applied to systems with nonlinear dynamics.

|  | **Model-free** | **Model-based** |
|---|---|---|
| **Advantages** | Applicable to systems with nonlinear and unknown dynamics | Estimation of the trajectory distribution is data-efficient. |
| **Disadvantages** | It is necessary to sample many trajectories to estimate the trajectory distribution. | Model learning can be very difficult. It is hard to apply to underactuated systems. |

**Table 4.2:** Objectives to obtain the unique solution in inverse reinforcement learning. The concept of maximizing the margin between the optimal policy and others was popular in the early studies on IRL. The maximum entropy principle is a dominant choice for recent IRL methods.

| Objectives | Employed by |
|---|---|
| *Maximum margin* | [Ng and Russell, 2000, Abbeel and Ng, 2004, Ratliff et al., 2006b,a, 2009, Silver et al., 2010, Zucker et al., 2011] |
| *Maximum entropy* | [Ziebart et al., 2008, Ramachandran and Amir, 2007, Choi and Kim, 2011b, Ziebart, 2010, Boularias et al., 2011, Kitani et al., 2012, Shiarlis et al., 2016, Ho and Ermon, 2016, Finn et al., 2016b] |
| *Other* | [Doerr et al., 2015, Arenz et al., 2016] |

Table 4.3 shows categorization of the existing IRL methods. As one can see, many IRL methods are model-based and use the linear reward function. On the contrary, model-free methods with nonlinear reward functions have not been investigated well.

In the next section, we review model-based IRL methods, and thereafter, we review model-free IRL methods.

## 4.4 Model-Based Inverse Reinforcement Learning Methods

In this section, we review model-based IRL methods, which leverage prior knowledge about system dynamics.

**Table 4.3:** Categorization of existing inverse reinforcement learning methods. However, tasks such as manipulation in robotic applications require a nonlinear reward function.

|  | **Model-free** | **Model-based** |
|---|---|---|
| **Linear reward** | [Boularias et al., 2011, Kalakrishnan et al., 2013] | [Abbeel and Ng, 2004, Ratliff et al., 2006b, Silver et al., 2010, Ramachandran and Amir, 2007, Choi and Kim, 2011b, Ziebart et al., 2008, Ziebart, 2010, Levine and Koltun, 2012, Hadfield-Menell et al., 2016] |
| **Nonlinear reward** | [Finn et al., 2016b, Ho and Ermon, 2016] | [Ratliff et al., 2006a, 2009, Silver et al., 2010, Grubb and Bagnell, 2010, Levine et al., 2011] |

---

**Algorithm 15** IRL by expected feature matching [Abbeel and Ng, 2004]

---

**Input:** Dataset of the demonstrations $\mathcal{D} = \{(\boldsymbol{x}_i, \boldsymbol{u}_i)\}_{i=1}^{N}$, termination threshold $\epsilon$
Randomly pick some policy $\pi_0^{\mathrm{L}}$
Compute $\boldsymbol{\mu}^{\mathrm{E}}$ using $\mathcal{D}$
Perform rollouts and $\boldsymbol{\mu}_0^{\mathrm{L}} = \boldsymbol{\mu}(\pi_0^{\mathrm{L}})$
Set $i = 1$
**repeat**
  Compute $t = \max_{\boldsymbol{w}:\|\boldsymbol{w}\|_2 \leq 1} \min_{j \in \{0,\ldots,i-1\}} \boldsymbol{w}^\top(\boldsymbol{\mu}^{\mathrm{E}} - \boldsymbol{\mu}_j^{\mathrm{L}})$
  Compute the optimal policy $\pi_i^{\mathrm{L}}$ based on $r(\boldsymbol{x}) = \boldsymbol{w}^\top \boldsymbol{\phi}(\boldsymbol{x})$
  Compute $\boldsymbol{\mu}_i^{\mathrm{L}} = \boldsymbol{\mu}(\pi_i^{\mathrm{L}})$
  Set $i \leftarrow i + 1$
**until** $t^i < \epsilon$
**return** $\pi^i : i = 0, \ldots, n$

---

### 4.4.1 Feature Expectation Matching

Abbeel and Ng [2004] proposed to match the *feature expectation* in order to solve IRL problems. If we assume the reward function is linear w.r.t. the features, the reward function is given by

$$r(\boldsymbol{x}) = \boldsymbol{w}^\top \boldsymbol{\phi}(\boldsymbol{x}), \quad (4.1)$$

where $\boldsymbol{\phi}(\boldsymbol{x})$ is the feature vector of the state $\boldsymbol{x}$, and $\boldsymbol{w}$ is a weight vector. Therefore, the expected reward of a policy $\pi$ is given by

$$\mathbb{E}[R|\pi] = \mathbb{E}\left[\sum_{t=0}^{T} \gamma^t r(\boldsymbol{x}_t) \middle| \pi\right] = \mathbb{E}\left[\sum_{t=0}^{T} \gamma^t \boldsymbol{w}^\top \boldsymbol{\phi}(\boldsymbol{x}_t) \middle| \pi\right] = \boldsymbol{w}^\top \mathbb{E}\left[\sum_{t=0}^{T} \gamma^t \boldsymbol{\phi}(\boldsymbol{x}_t) \middle| \pi\right]. \quad (4.2)$$

Abbeel and Ng [2004] defined the *feature expectation* of a policy $\pi$ as

$$\boldsymbol{\mu}(\pi) = \mathbb{E}\left[\sum_{t=0}^{T} \gamma^t \boldsymbol{\phi}(\boldsymbol{x}_t) \middle| \pi\right] \in \mathbb{R}^k. \quad (4.3)$$

Using this notation the value of a policy can be rewritten as

$$\mathbb{E}[R|\pi] = \boldsymbol{w}^\top \boldsymbol{\mu}(\pi), \quad (4.4)$$

where $R = \sum_{t=0}^{T} \gamma^t r(\boldsymbol{x}_t)$. Based on this matching of the feature expectation, Abbeel and Ng [2004] proposed to learn the policy from demonstrations so as to maximize the difference between the optimal policy and others. Maximization of the difference between the optimal policy and others was formulated as a quadratic program. By iteratively updating the learned policy, the algorithm finds the optimal policy close to the demonstrated policy. Algorithm 15 summarizes the method in [Abbeel and Ng, 2004].

The matching feature expectation appears also in other IRL methods, such as Ziebart et al. [2008]. However, matching the expected feature count is ambiguous since multiple policies can achieve the same expected feature counts. Therefore, it is necessary to use additional conditions that should be satisfied by the optimal policy.

### 4.4.2   Maximum Margin Planning

To obtain the unique solution in IRL, Ratliff et al. [2006b] proposed maximum margin Planning (MMP). The idea of MMP is to find the cost function that maximizes the difference between the optimal policy and others. MMP finds the cost function in which the cost of the demonstrated trajectory $\mathcal{C}(\boldsymbol{\tau}_{\text{demo}})$ is lower than the cost of other alternative trajectories $\mathcal{C}(\boldsymbol{\tau})$ by a certain margin. This constraint can be expressed as

$$\mathcal{C}(\boldsymbol{\tau}^{\text{demo}}) \leq \mathcal{C}(\boldsymbol{\tau}) - L(\boldsymbol{\tau}), \tag{4.5}$$

where $L(\boldsymbol{\tau})$ is the loss function. If the loss function $L(\boldsymbol{\tau})$ is large, the cost difference between the demonstrated trajectory and other trajectories is large. Since we need to consider only the minimizer of the right-hand side of (4.5), (4.5) can be rewritten as

$$\mathcal{C}(\boldsymbol{\tau}^{\text{demo}}) \leq \min\{\mathcal{C}(\boldsymbol{\tau}) - L(\boldsymbol{\tau})\}. \tag{4.6}$$

In MMP in [Ratliff et al., 2006b], it is assumed that the cost function is linear to the features of the trajectory as $\mathcal{C}(\boldsymbol{\tau}) = \boldsymbol{w}^\top \boldsymbol{\phi}(\boldsymbol{\tau})$ where $\boldsymbol{w}$ is the weight and $\boldsymbol{\phi}(\tau)$ are the trajectory features. If the trajectory features $\boldsymbol{\phi}(\tau)$ are linear to the state-action frequency counts $\boldsymbol{\mu} \in \mathbb{R}^{|\mathcal{X}||\mathcal{U}|}$, $\boldsymbol{\phi}(\tau)$ is given by $\boldsymbol{\phi}(\boldsymbol{\tau}) = F\boldsymbol{\mu}$ where $F \in \mathbb{R}^{d \times |\mathcal{X}||\mathcal{U}|}$ is the feature matrix.

Likewise, if the loss function $L(\boldsymbol{\tau})$ is linear to $\boldsymbol{\mu}$, the loss function of the trajectory is given by $L(\boldsymbol{\tau}) = \boldsymbol{l}^\top \boldsymbol{\mu}$ where $\boldsymbol{l} \in \mathbb{R}^{|\mathcal{X}||\mathcal{U}|}$ is the loss vector. Given a training set $\mathcal{D} = \{F_i, \boldsymbol{\tau}_i, \boldsymbol{l}_i\}_{i=1}^N$, the problem of finding $\boldsymbol{w}$ can be formalized as a quadratic program:

$$\min_{\boldsymbol{w}, \zeta_i} \frac{1}{2} \|\boldsymbol{w}\|^2 + \frac{1}{N} \sum_{i=1}^N \zeta_i \qquad (4.7)$$

$$\text{s.t.} \forall i, \boldsymbol{w}^\top \boldsymbol{\phi}_i(\boldsymbol{\tau}_i) \leq \min\left\{\boldsymbol{w}^\top \boldsymbol{\phi}_i(\boldsymbol{\tau}) - \boldsymbol{l}_i^\top \boldsymbol{\mu}\right\} + \zeta_i \qquad (4.8)$$

The slack variable $\{\zeta\}_{i=1}^N$ allows the violation of the constraints in a similar manner as in support vector machines [Vapnik, 1998]. If we use a slack variable $\zeta_i = \boldsymbol{w}^\top F_i \boldsymbol{\mu}_i - \min_{\boldsymbol{\mu}} \left\{\boldsymbol{w}^\top F_i \boldsymbol{\mu} - \boldsymbol{l}_i^\top \boldsymbol{\mu}\right\}$, the objective function can be obtained as

$$\mathcal{L}_{\text{MMP}}(\boldsymbol{w}) = \frac{1}{N} \sum_{i=1}^N \left(\boldsymbol{w}^\top F_i \boldsymbol{\mu}_i - \min_{\boldsymbol{\mu}} \left\{\boldsymbol{w}^\top F_i \boldsymbol{\mu} - \boldsymbol{l}_i^\top \boldsymbol{\mu}\right\}\right) + \frac{\lambda}{2} \|\boldsymbol{w}\|, \qquad (4.9)$$

which Ratliff et al. [2009] call the maximum margin objective where $\lambda > 0$ is the regularization parameter.

For solving this problem, a method based on subgradients is used in Ratliff et al. [2006b]. MMP assumes access to a MDP solver that returns the optimal trajectory by solving the problem

$$\boldsymbol{\tau}^* = \arg\min \mathcal{C}(\boldsymbol{\tau}), \qquad (4.10)$$

where $\mathcal{C}(\boldsymbol{\tau})$ is the cumulative cost of the trajectory $\boldsymbol{\tau}$. MMP uses the loss-augmented cost map $\tilde{\mathcal{C}}(\boldsymbol{\tau}) = \mathcal{C}(\boldsymbol{\tau}) - L(\boldsymbol{\tau})$ to plan the trajectory. Algorithm 16 summarizes the procedure of MMP.

The MMP framework was extended to LEARCH (LEArning to seaRCH), which is a framework for learning nonlinear cost functions efficiently [Ratliff et al., 2009, Silver et al., 2010, Zucker et al., 2011]. In LEARCH, exponential functional gradient descent was used for optimizing the maximum margin planning objective.

The policy obtained in MMP is based on efficient MDP solvers, which generate deterministic optimal policies. However, robotic systems with large configuration space dimensionality often require a

---

**Algorithm 16** Maximum margin planning Ratliff et al. [2006b]

---

**input:** Training set $\mathcal{D} = \{F_i, \boldsymbol{\tau}_i, \boldsymbol{l}_i\}_{i=1}^{N}$, regularization parameter $\lambda > 0$, stepsize sequence $\{\alpha_t\}$, iteration $T$
**while** $t < T$ **do**
  **for** $i = 1, ..., N$ **do**
    Compute the loss-augmented cost map $\tilde{c}_i = \boldsymbol{w}^\top F_i - \boldsymbol{l}_i^\top$
    Compute the optimal trajectory $\boldsymbol{\tau}_i^* = \arg\min \tilde{c}_i \boldsymbol{\mu}$
    Compute the state-action frequency couts $\boldsymbol{\mu}_i^*$
  **end for**
  Compute the subgradient $\boldsymbol{g} \in \partial \mathcal{L}_{\mathrm{MMP}}(\boldsymbol{w})$
  $\boldsymbol{w} \leftarrow \boldsymbol{w} - \alpha_t \boldsymbol{g}$
  (Optional) Project $\boldsymbol{w}$ on to any additional constraint
  $t \leftarrow t + 1$
**end while**
**return** $\boldsymbol{w}$

---

stochastic policy and approximations in planning [Ratliff et al., 2009]. In the next section, we review the maximum entropy IRL by Ziebart et al. [2008] that considers the distribution of the resulting trajectories.

### 4.4.3 Inverse Reinforcement Learning Based on the Maximum Entropy Principle

In recent studies on IRL, the maximum entropy principle [Jaynes, 1957] is often used to obtain the unique reward function. In the following section, we review IRL methods based on the maximum entropy principle.

#### 4.4.3.1 Maximum Entropy Inverse Reinforcement Learning

As described in §4.1, the IRL problem is ill-posed because a policy can be optimal for multiple reward functions. The max-margin approach described in the previous section works well when there is a single reward function that is clearly better than alternatives. However, in other cases optimizing a distribution over behaviors may be preferable.

The maximum entropy principle [Jaynes, 1957] suggests to choose a distribution that maximizes the entropy among the distributions that matches the feature expectations of the demonstrator [Dudík and Schapire, 2006, Ziebart et al., 2008]. Following this principle, Ziebart et al. [2008] proposed to learn a policy that maximizes the entropy

$$H(p(\boldsymbol{\tau})) = \sum p(\boldsymbol{\tau}) \ln \frac{1}{p(\boldsymbol{\tau})} \qquad (4.11)$$

subject to the constraints

$$\mathbb{E}_{\pi^{\mathrm{L}}}[\boldsymbol{\phi}(\boldsymbol{\tau})] = \mathbb{E}_{\pi^{\mathrm{E}}}[\boldsymbol{\phi}(\boldsymbol{\tau})], \qquad (4.12)$$

$$\sum_{\boldsymbol{\tau}} p(\boldsymbol{\tau}) = 1, \quad \forall \boldsymbol{\tau},\ p(\boldsymbol{\tau}) > 0, \qquad (4.13)$$

where $\mathbb{E}_{\pi^{\mathrm{L}}}[\boldsymbol{\phi}(\boldsymbol{\tau})]$ is the expected feature count with respect to the learner's policy and $\mathbb{E}_{\pi^{\mathrm{E}}}[\boldsymbol{\phi}(\boldsymbol{\tau})]$ is the expected feature count with respect to the expert's policy.

Among the distributions that satisfy $\mathbb{E}_{\pi^{\mathrm{L}}}[\boldsymbol{\phi}(\boldsymbol{\tau})] = \mathbb{E}_{\pi^{\mathrm{E}}}[\boldsymbol{\phi}(\boldsymbol{\tau})]$, the maximum entropy distribution follows

$$p(\boldsymbol{\tau}) \propto \exp\left(R(\boldsymbol{\tau})\right), \qquad (4.14)$$

where $p(\boldsymbol{\tau})$ is the probability of the trajectory $\boldsymbol{\tau}$, and $R(\boldsymbol{\tau}) = \boldsymbol{w}^\top \boldsymbol{\phi}(\boldsymbol{\tau})$ is the reward of $\boldsymbol{\tau}$. The parameter vector $\boldsymbol{w}$ is the Lagrangian multiplier for the feature matching constraint. Hence, we can see that, due to the feature matching constraint, the reward function is linear in the trajectory features. The probability of the trajectory can hence be expressed as

$$p(\boldsymbol{\tau}|\boldsymbol{w}) = \frac{1}{Z(\boldsymbol{w})} \exp\left(\boldsymbol{w}^\top \boldsymbol{\phi}(\boldsymbol{\tau})\right), \qquad (4.15)$$

where $Z(\boldsymbol{w})$ is the partition function given by $Z(\boldsymbol{w}) = \sum_{\boldsymbol{\tau}} \exp\left(\boldsymbol{w}^\top \boldsymbol{\phi}(\boldsymbol{\tau})\right)$.

However, Equation 4.15 only holds for deterministic environments. For stochastic environments, the trajectory distribution is also affected by the transition probabilities, i.e.,

$$p(\boldsymbol{\tau}|\boldsymbol{w}) = \frac{1}{Z(\boldsymbol{w})} \exp\left(\boldsymbol{w}^\top \boldsymbol{\phi}(\boldsymbol{\tau})\right) \prod_{\boldsymbol{x}_{t+1}, \boldsymbol{u}_t, \boldsymbol{x}_t \in \boldsymbol{\tau}} p(\boldsymbol{x}_{t+1}|\boldsymbol{u}_t, \boldsymbol{x}_t). \qquad (4.16)$$

The implication of this observation is that the agent is now trying to optimize

$$\tilde{R}(\boldsymbol{\tau}) = \boldsymbol{w}^\top \boldsymbol{\phi}(\boldsymbol{\tau}) + \sum_t \log p(\boldsymbol{x}_{t+1}|\boldsymbol{u}_t, \boldsymbol{x}_t),$$

where we have a bias term due to the stochasticity of the environment. This is one of the main theoretical drawbacks of maximum entropy IRL, which is addressed by follow-up work such as the maximum causal entropy IRL [Ziebart, 2010].

The parameter $\boldsymbol{w}$ of the reward can be obtained by maximizing the likelihood of the observed data under the maximum entropy distribution as

$$\boldsymbol{w}^* = \arg\max_{\boldsymbol{w}} \mathcal{L}_{\text{ME}}(\boldsymbol{w}) = \arg\max_{\boldsymbol{w}} \sum_{\boldsymbol{\tau}^{\text{demo}}} \ln p(\boldsymbol{\tau}^{\text{demo}}|\boldsymbol{w}). \quad (4.17)$$

Since maximizing the likelihood is equivalent to the M-projection, this problem formulation can be interpreted as M-projection onto the manifold of the maximum entropy distribution, which we discussed in §2.7.1. Since the objective function $\mathcal{L}_{\text{ME}}(\boldsymbol{w})$ is convex, this optimization can be solved using gradient-based methods. The gradient is given by the difference between the empirical feature counts from demonstrations and the expected feature counts from the learner's policy as

$$\nabla \mathcal{L}_{\text{ME}}(\boldsymbol{w}) = \mathbb{E}_{\pi^{\text{E}}}[\boldsymbol{\phi}(\boldsymbol{\tau})] - \sum_{\boldsymbol{\tau}} p(\boldsymbol{\tau}|\boldsymbol{w})\boldsymbol{\phi}(\boldsymbol{\tau}) = \mathbb{E}_{\pi^{\text{E}}}[\boldsymbol{\phi}(\boldsymbol{\tau})] - \sum_{\boldsymbol{x}_i} D_{\boldsymbol{x}_i} \boldsymbol{\phi}(\boldsymbol{x}_i).$$
$$(4.18)$$

If $\boldsymbol{\phi}(\boldsymbol{\tau}) = \sum_{t=0}^{T} \boldsymbol{\phi}(\boldsymbol{x}_t)$, then the expectation over the state-features $\boldsymbol{\phi}(\boldsymbol{x})$ can be computed by estimating the expected state visitation frequencies $D_{\boldsymbol{x}_i}$ of the current reward model, at least in discrete domains. For computing these frequencies, a backward-forward message passing algorithm can be used. Algorithm 17 summarizes the procedure for computing the state visitation frequencies.

Although the maximum entropy IRL proposed by Ziebart et al. [2008], Rust [1994] works well in MDP problems, it assumes that the state transition distribution is known, which is not the case in many robotic applications. Sampling-based or model learning extensions must be applied for problems where the model is not specified.

---

**Algorithm 17** Expected edge frequency calculation [Ziebart et al., 2008]

---

**Backward pass**
Set $Z_{x_\text{terminal}} = 1$
Recursively compute for $N$ iterations
$Z_{\boldsymbol{u}_{i,j}} = \sum_k p(\boldsymbol{x}_k|\boldsymbol{x}_i, \boldsymbol{u}_{i,j}) \exp(R(\boldsymbol{x}_i|\boldsymbol{w})) Z_{\boldsymbol{x}_k}$
$Z_{\boldsymbol{x}_i} = \sum_j Z_{\boldsymbol{u}_{i,j}}$
**Local action probability computation**
$p(\boldsymbol{u}_{i,j}|\boldsymbol{x}_i) = \frac{Z_{\boldsymbol{u}_{i,j}}}{Z_{\boldsymbol{x}_i}}$
**Forward pass**
Set $D_{s_i,t} = p(\boldsymbol{x}_i = \boldsymbol{x}_\text{initial})$
Recursively compute for $t = 1$ to $N$
$D_{\boldsymbol{x}_k,t+1} = \sum_{\boldsymbol{x}_i} \sum_{\boldsymbol{u}_{i,j}} D_{\boldsymbol{x}_k,t} p(\boldsymbol{u}_{i,j}|\boldsymbol{x}_i) p(\boldsymbol{x}_k|\boldsymbol{x}_i, \boldsymbol{u}_{i,j})$
**Summing frequencies**
$D_{\boldsymbol{x}_i} = \sum_t D_{\boldsymbol{x}_i,t}$

---

#### 4.4.3.2 Maximum Causal Entropy Inverse Reinforcement Learning

In order to fix the theoretical drawbacks of max-ent IRL in case of stochastic dynamics, Ziebart [2010] proposed to use the maximum causal entropy for IRL. The key idea of the causal entropy is that action choices need to be causal, i.e., the action selection at time step $t$ needs to be independent from future states in the trajectory. Using these insights, a new algorithm can be developed that also incorporates the stochasticity of the dynamics in the reward estimation. Contrary to maximum entropy IRL, maximum causal entropy IRL removes the "bonus entropy" that is due to the stochastic dynamics of an environment itself. This prevents learning policies that simply attempt to target areas in state-space of high stochasticity.

Maximum causal entropy IRL Ziebart [2010], tries to find the policy $\pi^*(\boldsymbol{u}|\boldsymbol{x})$, which maximizes the causal entropy $H(\boldsymbol{u}_{1:T}||\boldsymbol{x}_{1:T})$ of the actions given the states, i.e,

$$\pi^*(\boldsymbol{u}|\boldsymbol{x}) = \operatorname*{argmax}_{\pi^\text{L}(\boldsymbol{u}|\boldsymbol{x})} H(\boldsymbol{u}_{1:T}||\boldsymbol{x}_{1:T}) \tag{4.19}$$

subject to the constraint of feature expectation matching

$$\mathbb{E}_{\pi^{\mathrm{L}}}[\boldsymbol{\phi}(\boldsymbol{\tau})] = \mathbb{E}_{\pi^{\mathrm{E}}}[\boldsymbol{\phi}(\boldsymbol{\tau}^{\mathrm{demo}})],$$
$$\sum_{\boldsymbol{u}} \pi^{\mathrm{L}}(\boldsymbol{u}|\boldsymbol{x}) = 1, \quad \pi^{\mathrm{L}}(\boldsymbol{u}|\boldsymbol{x}) \geq 0 \;, \qquad (4.20)$$

where the feature function $\boldsymbol{\phi}(\boldsymbol{\tau}) = \sum_t \boldsymbol{\phi}_t(\boldsymbol{x}_t, \boldsymbol{u}_t)$ is given by the sum over state-action features. The causal entropy is defined as

$$\begin{aligned}
H(\boldsymbol{u}_{1:T}||\boldsymbol{x}_{1:T}) &= \sum_{t=1}^{T} H(\boldsymbol{u}_t|\boldsymbol{u}_{1:t-1}, \boldsymbol{x}_{1:t}) \qquad (4.21) \\
&= -\sum_{t=1}^{T} \sum_{\boldsymbol{u}_{1:t}, \boldsymbol{x}_{1:t}} p(\boldsymbol{u}_{1:t}, \boldsymbol{x}_{1:t}) \ln\left(\pi(\boldsymbol{u}_t|\boldsymbol{x}_{1:t}, \boldsymbol{u}_{1:t-1})\right) \;,
\end{aligned}$$

where $H(\boldsymbol{u}_t|\boldsymbol{u}_{1:t-1}, \boldsymbol{x}_{1:t})$ is the conditional entropy and $p(\boldsymbol{u}_{1:t}, \boldsymbol{x}_{1:t})$ is the joint distribution over all states and actions until time step $t$. Contrary to the conditional entropy $H(\boldsymbol{u}_{1:T}|\boldsymbol{x}_{1:T})$, that is implicitly used in standard max-ent IRL, the causal entropy $H(\boldsymbol{u}_{1:T}||\boldsymbol{x}_{1:T})$ conditions action choices at time step $t$ only on states until time step $t$, while the conditional entropy would make the action choice also dependent on future states (i.e., it ignores the causality).

Under the assumption that the system is Markovian, $p(\boldsymbol{x}_t|\boldsymbol{x}_{1:t-1}, \boldsymbol{u}_{1:t-1})$ reduces to $p(\boldsymbol{x}_t|\boldsymbol{x}_{t-1}, \boldsymbol{u}_{t-1})$, and $\pi(\boldsymbol{u}_t|\boldsymbol{x}_{1:t}, \boldsymbol{u}_{1:t-1})$ reduces to $\pi(\boldsymbol{u}_t|\boldsymbol{x}_t)$. Causal entropy can be maximized using dynamic programming [Ziebart, 2010] resulting in equations similar to those found in soft value-iteration methods.

### 4.4.3.3 IRL from Failed Demonstrations

Although the usual aim of inverse reinforcement learning is to learn an optimal policy from demonstrated successful trajectories, failed demonstrations also contain information that can be used for learning. Shiarlis et al. [2016] extends the maximum causal entropy IRL [Ziebart, 2010] method to learning from failed demonstrations. When using the maximum entropy approach for learning from successful demonstrations, the learned feature expectations should be similar to the demonstrated ones. In order to take failed demonstrations into account, Shiarlis et al.

[2016] modifies the maximum causal entropy IRL [Ziebart, 2010] optimization problem so that the optimized policy favors trajectories with features which are dissimilar to the features found in failed demonstrations

$$\max_{\pi^{\mathrm{L}}(\boldsymbol{u}|\boldsymbol{x}),\boldsymbol{w},\boldsymbol{z}} H(\boldsymbol{u}_{1:T}||\boldsymbol{x}_{1:T}) + \sum_{k=1}^{K} w_k z_k - \frac{\lambda}{2}||\boldsymbol{w}||^2 \quad (4.22)$$

subject to

$$\mathbb{E}_{\pi^{\mathrm{L}}(\boldsymbol{u}|\boldsymbol{x})}[\boldsymbol{\phi}(\boldsymbol{\tau}_{\mathrm{S}})] = \mathbb{E}_{\pi^{\mathrm{E}}}[\boldsymbol{\phi}(\boldsymbol{\tau}_{\mathrm{S}}^{\mathrm{demo}})],$$
$$\mathbb{E}_{\pi^{\mathrm{L}}(\boldsymbol{u}|\boldsymbol{x})}[\boldsymbol{\phi}(\boldsymbol{\tau}_{\mathrm{F}})] - \mathbb{E}_{\pi^{\mathrm{E}}}[\boldsymbol{\phi}(\boldsymbol{\tau}_{\mathrm{F}}^{\mathrm{demo}})] = z_k,$$
$$\sum_{\boldsymbol{u}} \pi^{\mathrm{L}}(\boldsymbol{u}|\boldsymbol{x}) = 1 \ , \ \pi^{\mathrm{L}}(\boldsymbol{u}|\boldsymbol{x}) \geq 0 \ ,$$

where $\lambda$ is a constant, $K$ is the number of features, and $\boldsymbol{w}$ are feature weights to optimize. While the original maximum causal entropy approach used only features of successful demonstrations $\boldsymbol{\phi}(\boldsymbol{\tau}_{\mathrm{S}}^{\mathrm{demo}})$ the approach of Shiarlis et al. [2016] uses also failed demonstration features $\boldsymbol{\phi}(\boldsymbol{\tau}_{\mathrm{F}})$. The term $\sum_{k=1}^{K} w_k z_k$ favors large distances between policy generated features and features in failed demonstrations. $\frac{\lambda}{2}||\boldsymbol{w}||^2$ is a regularization term to keep $\boldsymbol{w}$ small enough. In order to find a solution to the program in Equation 4.22, Shiarlis et al. [2016] performs gradient ascent to find the feature weights while incrementally decreasing $\lambda$ until hitting a $\lambda$ threshold. The idea in this procedure is to first emphasize finding good weights for successful demonstrations and then focus on finding weights for failed demonstrations.

### 4.4.3.4 Connection of Maximum Entropy Methods to Economics

For discrete MDPs, the Boltzmann policy form and closely-related dynamic programs have been developed in the econometrics community under the rubric of "structural estimation" from a completely different analysis. Notably, Rust [1994] derived predictive distributions of agents' actions by developing a framework for learning cost functions and predictive stochastic policies for agents acting according to a Markov Decision Process. Intriguingly, the MaxEnt policy structure and the dynamic programming algorithms derived from the maximum entropy formulation arise as well by considering an economist with only partial access to the prediction problem and including random "shocks" in a model of what would otherwise be optimal behavior. These close connections between operations research ("structural estimation"), control theory ("inverse optimal control") and machine learning ("inverse reinforcement learning") deserve much deeper investigation and better cross-fertilization between communities.

### 4.4.4 Miscellaneous Important Model-Based IRL Methods

Although the maximum entropy principle is becoming dominant in recent studies on IRL, various other model-based IRL methods have been proposed. We review some of them in the following sections.

#### 4.4.4.1 Linearly-Solvable MDPs

The linearly-solvable MDP approach of Dvijotham and Todorov [2010] differs from standard inverse reinforcement learning approaches since it estimates a value function instead of a reward or cost function. A reward function can be used to optimize a policy under different system dynamics but a value function may require system dynamics similar to those used for learning the value function.

The linearly-solvable MDP approach of Dvijotham and Todorov [2010] is designed to not require solving an MDP repeatedly. Dvijotham and Todorov [2010] assume a special kind of linearly-solvable MDP where the system dynamics are divided into passive dynamics and policy specific active dynamics. The cost function is a combination of state specific cost $c(\boldsymbol{x})$ and the cost on the difference between passive dynamics $p(\boldsymbol{x}_{t+1}|\boldsymbol{x}_t)$ and policy specific dynamics $\pi(\boldsymbol{x}_{t+1}|\boldsymbol{x}_t)$:

$$c(\boldsymbol{x}_t, \pi) = c(\boldsymbol{x}_t) + D_{\mathrm{KL}}(\pi||p) \ . \tag{4.23}$$

While the maximum entropy approach of [Ziebart et al., 2008] prefers exponentially larger rewards, Dvijotham and Todorov [2010] prefers exponentially larger value functions of the next state which is influenced by the policy $\pi$:

$$\pi(\boldsymbol{x}_{t+1}|\boldsymbol{x}_t) = \frac{p(\boldsymbol{x}_{t+1}|\boldsymbol{x}_t) z(\boldsymbol{x}_{t+1})}{Z} \ , \tag{4.24}$$

where $z(\boldsymbol{x}_{t+1}) = \exp\left(V(\boldsymbol{x}_{t+1})\right)$ is the desirability function, $Z$ is for normalization, and $V(\boldsymbol{x}_{t+1})$ is the value function. Note that the policy $\pi(\boldsymbol{x}_{t+1}|\boldsymbol{x}_t)$ is a scaled version of the passive transition probabilities $p(\boldsymbol{x}_{t+1}|\boldsymbol{x}_t)$. The IRL problem is then to estimate the value function from state transition samples. Dvijotham and Todorov [2010] finds the maximum likelihood value function from an unconstrained convex optimization problem. The advantage of the approach is that it does not require solving the MDP repeatedly. Disadvantages are that in continuous states spaces Dvijotham and Todorov [2010] needs to approximate the value function which may be more challenging then approximating reward functions which is the common approach in IRL. Moreover, a learned reward function can be used under different dynamics while this can be challenging for a value function which has been optimized for specific application dynamics.

#### 4.4.4.2 IRL Methods Based on a Bayesian Framework

The Bayesian framework is a powerful tool in machine learning which allows updating the current hypothesis based on new evidence. Ramachandran and Amir [2007] proposed an IRL method based on the Bayesian framework. In this framework, the action of the expert is considered as *evidence* that can be used to update a prior on reward functions. As in [Ziebart et al., 2008], a (different) log-linear distribution is assumed, and the posterior probability of the reward function can be computed using Bayes theorem as

$$p(R|\boldsymbol{\tau}) = \frac{p(\boldsymbol{\tau}|R)p(R)}{p(\boldsymbol{\tau})} = \frac{1}{Z}\exp(\alpha E(\boldsymbol{\tau}, R))p(R), \qquad (4.25)$$

which can be considered as a Boltzmann-type distribution with energy $E(\boldsymbol{\tau}, R)$. Computing the mean of this posterior distribution requires to recover the reward function and to learn the optimal policy from demonstrations. In the study by Ramachandran and Amir [2007], an MCMC algorithm was used to generate samples from distributions and the sample mean was used as an estimate of the mean of the true distribution.

Instead of computing the posterior mean, Choi and Kim [2011b] proposed to use maximum-a-posterior(MAP) inference. The IRL prob-lem with MAP inference can be formulated as finding the reward function $R_{\mathrm{MAP}}$ that maximizes the posterior

$$R_{\mathrm{MAP}} = \arg\max_{R} p(R|\mathcal{D}) = \arg\max_{R} \left[\ln p(\mathcal{D}|R) + \ln p(R)\right], \quad (4.26)$$

where $\mathcal{D} = \{(\boldsymbol{x}_t, \boldsymbol{u}_t)\}$ is a set of state-action pairs demonstrated by the expert. The likelihood $p(R|\mathcal{D})$ can be interpreted as a measure of the compatibility of the reward function $R$ with the demonstrated behavior data $\mathcal{D}$. For solving this problem, the method in Choi and Kim [2011b] used gradient-based optimization. Choi and Kim [2011b] suggested that MMP, Maximum entropy IRL, and other IRL methods can be interpreted in a Bayesian framework.

### 4.4.5 Learning Nonlinear Reward Functions

While research on inverse reinforcement learning originally focused mostly on learning reward functions linear with respect to feature vectors [Abbeel and Ng, 2004, Ziebart et al., 2008, Ratliff et al., 2006a, Boularias et al., 2011], many tasks, for example in robotics, require nonlinear reward functions [Silver et al., 2010, Ratliff et al., 2006b, Grubb and Bagnell, 2010, Levine et al., 2011, Finn et al., 2016b]. We discuss below such model-based approaches for modeling nonlinear rewards.

#### 4.4.5.1 Boosting Methods

The earliest approaches to rich reward function learning from model classes with high representational power was the use of *gradient-boosting*. These methods, typified by Ratliff et al. [2006b], Silver et al. [2010], Ratliff et al. [2009] can use arbitrary supervised learning algorithms in an ensemble to create highly non-linear cost functions. This approach has been used to learn locomotion strategies by demonstration Zucker et al. [2011] as well as to learn to match the real-world, rough, terrain driving strategies Silver et al. [2010, 2016, 2013]. These are among the easiest and most general approaches to implement, and an example of their use is discussed in 4.8.2.

#### 4.4.5.2 Deep Network Methods

Deep neural approaches to complex IRL cost functions were first demonstrated in Grubb and Bagnell [2010], Bradley [2010]. These approaches both build on the maximum margin formalism (although apply equally to related ones like Maximum Entropy), and use variants of backpropagation to learn sophisticated cost functions from demonstrations for interpreting sensor data.

#### 4.4.5.3 Gaussian Process IRL

To learn a nonlinear reward function, Levine et al. [2011] use a Gaussian Process (GP) approach based on the maximum entropy principle [Ziebart et al., 2008]. The original maximum entropy based approach Ziebart et al. [2008] uses linear reward features for the reward function. Levine et al. [2011] use GP inverse reinforcement learning (GPIRL) to represent a reward function which is nonlinear in the features. In general, a GP [Rasmussen and Williams, 2006] defines a probability distribution over possible outputs given some input coordinates, and, kernel hyperparameters define the actual shape of the GP. In GPIRL, the kernel hyperparameters $\boldsymbol{\theta}$ define the shape of the reward function, manually chosen feature coordinates $\boldsymbol{\phi_u}$ correspond to input coordinates, outputs correspond to demonstrated actions, and a GP models the probability distribution over true actions $\boldsymbol{u}$. The probability distribution over $\boldsymbol{u}$ and $\boldsymbol{\theta}$ is

$$p(\boldsymbol{u},\boldsymbol{\theta}|\mathcal{D},\boldsymbol{\phi_u}) \propto \left[\int_{\boldsymbol{r}} p(\mathcal{D}|\boldsymbol{r})p(\boldsymbol{r}|\boldsymbol{u},\boldsymbol{\theta},\boldsymbol{\phi_u})d\boldsymbol{r}\right]p(\boldsymbol{u},\boldsymbol{\theta}|\boldsymbol{\phi_u}) \,, \qquad (4.27)$$

where $p(\mathcal{D}|\boldsymbol{r})$ is the distribution over demonstrated trajectories and is given by the maximum entropy principle yielding trajectories exponentially more likely closer to larger rewards. $p(\boldsymbol{r}|\boldsymbol{u},\boldsymbol{\theta},\boldsymbol{\phi_u})$ is the conditional GP posterior reward probability, and $p(\boldsymbol{u},\boldsymbol{\theta}|\boldsymbol{\phi_u})$ is the prior GP probability for $\boldsymbol{u}$ and $\boldsymbol{\theta}$. In order to compute 4.27, Levine et al. [2011] use several approximations. The choice of $\boldsymbol{\phi_u}$ is particularly important since it has a large impact on both whether the solution covers the true reward function and on the computational requirements: GPs are computationally intensive because of the required covariance matrix inversion where the size of the matrix depends on input space size.

### 4.4.6 Guided Cost Learning

Recently, Finn et al. [2016b] extended the use of non-linear neural cost function approach described above Grubb and Bagnell [2010] using an adaptive sampling scheme rather then an analytic approximation as the policy optimization step in an unknown Markov Decision Process. In order to solve the cost function non-uniqueness problem as well as imperfect demonstration, Finn et al. [2016b] use the popular maximum entropy principle Ziebart et al. [2008]. For optimizing a policy and learning the cost function, the approach of Finn et al. [2016b] repeats two steps: 1) updates the cost function based on samples from both the policy and demonstrations, 2) updates the policy based on the new cost function.

Guided cost learning finds the maximum likelihood solution under the maximum entropy principle as in [Ziebart et al., 2008]. Under the maximum entropy assumption, the probability distribution of the trajectory $\boldsymbol{\tau}$ is given by $p(\boldsymbol{\tau}) = \frac{1}{Z}\exp(-c_{\boldsymbol{w}}(\boldsymbol{\tau}))$, where $c_{\boldsymbol{w}}$ is the cost function parameterized with a vector $\boldsymbol{w}$. The objective function $\mathcal{L}_{\text{GCL}}$ of the guided cost learning is given by the negative log-likelihood of the maximum entropy distribution

$$\mathcal{L}_{\text{GCL}} = \frac{1}{N} \sum_{\boldsymbol{\tau}_j \in \mathcal{D}_{\text{demo}}} c_{\boldsymbol{w}}(\boldsymbol{\tau}_i) + \ln Z \tag{4.28}$$

$$\approx \frac{1}{N} \sum_{\boldsymbol{\tau}_i \in \mathcal{D}_{\text{demo}}} c_{\boldsymbol{w}}(\boldsymbol{\tau}_i) + \ln \frac{1}{M} \sum_{\boldsymbol{\tau}_j \in \mathcal{D}_{\text{samp}}} \frac{\exp(-c_{\boldsymbol{w}}(\boldsymbol{\tau}_j))}{q(\boldsymbol{\tau}_j)}, \tag{4.29}$$

where $\mathcal{D}_{\text{demo}}$ is the set of demonstrated trajectories, $\mathcal{D}_{\text{samp}}$ is the set of samples, and $q$ is the distribution from which the $\boldsymbol{\tau}_j$ is sampled. The gradient of the cost $c_{\boldsymbol{w}}$ with respect to the parameter can be efficiently computed when the cost is represented by a neural network.

Algorithm 18 summarizes the approach. In more detail, at each iteration, Finn et al. [2016b] samples additional trajectories using the current policy and a black box simulator. Next, the cost function is updated based on all sampled trajectories and the demonstrations. The parameters of the neural network, representing the cost function, are

---

**Algorithm 18** Guided cost learning Finn et al. [2016b]

---

Initialize $q_k(\boldsymbol{\tau})$ at either a random initial controller or from demonstrations

**for** iteration $i = 1$ to $I$ **do**

    Generate samples $\mathcal{D}_{\text{traj}}$ from $q_k(\boldsymbol{\tau})$

    Append samples: $\mathcal{D}_{\text{samp}} \leftarrow \mathcal{D}_{\text{samp}} \cup \mathcal{D}_{\text{traj}}$

    Use $\mathcal{D}_{\text{samp}}$ to update the cost $c_{\boldsymbol{w}}$ using Algorithm 19

    Update $q_k(\boldsymbol{\tau})$ using $\mathcal{D}_{\text{traj}}$ and the method from Levine and Abbeel [2014] to obtain $q_{k+1}(\boldsymbol{\tau})$

**end for**

**return**  optimized cost parameters $\boldsymbol{w}$ and trajectory distribution $q(\boldsymbol{\tau})$

---

updated based on the gradient computed using the exponential cost typical for maximum entropy based approaches. For updating the policy based on the new cost function and samples, Finn et al. [2016b] uses a constrained version of linear quadratic regular (LQR) based trajectory optimization together with linearizing dynamics of local approximate Gaussian distributions estimated from the samples [Levine and Abbeel, 2014].

The approach of Finn et al. [2016b] has several interesting properties. Firstly, the policy optimization part of the approach is designed for smooth continuous trajectories found e.g. in robotics. Secondly, the approach requires a black box simulator but no explicit dynamics model.

Recently, Finn et al. [2016a], Ho and Ermon [2016] identified the close connection between Inverse Reinforcement Learning and the more recent generative adversarial networks [Goodfellow et al., 2014]. In generative adversarial networks, a generative model $G$ is trained to generate data samples so as to mimic the true data distribution, while the discriminator $D$ is trained to discriminate the data generated by $G$ and the true data. These works demonstrate that optimization/RL play the role of a *generator* while the learned cost function plays the role of a *discriminator*, albeit with the generalization of applying to any trajectory a system could take. This viewpoint sheds light on the

**Algorithm 19** Nonlinear IOC with stochastic gradients [Finn et al., 2016b]

**for** iteration $k = 1$ to $K$ **do**
    Sample demonstration batch $\hat{\mathcal{D}}_{\text{demo}} \subset \mathcal{D}_{\text{demo}}$
    Sample background batch $\hat{\mathcal{D}}_{\text{samp}} \subset \mathcal{D}_{\text{samp}}$
    Append demonstration batch to background batch
    $\hat{\mathcal{D}}_{\text{samp}} \leftarrow \hat{\mathcal{D}}_{\text{demo}} \cup \hat{\mathcal{D}}_{\text{samp}}$
    Estimate $\frac{d\mathcal{L}_{\text{GCL}}}{d\boldsymbol{w}}(\boldsymbol{w})$ using $\hat{\mathcal{D}}_{\text{demo}}$ and $\hat{\mathcal{D}}_{\text{samp}}$
    Update parameters $\boldsymbol{w}$ using gradient $\frac{d\mathcal{L}_{\text{GCL}}}{d\boldsymbol{w}}(\boldsymbol{w})$
**end for**
**return** optimized cost parameters $\boldsymbol{w}$

instabilities of GANs and the potential power of combining algorithms used in each field.

## 4.5 Model-Free Inverse Reinforcement Learning Methods

In robotics and other application fields, exact dynamics models are often difficult to come by. Model-free IRL methods side step the problem by not requiring such prior knowledge. Model-free IRL methods often employ sampling-based approaches to estimate the trajectory distribution. Although this approach requires many samples of trajectories in the learning process, it avoids the explicit learning of system dynamics.

### 4.5.1 Relative Entropy Inverse Reinforcement Learning

Although model-based IRL methods assume that the system dynamics, e.g. state transition probability, is known, model-free IRL methods do not require such prior knowledge on the system dynamics. Relative entropy IRL in [Boularias et al., 2011] is one of such model-free IRL methods. Boularias et al. [2011] proposed to minimize the relative entropy between a prior trajectory distribution $q_0(\boldsymbol{\tau})$ induced by a baseline policy and the trajectory distribution $p(\boldsymbol{\tau})$ induced by the learner's policy. For minimizing the relative entropy without prior knowledge of the system dynamics, importance-sampling is used to estimate the expected feature count in [Boularias et al., 2011]. Relative entropy IRL also assumes that the reward is given as a linear function of the feature vector as $R(\boldsymbol{\tau}) = \boldsymbol{w}^\top \boldsymbol{\phi}(\boldsymbol{\tau})$. This problem can be formulated as minimizing the relative entropy

$$\min \sum p(\boldsymbol{\tau}) \ln \frac{p(\boldsymbol{\tau})}{q_0(\boldsymbol{\tau})}, \qquad (4.30)$$

subject to the constraints

$$\forall i \in \{1, ... k\}, \ |\mathbb{E}_{\pi^\mathrm{L}}[\phi_i(\boldsymbol{\tau})] - \mathbb{E}_{\pi^\mathrm{E}}[\phi_i(\boldsymbol{\tau})]| \le \epsilon_i, \qquad (4.31)$$

$$\sum_{\boldsymbol{\tau} \in \mathcal{T}} p(\boldsymbol{\tau}) = 1, \qquad (4.32)$$

$$\forall \boldsymbol{\tau} \in \mathcal{T}, \ p(\boldsymbol{\tau}) \ge 0, \qquad (4.33)$$

where $\mathbb{E}_{\pi^\mathrm{E}}[\phi_i(\boldsymbol{\tau})]$ is the empirical expectation of the $i$th feature vector calculated from demonstrations, $\mathbb{E}_{\pi^\mathrm{L}}[\phi_i(\boldsymbol{\tau})] = \sum_{\boldsymbol{\tau}} p(\boldsymbol{\tau})\phi_i(\boldsymbol{\tau})$ is the expectation of the feature vector with respect to the learner's policy, $k$ is the number of features, $\mathcal{T}$ is a set of feasible trajectories, and the threshold $\epsilon_i$ is calculated by using Hoeffding's bound. The Lagrangian of this problem is given by

$$\mathcal{L}_\mathrm{RE}(p, \boldsymbol{w}, \eta) = \sum p(\boldsymbol{\tau}) \ln \frac{p(\boldsymbol{\tau})}{q_0(\boldsymbol{\tau})} - \boldsymbol{w}^\top \left( \sum_{\boldsymbol{\tau}} p(\boldsymbol{\tau})\boldsymbol{\phi}(\boldsymbol{\tau}) - \mathbb{E}_{\pi^\mathrm{E}}[\boldsymbol{\phi}(\boldsymbol{\tau})] \right)$$
$$- \sum_{i=1}^{k} |w_i|\epsilon_i + \eta \left( \sum_{\boldsymbol{\tau} \in \mathcal{T}} p(\boldsymbol{\tau}) - 1 \right). \qquad (4.34)$$

The dual problem is given by maximizing the dual function

$$g_\mathrm{RE}(\boldsymbol{w}) = \boldsymbol{w}^\top \mathbb{E}_{\pi^\mathrm{E}}[\boldsymbol{\phi}(\tau)] - \ln Z(\boldsymbol{w}) - \sum_{i=1}^{k} |w_i|\epsilon_i. \qquad (4.35)$$

This dual problem can be solved by using a sub-gradient-based method and importance sampling in Boularias et al. [2011]. Since the expected feature count is estimated through sampling, this method can be applied to a system with unknown dynamics.

---

**Algorithm 20** Generative adversarial imitation learning

**Input:** Expert trajectories $\mathcal{D} = \{\boldsymbol{\tau}_i\}_{i=1}^N$, initial policy and discriminator parameters $\boldsymbol{\theta}_0, \boldsymbol{w}_0$
**for** iteration $i = 1$ to $K$ **do**
  Sample Trajectories $\boldsymbol{\tau}_i \sim \pi_i^{\mathrm{L}}$
  Update the discriminator parameters from $\boldsymbol{w}_i$ to $\boldsymbol{w}_{i+1}$ with the gradient
  $\mathbb{E}_{\pi_i^{\mathrm{L}}}\left[\nabla_{\boldsymbol{w}} \ln(D_{\boldsymbol{w}}(s,a))\right] + \mathbb{E}_{\pi^{\mathrm{E}}}\left[\nabla_{\boldsymbol{w}} \ln\left(1 - D_{\boldsymbol{w}}(s,a)\right)\right]$
  Update a policy $\pi_i^{\mathrm{L}}$ using the TRPO rule with the cost function $\ln(D_{\boldsymbol{w}_{i+1}}(s,a))$, which takes a KL-constrained natural gradient step with
  $\mathbb{E}_{\pi_i^{\mathrm{L}}}\left[\nabla_{\boldsymbol{\theta}} \ln \pi^{\mathrm{L}}(\boldsymbol{u}|\boldsymbol{x}) Q(\boldsymbol{x},\boldsymbol{u}) - \lambda \nabla_{\boldsymbol{\theta}} H(\pi^{\mathrm{L}})\right],$
  where $Q(\bar{\boldsymbol{x}}, \bar{\boldsymbol{u}}) = \mathbb{E}_{\pi_i^{\mathrm{L}}}\left[\ln\left(D_{\boldsymbol{w}_{i+1}}(\boldsymbol{x},\boldsymbol{u})\right) | \boldsymbol{x}_0 = \bar{\boldsymbol{x}}, \boldsymbol{u}_0 = \bar{\boldsymbol{u}}\right]$
**end for**
**return** optimized policy parameters $\boldsymbol{\theta}$

---

**4.5.2 Generative Adversarial Imitation Learning**

Recently, Ho and Ermon [2016] proposed generative adversarial imitation learning (GAIL) by leveraging the connection noted above between GANs [Goodfellow et al., 2014] and IRL.[1] This viewpoint enables constraining the behavior of the agent to be approximately optimal according to an unknown reward function without explicitly attempting to recover that reward function.

Ho and Ermon [2016] trained a policy that reproduces the expert's behavior and a discriminator that distinguishes trajectories induced by the learner's policy from trajectories demonstrated by the expert. The state-action occupancy induced by the expert's policy in GAIL is analogous to the true data distribution in GANs. Algorithm 20 summarizes GAIL. Ho and Ermon [2016] indicated that IRL is a dual of the occupancy measure matching under the maximum entropy principle. Based

---

[1]GAIL [Ho and Ermon, 2016] cannot be fully classified as an IRL approach since GAIL does not recover the reward function. However, we introduce the study [Ho and Ermon, 2016] in the IRL section since it is relevant to the concept of IRL.

on this consideration, the objective function

$$\mathcal{L}_{\text{GA}} = \mathbb{E}_{\pi_{\boldsymbol{\theta}}^{\text{L}}}\left[\ln(D_{\boldsymbol{w}}(\boldsymbol{x}, \boldsymbol{u}))\right] - \mathbb{E}_{\pi^{\text{E}}}\left[\ln(1 - D_{\boldsymbol{w}}(\boldsymbol{x}, \boldsymbol{u}))\right] - \lambda H(\pi_{\boldsymbol{\theta}}^{\text{L}}) \quad (4.36)$$

is optimized to match the occupancy measure, where $\pi_{\boldsymbol{\theta}}^{\text{L}}$ is the learner's policy parameterized with $\boldsymbol{\theta}$, $D_{\boldsymbol{w}}$ is the discriminator network parameterized with $\boldsymbol{w}$, $H(\pi_{\boldsymbol{\theta}}^{\text{L}}) \equiv \mathbb{E}_{\pi_{\boldsymbol{\theta}}^{\text{L}}}[-\ln \pi_{\boldsymbol{\theta}}^{\text{L}}(\boldsymbol{u}|\boldsymbol{x})]$ is the $\gamma$-discounted causal entropy of the policy $\pi_{\boldsymbol{\theta}}^{\text{L}}$ in [Bloem and Bambos, 2014]. Through optimizing $\mathcal{L}_{\text{GA}}$, the discriminator network $D_{\boldsymbol{w}}$ and the policy $\pi_{\boldsymbol{\theta}}^{\text{L}}$ are trained. Here, trust region policy optimization (TRPO) proposed by Schulman et al. [2015] is used to optimize $\mathcal{L}_{\text{GA}}$ with respect to the policy parameter $\boldsymbol{\theta}$. TRPO employs the constraint between the current and updated policies in order to avoid unstable policy updates. For this purpose, the KL divergence is used as a measure of the dissimilarity of policies in TRPO.

Recent work by Baram et al. [2017] extended GAIL to the model-based approach. Baram et al. [2017] proposed to make the computation for training a stochastic policy fully differentiable by using a forward model. The empirical results show that the model-based GAIL outperforms the model-free GAIL in continuous control tasks. In addition, the work by Henderson et al. [2018] extended GAIL to the option framework for a hierarchical policy.

## 4.6 Interpretation of IRL with the Maximum Entropy Principle

As we have seen so far, many IRL methods iteratively estimate the reward function to make the demonstrations appear more optimal than other policies, then update the policy under the updated reward function, and execute the policy to get more samples which the reward function attempts to distinguish. This process is summarized in Figure 4.1. To obtain the unique solution of the "ill-posed" IRL problem, the maximum entropy principle is often used. Here, we discuss the interpretation of IRL with the maximum entropy principle.

Let us consider a prior trajectory distribution $p_0(\boldsymbol{\tau})$ and the trajectory distribution $p(\boldsymbol{\tau})$ induced by the learner's policy. Information geometry suggests to minimize the KL divergence $D_{\text{KL}}(p(\boldsymbol{\tau})||p_0(\boldsymbol{\tau}))$ from $p(\boldsymbol{\tau})$ to $p_0(\boldsymbol{\tau})$ [Amari, 2016]. The maximum entropy principle in [Jaynes, 1957] suggests to choose a distribution that maximizes the entropy among the distributions that achieve at least the same total reward. Entropy $H(p(\boldsymbol{\tau}))$ is defined as

$$H(p(\boldsymbol{\tau})) \simeq \sum p(\boldsymbol{\tau}) \ln \frac{1}{p(\boldsymbol{\tau})}, \tag{4.37}$$

whereas the KL divergence $D_{\text{KL}}(p(\boldsymbol{\tau})||p_0(\boldsymbol{\tau}))$ is defined as

$$D_{\text{KL}}(p(\boldsymbol{\tau})||p_0(\boldsymbol{\tau})) \simeq \sum p(\boldsymbol{\tau}) \ln \frac{p(\boldsymbol{\tau})}{p_0(\boldsymbol{\tau})}. \tag{4.38}$$

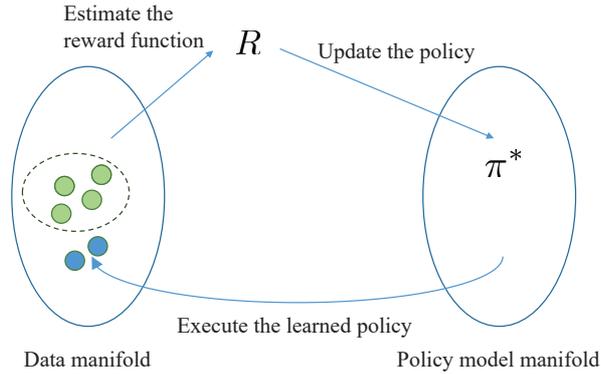

**Figure 4.1:** Illustration of many IRL approaches. Such IRL methods iteratively estimate the reward function to make the demonstrations appear more optimal than the current policy, then update the policy under the new reward function, and execute the policy virtually or physically to get more samples which the reward function attempts to distinguish.

Therefore, maximizing the entropy $H(p(\tau))$ is equivalent to minimizing the KL divergence $D_{\text{KL}}(p(\boldsymbol{\tau})||p_0(\boldsymbol{\tau}))$ under the assumption that $p_0(\boldsymbol{\tau})$ is the uniform distribution. Alternate prior distributions can be easily taken into account by simply adding a "feature" that is $\log p_0(\boldsymbol{\tau})$ either with a weight fixed to 1.0 or allowed to adapt and learn.

The maximum causal entropy distribution [Ziebart et al., 2013] can be understood to assume to remove the effects of stochastic dynamics as well. For learning tasks involving physical systems, it is often desirable to consider alternate $p_0(\boldsymbol{\tau})$, particularly by exploiting information in the system dynamics. For this reason, Dvijotham and Todorov [2010] proposed to use the trajectory distribution induced by the passive dynamics $p(\boldsymbol{x}_{t+1}|\boldsymbol{x}_t)$ of the system as the KL divergence term $p_0(\boldsymbol{\tau})$ of the cost function. Kalakrishnan et al. [2013] also approximated a trajectory distribution using trajectories sampled from the system dynamics. These methods consider the passive dynamics of the system in their problem formulation.

The relative entropy IRL approach by Boularias et al. [2011] attempts to minimize the KL divergence $D_{\text{KL}}(p(\boldsymbol{\tau})||p_0(\boldsymbol{\tau}))$, with feature matching constraints. By using importance sampling, the expected feature counts are approximated without prior knowledge of the system dynamics. Since the trajectories sampled from the actual system follow the system dynamics, we can consider that the expected feature counts approximated using importance sampling implicitly encode the system dynamics. Arenz et al. [2016] use the M-projection to obtain the data state distribution analytically, and then use the I-projection to obtain the policy given the analytic model of the data distribution. Methods that directly try to minimize the KL to the data distribution $D_{\text{KL}}(p(\boldsymbol{\tau})||q^{\text{demo}}(\boldsymbol{\tau}))$, where $q^{\text{demo}}(\boldsymbol{\tau})$ is the trajectory distribution induced by the expert policy, have not been widely researched in imitation learning to our knowledge. However, some recent research shows that any $f$-divergence can be minimized [Nowozin et al., 2016] in GANs and given the close connection to IOC methods we expect that investigations into this area may be profitable.

## 4.7 Inverse Reinforcement Learning under Partial Observability

Partial observability is common in robotics and other domains due to sensor noise and occlusions caused by objects, robots, humans, and the environment. Moreover, the whole process of IRL can be seen as a process where the agent has incomplete observations about the true reward function. Here, we discuss the cases when the expert and learner make partial observations, and, the case of formally framing IRL as the learner making partial observations about the reward function. Section 4.7.1 discusses the case when the learner partially observes the demonstrations, Section 4.7.2 then discusses the case when the expert makes partial observations when performing demonstrations, Section 4.7.3 describes how IRL can be framed as a partially observable Markov decision process, and Section 4.7.4 discusses a model for optimizing the behavior of both the expert and learner when the reward function is partially observable.

### 4.7.1 IRL from Partially Observable Demonstrations

Recently, inverse reinforcement learning with partially observable expert demonstrations has gained interest in vision research [Kitani et al., 2012] and robotics [Boularias et al., 2012, Bogert and Doshi, 2014, 2015, Bogert et al., 2016].

Noisy sensors are a common source of partial observability. To forecast human activities from noisy images, [Kitani et al., 2012] extends maximum entropy IRL [Ziebart et al., 2008] into domains where the learner only partially observes expert demonstrations. To handle partial observability, Kitani et al. [2012] proposes to use a hidden variable Markov decision process (hMDP). In hMDP, observation probabilities are part of the joint maximum entropy state-observation probability distribution

$$p(\boldsymbol{\tau}|\boldsymbol{o},\boldsymbol{\theta}) \approx \frac{\exp(\boldsymbol{w}^\top \boldsymbol{\phi}'_{\boldsymbol{\tau}})}{Z(\boldsymbol{w})} \qquad (4.39)$$

which is similar to the maximum entropy IRL trajectory probability distribution in (4.15), but, the state features $\boldsymbol{\phi}'_{\boldsymbol{\tau}}$ in (4.39) include the logarithm of the probability of the observations $\boldsymbol{o}$. For simplicity, in [Kitani et al., 2012], the observation probability is Gaussian.

Boularias et al. [2012] deal with noisy features using a graphical model based on Markov random fields (MRFs) that allows correlation between actions of similar states. Intuitively, utilizing correlations reduces noise due to the smoothing effect on observations over similar states. In many problems state similarity is easy to determine. For example in navigation, Euclidean distance can be used as a similarity measure. Boularias et al. [2012] demonstrate the approach in a simulated navigation and in a simulated grasping task. One disadvantage of the approach is that the algorithms presented in [Boularias et al., 2012] are computationally heavy.

Motivated by occlusions in robotic problems, Bogert and Doshi [2014, 2015], Bogert et al. [2016] study the problem of reward learning from partially occluded demonstrations. Moreover, the demonstrations are performed by multiple experts. Contrary to [Natarajan et al., 2010], the experts' policies are not independent from each other but take other experts into account. The methods developed in [Bogert and Doshi, 2014, 2015, Bogert et al., 2016] are based on maximum entropy IRL [Ziebart et al., 2008]. To handle partial observability, Bogert and Doshi [2014] simply do not consider occluded states and actions, but, instead, compute feature expectations only for observable states. Bogert and Doshi [2014] demonstrate the approach in multi-robot patrolling: the learner has to find out the reward functions of patrolling robots in order to plan a route around them. Bogert and Doshi [2015] consider also uncertain transition functions. Instead of discarding partially observed time steps, Bogert et al. [2016] follow a different approach by treating missing data as hidden variables and presents an expectation maximization (EM) approach for a locally optimal solution. Bogert et al. [2016] demonstrates the EM approach in a simulated reconnaissance scenario with dynamically changing occlusions and shows how a robot learns to perform a sorting task demonstrated by a human.

### 4.7.2 IRL with Incomplete Expert Observations

Usually the basic premise in IRL is that the expert observes the world state fully. However, similarly to the learner, the expert may only partially observe the world when demonstrating the task. Thus instead of an MDP model a partially observable Markov decision process (POMDP) model is needed for the expert. The formal POMDP model is identical to the MDP model except that a POMDP additionally includes observation probabilities conditioned on the next state and current action. Policy computation for POMDPs is challenging compared to MDPs. The same applies to IRL in POMDPs [Choi and Kim, 2011a]. Choi and Kim [2011a] extend classical IRL algorithms [Ng and Russell, 2000, Abbeel and Ng, 2004] to two different POMDP settings: 1) learning from a given expert's policy and 2) learning from expert trajectories. Learning from a given policy is a simpler problem than learning from trajectories. Because of the computational difficulty the demonstrations on benchmark problems are relatively simple.

### 4.7.3 Active Inverse Reinforcement Learning as a POMDP

With active inverse reinforcement learning we refer to learning the reward function when the robot is able to influence the demonstrations [Daniel et al., 2015]. An appealing way is to model the process of active inverse reinforcement learning as a partially observable Markov decision process (POMDP) where the reward function is a hidden quantity which the agent partially observes. Solving the POMDP then yields optimal actions for both gathering information about the reward function and other task specific objectives. Computational methods exist for both parametric [Dearden et al., 1999, Poupart and Vlassis, 2008] and non-parametric [Doshi-Velez et al., 2012, 2015] learning of the reward function when the IRL problem itself is modeled as a POMDP. The main disadvantage of POMDPs is the high computational complexity. The current application of POMDPs for active IRL in robotic applications is limited but an interesting avenue for future work since POMDPs offer a principled way of modeling IRL. For example, POMDPs do not suffer from the exploration-exploitation dilemma which could be a useful property in active IRL.

### 4.7.4 Cooperative Inverse Reinforcement Learning

In the vein of the approaches discussed above, Hadfield-Menell et al. [2016] frame the problem of IRL as learning a hidden reward function as a partially observable Markov decision process (POMDP). Hadfield-Menell et al. [2016] define and study the cooperative inverse reinforcement learning (CIRL) problem. A CIRL is a two player game where the human observes the reward function but the robot not. Traditional IRL [Ng and Russell, 2000] assumes that the demonstrator is acting based on an optimal policy. Hadfield-Menell et al. [2016] show that in CIRL, the human may accept sub-optimal reward if it can provide the robot with more information. CIRL defines optimal behavior for *both* the human and the robot when optimizing reward for the human. CIRL potentially leads to policies where human teaching and robot learning are jointly optimized. Hadfield-Menell et al. [2016] show that finding optimal policies for the human and robot in CIRL corresponds to solving a POMDP. A drawback of the POMDP model is that in practice exact optimal solutions for the model are hard to come by but the POMDP model can be used as a theoretical tool and a basis for practical solutions.

Hadfield-Menell et al. [2016] demonstrate the CIRL framework in simple simulated scenarios. Considering more complicated robotic experiments, the traditional way of IRL of performing close to optimal demonstrations could be easier for a human compared to teaching a robot optimally. In order to perform demonstrations which teach the robot optimally, the human has to consider how the robot optimizes learning in addition to the actual task being demonstrated.

## 4.8 Robot Applications with Inverse Reinforcement Learning Methods

Inverse reinforcement learning has been used for tasks such as parsing sentences Neu and Szepesvári [2009], car driving Abbeel and Ng [2004], path planning Ratliff et al. [2006b], Silver et al. [2010], Zucker et al. [2011], and robot motions Boularias et al. [2011], Finn et al. [2016b]. First, we review applications of model-based inverse reinforcement learning methods. Since model-based IRL methods assume that the dynamics of the system is available, they have been applied to problems where the system dynamics is completely known such as a driving simulator. Thereafter, we review applications of model-free inverse reinforcement learning methods. Since model-free IRL methods do not require prior knowledge of the system dynamics, they can be applied to robotic tasks where the dynamics of a manipulator is hard to obtain.

### 4.8.1 Learning to Drive a Car in a Simulator

Simulating car-driving is a typical application which can be modeled as an MDP problem. It is often assumed that the policy is stationary (independent of time) and that the state-action space can be approx-imated by a set of discrete states and actions. Abbeel et al. demonstrated the performance of IRL in a car-driving simulation shown in Figure 4.3 [Abbeel and Ng, 2004]. In the car simulation, five actions were available, three of which were to steer the car to one of the lanes, and two of which were to drive off the road on the left or the right side. The expert's features were computed from a single trajectory of 1200 samples. In this experiment, different driving styles were demonstrated by the expert. The results show that the method in [Abbeel and Ng, 2004] is able to imitate different driving styles.

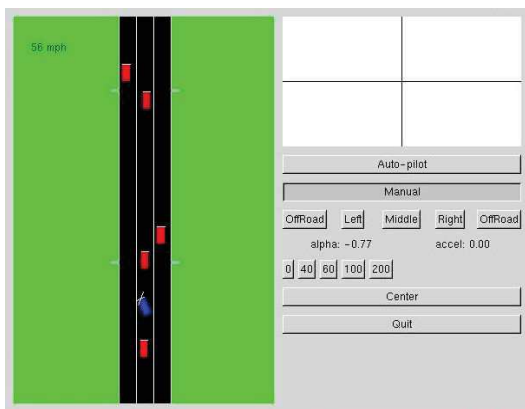

**Figure 4.2:** Screen shot of the driving simulator used in [Abbeel and Ng, 2004]. A time-invariant policy was learned using a model-based IRL method. Experimental results show that a different driving style can be learning using different demonstration data.

### 4.8.2 Learning Path Planning with MMP

Ratliff et al. [2006b], Silver et al. [2010] apply maximum margin planning (MMP) and LEARCH for finding a path with minimum accumulated cost (see Figure 4.3). Interestingly, from raw perceptual data, lattice planners can be taught human-like rough terrain driving more efficiently compared to manually programmed behavior Silver et al. [2010]. LEARCH learns the cost as a function of features and the optimal path can be found by using classic motion planning methods on the recovered cost function. The features of the MDP are based on visual (images/lidar) input as shown in Figure 4.4. The learned cost function reproduces paths incrementally more similar to the example path as the learning evolves. MMP and LEARCH have been applied to various robotic systems, including footstep planning for a quadruped robot [Zucker et al., 2011].

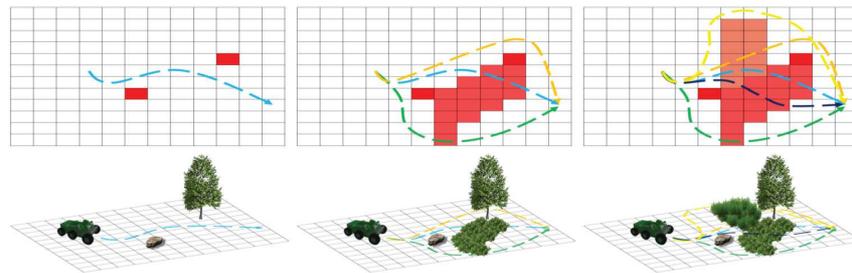

**Figure 4.3:** The learning to search (LEARCH) approach for identifying a cost function has been applied to various robotic applications including learning rough terrain navigation from sensor data. The approach iterates between building a discriminative classifier between states visited by the learner and the demonstrator, updating the cost function with the discriminative classifier, and then using classical path planning methods to identify a new proposed optimal plan.

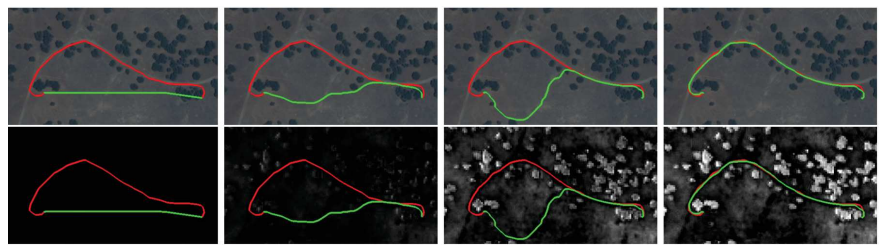

**Figure 4.4:** Examples of path planning with LEARCH [Silver et al., 2010]. Top figures show the satellite images and the bottom figures show the costs. The cost function evolves from left to right in the learning process. The red line represents the example path and the green represents the current plan. The learned cost function reproduces paths more similar to the example path as the learning evolves. The upper set of images shows the raw visual (camera) data being interpreted by the learner, the lower images show the interpretation in terms of costs (white expensive, dark low-cost).

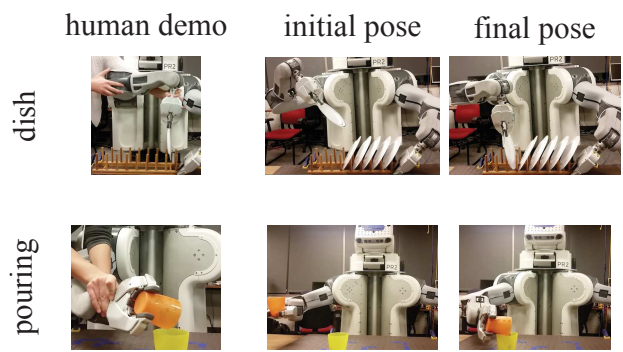

**Figure 4.5:** Learning house-keeping tasks in [Finn et al., 2016b]. Tasks that require a nonlinear reward function and a complex policy were learned using guided cost learning.

### 4.8.3 Learning Motion Planning with Deep Guided-Cost Learning

Learning manipulation tasks often requires nonlinear reward functions. Finn et al. [2016b] applied guided cost learning to house-keeping tasks such as moving dishes and pouring water shown in Figure 4.5. Demonstrations were recorded using kinesthetic teaching with a PR2 robot. As we described in §4.4.6, guided cost learning uses a neural network to represent the reward function. The state of the system was represented by vision-based features obtained by using an unsupervised learning method [Finn et al., 2016b]. The experimental results show that guided cost learning can be used to learn robotic manipulation tasks that require a nonlinear reward function under unknown dynamics.

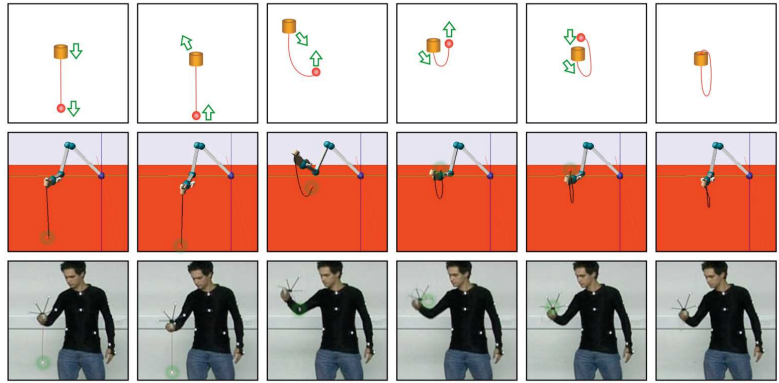

**Figure 4.6:** Learning ball-in-the-cup in [Boularias et al., 2011]. The KL divergence between the expert policy and the learner's policy was minimized using a sampling-based method.

### 4.8.4 Learning a Ball-in-a-Cup task with Relative Entropy Inverse Reinforcement Learning

Learning robotic tasks with an underactuated manipulator is nontrivial because the dynamics of the system is hard to estimate. Since model-based IRL methods require an accurate model of the system dynamics, applying model-based IRL methods to such tasks can be challenging. Boularias et al. [2011] applied the model-free Relative Entropy Inverse Reinforcement Learning (RE-IRL) approach to the Ball-in-a-cup task with an underactuated robot shown in Figure 4.6. A human demonstrated the ball-in-a-cup motion 17 times, and the motions were recorded using a 3D motion capture system. Robotic simulations showed successful learning of the demonstrated motion.

# 5

# Challenges in Imitation Learning for Robotics

We have surveyed the state of the art in imitation learning for robotics. Although imitation learning has progressed rapidly, it is clear that there are still many problems and challenges which need to be investigated. In this section, we highlight open questions and technical challenges in imitation learning.

## 5.1 Behavioral Cloning vs Inverse Reinforcement Learning

Behavioral cloning (BC) and inverse reinforcement learning (IRL) methods form the two major classes of imitation learning methods. As discussed in § 2, "BC vs IRL" is the first question that one needs to answer when applying imitation learning to the problem at hand.

Recovering the reward function can be interpreted as inferring the expert's intent since the reward function encodes the objective for the desired task. For example, when learning from a sequence of images without kinematic information of the expert, it is not clear how to apply behavioral cloning. In such a case, we need to infer what is desired by the expert and then estimate a policy to achieve the inferred goal.

For example, to address the problem of imitation from observation, the recent work by Sermanet et al. [2017] and Liu et al. [2017] proposed methods for recovering the reward function from visual features extracted by deep neural networks. Thus, IRL is a reasonable choice for such problems where inference of the expert's intent is necessary even if the policy itself is more compact than a reward function.

When both behavioral cloning or inverse reinforcement learning can be applied to a given problem, it is essential to consider "*what is the most parsimonious description of the desired behavior, reward or policy?*". Ho and Ermon [2016] recently indicated that under the maximum entropy assumption recovering the reward function is the dual of matching the expectation of states and actions. This implies that BC and IRL can be equivalent under certain assumptions since BC methods learn a policy by matching the expectation of states and actions and IRL methods learn a policy based on the reward function recovered by matching the expectation of states and actions. Since IRL recovers the "hidden" reward function, IRL often adds complexity to the solution approach compared to BC. Thus, in order to select BC or IRL, it is essential to clarify whether recovering the reward function is beneficial or not.

For instance, recovering a reward function for a manipulation task is often difficult since it is not trivial to extract features of the given scene which are relevant to the task. On the other hand, the distribution of the demonstrated trajectories for manipulation can be often learned without recovering the reward function. When the distribution of necessary trajectories can be predicted for a given context, the task can be performed without any knowledge about the reward function of the task. In this case, the distribution of the demonstrated trajectories can be considered a parsimonious description of the desired behavior.

As another example, learning a reward function for footstep planning for a quadruped robot enables generalizing the footstep planning strategy to different terrains. If the reward function that tells "which footstep location is stable" is recovered, footstep locations can be adaptively selected based on this criteria. Such generalization is hard to obtain if we only learn the distribution of the footstep locations. In this case, the reward function is considered a parsimonious description of the desired behavior, which enables good generalization of skills.

Overall, the answer to the question "BC vs IRL" totally depends on the problem setting. It is essential to analyze what and how the task should be performed when applying imitation learning methods.

## 5.2 Open Questions in Imitation Learning

We have discussed the state of the art in imitation learning in this survey. Although imitation learning methods so far have demonstrated great capability, it is clear that there still exists several challenges to be solved. In this section, we highlight open questions in imitation learning and try to clarify what problems need to be solved.

### 5.2.1 Problems Related to Demonstrated Data

The first step of imitation learning is to collect expert demonstration data. However, it is often not trivial to obtain appropriate data to achieve satisfactory performance in imitation learning. Below we list questions related to data collection.

**How to learn from multiple experts?** It is known that imitation learning methods work well for demonstrations performed by one expert rather than multiple experts [Camacho and Michie, 1995]. Therefore, when multiple human experts give instructions to a robotic system, one could extract one expert from multiple experts. However, this problem has not been sufficiently addressed.

**How to deal with undesirable motions in demonstrations?** Many imitation learning methods assume that demonstrated behavior is (sub-)optimal. However, in practice, demonstrated behavior often contains undesirable motions which may may result in low performance policies. To address this issue, reinforcement learning can be used to improve the learned policy [Kober et al., 2013, Mnih et al., 2015, Silver et al., 2016]. Nevertheless, explicitly detecting unnecessary motion and removing it from demonstrated behavior is still an open problem.

**How to learn from raw sensory inputs without embodiment information?** When learning only from vision we cannot directly measure the kinematic information of the expert. While learning from raw sensory inputs without embodiment information is challenging, humans can do it based on prior knowledge. Recent work by Sermanet et al. [2017] shows that the reward function can be inferred from few demonstrations by using visual representations learned by deep models.

**How to deal with different viewpoints?** Current imitation learning methods are usually limited to the case where the demonstration is supplied in the first-person, i.e., a sequence of states and actions is provided similarly to how the learner would observe the task. However, humans can learn by observing the behavior of other humans. When learning from the third-person view it is necessary to infer how the task should be performed. Recent work on third-person imitation learning [Stadie et al., 2017] addresses this problem in some simple environments.

**How to leverage past demonstrations of other related tasks, to learn more quickly the current task?** While it is challenging to learn a very complex task from one demonstration, humans can learn from few demonstrations because they have so much prior knowledge. In principle, this knowledge could be captured and reused for other tasks. Recent work such as [Gupta et al., 2017, Finn et al., 2017a,b, Duan et al., 2017] addresses this research direction.

### 5.2.2 Open Questions Related to Design Choices

When we implement imitation learning in an actual robotic system, we need to make several design choices as we discussed in Chapter 2. There are still several open questions when making such design choices.

**What is the best similarity measure of policies?** To obtain a policy that imitates experts' behavior, it is essential to measure the similarity of policies. Although we discussed some similarity measure such as KL divergence and Euclidean distance, there exist many other options. For example, recently the Wasserstein divergence (aka Earth-mover distance) [Arjovsky et al., 2017] has been shown to improve the performance of generative adversarial networks (GANs) [Goodfellow et al., 2014] which have inspired some recent imitation learning approaches [Ho and Ermon, 2016, Finn et al., 2016a]. Exploring new similarity measures is a promising way to discover new imitation learning methods which may work in situations not handled by current methods.

**How to learn from multiple instruction types?** In practice, various types of instructions are available, such as corrective motion from operators, preferences on optional actions and evaluation of the performance. To achieve intuitive human-robot interaction and efficient learning, it is necessary to utilize various instruction types. Although some methods incorporate multiple instruction types Jain et al. [2015], this research direction has not been well-investigated yet.

**How to incorporate prior knowledge? How to do it explicitly?** Although prior knowledge of the system or environment, e.g., kinematics and the mass of a manipulator, are often available, many imitation learning methods utilize only demonstrations. However, incorporating available prior knowledge will be useful for system control and trajectory planning. On the other hand, many methods use implicit prior knowledge such as assuming a Gaussian distribution of samples. Methods that explicitly incorporate prior knowledge could lower the amount of demonstration data required and make new robotic applications possible.

**How to learn from various sensors?** Many studies on imitation learning implicitly select sensory information appropriate for their method. However, in practice, we can use various redundant sensory information such as tactile information, RGB-D images, audio information, and encoders in robot joints. Fusing of various sensory information will lead to more robust and adaptive behavior.

**How to learn tasks humans cannot do?** Imitation learning methods assume that demonstrations of the desired task are available. However, it is often the case that human operators cannot appropriately demonstrate the given task, especially in cases where a robot has a physical advantage compared to a human. For example, a robotic system may have more than two arms making it challenging for the human operator to demonstrate the desired behavior. To achieve performance beyond human capability, methods that iteratively improve the performance of the system will be necessary.

**How to choose a trajectory representation?** In §3.5.1, we discussed several different trajectory representations. An interesting open question is how to choose among the trajectory representations. We gave in §3.5.2 some suggestions how to choose based on the different properties of the representations. However, there is no definite answer on how to select a trajacectory presentation. Note that choosing a trajectory representation is analogous to model selection in machine learning [Bishop, 2006]. Considering trajectory representation selection as a model selection problem could lead to interesting advances.

### 5.2.3 Problems Related to Algorithms

When we want to overcome limitations in current imitation learning, we also need to face several open questions related to algorithmic aspects of imitation learning.

**How to generalize skills with complex conditions?** Many methods model the distribution over demonstrated trajectories and generalize the skill by conditioning the distribution Khansari-Zadeh and Billard [2011], Paraschos et al. [2013] for example on different start or end positions. However, such methods might not scale to high dimensional conditions. Although some work addresses scaling up generalization of skills with high dimensional inputs Schulman et al.

[2013], further investigation is necessary. Recent work by Finn et al. [2017b], Sermanet et al. [2017], Liu et al. [2017], and Rahmatizadeh et al. [2017] proposed methods for learning from visual information using deep neural networks, which is a promising way to address the skill generalization with complex conditions.

**How to find solutions with guarantees?** In current imitation learning, there are performance guarantees, e.g., stability of DMPs [Ijspeert et al., 2002a, Schaal et al., 2004] and a proof of low error in DAGGER [Ross et al., 2011]. However, currently, for many imitation learning methods there are no performance guarantees. Especially in robotics, guarantees such as stability or convergence can be very important in practice. Finding guarantees for common imitation learning methods is a worthy research direction.

**How to scale up with respect to the number of dimensions?** Motion planning in a robotic system requires a high dimensional solution. For example, a humanoid robot often has over 50 joints. However, existing imitation learning methods are often inefficient for such high dimensional motion due to the different embodiment of the learner and the expert. Recent studies show that the dimensionality of the input space can be scaled up using convolutional neural networks. However, current methods for high dimensional inputs are often limited to 2D images. Incorporating high dimensional sensory inputs is still an open question. In addition, scaling up the dimensionality of actions is also an open problem. Incorporating dimensionality reduction in imitation learning is an interesting research direction [Sugiyama et al., 2010, Tangkaratt et al., 2015]

**How to find globally optimal solutions in high dimensional spaces? How to make it tractable?** In robotic applications, it is essential to find solutions in a continuous and high dimensional space. Many imitation learning methods find locally optimal solutions close to the behavior demonstrated by experts. However, there may exist a better solution which is different from the demonstrated behavior.

**How to perform imitation by multiple agents?** In multi-agent domains, an agent needs to consider how the other agents' behavior may influence the outcome. Prior work [Waugh et al., 2011, Kuleshov and Schrijvers, 2015] addresses how to infer the reward function, which represents the equilibrium of agents' strategies, from observed behavior of multiple agents. However, the results are still quite limited to simple problem settings and have not migrated to large scale robot applications.

**How to perform incremental/active learning in IRL?**
Although many inverse reinforcement learning (IRL) methods assume a sufficient number of demonstrations, it is often not the case in practice. When the policy learned from the initial dataset of demonstrations does not show satisfactory performance, the policy can be incrementally improved. Silver et al. [2012], Lopes et al. [2009] proposed methods for IRL with active learning. Such incremental IRL methods have not been investigated sufficiently.

### 5.2.4 Performance Evaluation

Since the purpose and target applications of imitation learning are very broad, benchmarking imitation learning methods can be challenging. The following open questions are related to performance evaluation in imitation learning.

**How to establish benchmark problems for imitation learning?**
Unlike other machine learning fields, there is no widely accepted set of benchmark problems for imitation learning. Although efforts for benchmarking different techniques have been made, e.g. [Lemme et al., 2015], there is no clear way to compare performance between methods. Benchmark problems such as data mining and computer vision communities should be established.

**What metric should be used to evaluate imitation learning**

**methods?** There are various ways to quantify imitation learning performance. However, there is no established way to evaluate imitation learning methods, nor are there yet large scale benchmarks that make it effective and easy to compare and contrast approaches.

# Acknowledgements

The research leading to this work has received funding from the European Union's Horizon 2020 research and innovation programme under grant agreements #645582 (RoMaNS) and #640554 (SKILLS4ROBOTS). J. A. Bagnell's gratefully acknowledges the support of the National Science Foundation's NRI grant on "National Robotics Initiative: Purposeful Prediction" (Grant 1227495) and the Office of Naval Research's (Grant N000141512365) "Learning to Reason with Inference Machines".

T. Osa was supported by JST CREST Grant Number JPMJCR1403 and KAKENHI Grant Number 17H00757. P. Abbeel acknowledges the Office of Naval Research's Presidential Early Career Awards for Scientists and Engineers (PECASE).